\pdfoutput=1

\documentclass[11pt]{article}

\usepackage[]{ACL2023}

\usepackage{amsmath,amsfonts,bm}

\def\eqref#1{equation~\ref{#1}}

\def\1{\bm{1}}

\def\vb{{\bm{b}}}
\def\vc{{\bm{c}}}

\def\ve{{\bm{e}}}

\def\vk{{\bm{k}}}

\def\vo{{\bm{o}}}

\def\vv{{\bm{v}}}

\def\vx{{\bm{x}}}
\def\vy{{\bm{y}}}
\def\vz{{\bm{z}}}

\def\mA{{\bm{A}}}
\def\mB{{\bm{B}}}
\def\mC{{\bm{C}}}

\def\mL{{\bm{L}}}
\def\mM{{\bm{M}}}

\def\mU{{\bm{U}}}

\def\mW{{\bm{W}}}
\def\mX{{\bm{X}}}

\DeclareMathAlphabet{\mathsfit}{\encodingdefault}{\sfdefault}{m}{sl}
\SetMathAlphabet{\mathsfit}{bold}{\encodingdefault}{\sfdefault}{bx}{n}

\usepackage{times}
\usepackage{latexsym}

\usepackage[T1]{fontenc}

\usepackage[utf8]{inputenc}

\usepackage{microtype}

\usepackage{inconsolata}
\usepackage{times}
\usepackage{latexsym}
\usepackage{comment}
\usepackage{amsmath,amsfonts,amssymb,amsthm,mathtools}
\usepackage{comment}
\usepackage{graphicx}
\usepackage{xcolor}
\usepackage{color,soul}
\usepackage{booktabs}
\usepackage{mathtools}
\usepackage{linguex}
\usepackage{subfigure}
\usepackage{array}
\usepackage{float}
\usepackage{bm}
\usepackage{arydshln}
\usepackage{tikz}
\usepackage{multirow}

\allowdisplaybreaks

\definecolor{red_max}{HTML}{800000}
\definecolor{blue_max}{HTML}{00004C}

\usepackage{capt-of}

\usepackage{tikz-dependency}

\newcommand{\eos}{\texttt{<\text{/s}>}}

\newcommand\numberthis{\addtocounter{equation}{1}\tag{\theequation}}

\definecolor{colorpath1}{HTML}{999999}
\definecolor{colorpath2}{HTML}{1764AB}
\definecolor{colorpath3}{HTML}{E27878}
\definecolor{yellowpaper}{HTML}{D6B656}
\definecolor{bluepaper}{HTML}{6C8EBF}
\definecolor{greypaper}{HTML}{5C5C5C}

\definecolor{custom_green}{HTML}{44AE3D}
\definecolor{custom_red}{HTML}{d62728}
\definecolor{color0}{HTML}{FFFFFF}
\definecolor{color1}{HTML}{00004C}
\definecolor{color2}{HTML}{00004C}
\definecolor{color3}{HTML}{00004C}
\definecolor{color4}{HTML}{00004F}
\definecolor{color5}{HTML}{00004F}
\definecolor{color6}{HTML}{00004F}
\definecolor{color7}{HTML}{00004F}
\definecolor{color8}{HTML}{000052}
\definecolor{color9}{HTML}{000052}
\definecolor{color10}{HTML}{000052}
\definecolor{color11}{HTML}{000052}
\definecolor{color12}{HTML}{000055}
\definecolor{color13}{HTML}{000055}
\definecolor{color14}{HTML}{000055}
\definecolor{color15}{HTML}{000055}
\definecolor{color16}{HTML}{000058}
\definecolor{color17}{HTML}{000058}
\definecolor{color18}{HTML}{000058}
\definecolor{color19}{HTML}{000058}
\definecolor{color20}{HTML}{00005A}
\definecolor{color21}{HTML}{00005A}
\definecolor{color22}{HTML}{00005A}
\definecolor{color23}{HTML}{00005A}
\definecolor{color24}{HTML}{00005D}
\definecolor{color25}{HTML}{00005D}
\definecolor{color26}{HTML}{00005D}
\definecolor{color27}{HTML}{00005D}
\definecolor{color28}{HTML}{000060}
\definecolor{color29}{HTML}{000060}
\definecolor{color30}{HTML}{000060}
\definecolor{color31}{HTML}{000060}
\definecolor{color32}{HTML}{000063}
\definecolor{color33}{HTML}{000063}
\definecolor{color34}{HTML}{000063}
\definecolor{color35}{HTML}{000063}
\definecolor{color36}{HTML}{000066}
\definecolor{color37}{HTML}{000066}
\definecolor{color38}{HTML}{000066}
\definecolor{color39}{HTML}{000066}
\definecolor{color40}{HTML}{000068}
\definecolor{color41}{HTML}{000068}
\definecolor{color42}{HTML}{000068}
\definecolor{color43}{HTML}{00006B}
\definecolor{color44}{HTML}{00006B}
\definecolor{color45}{HTML}{00006B}
\definecolor{color46}{HTML}{00006B}
\definecolor{color47}{HTML}{00006E}
\definecolor{color48}{HTML}{00006E}
\definecolor{color49}{HTML}{00006E}
\definecolor{color50}{HTML}{00006E}
\definecolor{color51}{HTML}{000071}
\definecolor{color52}{HTML}{000071}
\definecolor{color53}{HTML}{000071}
\definecolor{color54}{HTML}{000071}
\definecolor{color55}{HTML}{000074}
\definecolor{color56}{HTML}{000074}
\definecolor{color57}{HTML}{000074}
\definecolor{color58}{HTML}{000074}
\definecolor{color59}{HTML}{000076}
\definecolor{color60}{HTML}{000076}
\definecolor{color61}{HTML}{000076}
\definecolor{color62}{HTML}{000076}
\definecolor{color63}{HTML}{000079}
\definecolor{color64}{HTML}{000079}
\definecolor{color65}{HTML}{000079}
\definecolor{color66}{HTML}{000079}
\definecolor{color67}{HTML}{00007C}
\definecolor{color68}{HTML}{00007C}
\definecolor{color69}{HTML}{00007C}
\definecolor{color70}{HTML}{00007C}
\definecolor{color71}{HTML}{00007F}
\definecolor{color72}{HTML}{00007F}
\definecolor{color73}{HTML}{00007F}
\definecolor{color74}{HTML}{00007F}
\definecolor{color75}{HTML}{000082}
\definecolor{color76}{HTML}{000082}
\definecolor{color77}{HTML}{000082}
\definecolor{color78}{HTML}{000082}
\definecolor{color79}{HTML}{000084}
\definecolor{color80}{HTML}{000084}
\definecolor{color81}{HTML}{000084}
\definecolor{color82}{HTML}{000084}
\definecolor{color83}{HTML}{000087}
\definecolor{color84}{HTML}{000087}
\definecolor{color85}{HTML}{000087}
\definecolor{color86}{HTML}{00008A}
\definecolor{color87}{HTML}{00008A}
\definecolor{color88}{HTML}{00008A}
\definecolor{color89}{HTML}{00008A}
\definecolor{color90}{HTML}{00008D}
\definecolor{color91}{HTML}{00008D}
\definecolor{color92}{HTML}{00008D}
\definecolor{color93}{HTML}{00008D}
\definecolor{color94}{HTML}{000090}
\definecolor{color95}{HTML}{000090}
\definecolor{color96}{HTML}{000090}
\definecolor{color97}{HTML}{000090}
\definecolor{color98}{HTML}{000092}
\definecolor{color99}{HTML}{000092}
\definecolor{color100}{HTML}{000092}
\definecolor{color101}{HTML}{000092}
\definecolor{color102}{HTML}{000095}
\definecolor{color103}{HTML}{000095}
\definecolor{color104}{HTML}{000095}
\definecolor{color105}{HTML}{000095}
\definecolor{color106}{HTML}{000098}
\definecolor{color107}{HTML}{000098}
\definecolor{color108}{HTML}{000098}
\definecolor{color109}{HTML}{000098}
\definecolor{color110}{HTML}{00009B}
\definecolor{color111}{HTML}{00009B}
\definecolor{color112}{HTML}{00009B}
\definecolor{color113}{HTML}{00009B}
\definecolor{color114}{HTML}{00009E}
\definecolor{color115}{HTML}{00009E}
\definecolor{color116}{HTML}{00009E}
\definecolor{color117}{HTML}{00009E}
\definecolor{color118}{HTML}{0000A0}
\definecolor{color119}{HTML}{0000A0}
\definecolor{color120}{HTML}{0000A0}
\definecolor{color121}{HTML}{0000A0}
\definecolor{color122}{HTML}{0000A3}
\definecolor{color123}{HTML}{0000A3}
\definecolor{color124}{HTML}{0000A3}
\definecolor{color125}{HTML}{0000A6}
\definecolor{color126}{HTML}{0000A6}
\definecolor{color127}{HTML}{0000A6}
\definecolor{color128}{HTML}{0000A6}
\definecolor{color129}{HTML}{0000A9}
\definecolor{color130}{HTML}{0000A9}
\definecolor{color131}{HTML}{0000A9}
\definecolor{color132}{HTML}{0000A9}
\definecolor{color133}{HTML}{0000AC}
\definecolor{color134}{HTML}{0000AC}
\definecolor{color135}{HTML}{0000AC}
\definecolor{color136}{HTML}{0000AC}
\definecolor{color137}{HTML}{0000AE}
\definecolor{color138}{HTML}{0000AE}
\definecolor{color139}{HTML}{0000AE}
\definecolor{color140}{HTML}{0000AE}
\definecolor{color141}{HTML}{0000B1}
\definecolor{color142}{HTML}{0000B1}
\definecolor{color143}{HTML}{0000B1}
\definecolor{color144}{HTML}{0000B1}
\definecolor{color145}{HTML}{0000B4}
\definecolor{color146}{HTML}{0000B4}
\definecolor{color147}{HTML}{0000B4}
\definecolor{color148}{HTML}{0000B4}
\definecolor{color149}{HTML}{0000B7}
\definecolor{color150}{HTML}{0000B7}
\definecolor{color151}{HTML}{0000B7}
\definecolor{color152}{HTML}{0000B7}
\definecolor{color153}{HTML}{0000BA}
\definecolor{color154}{HTML}{0000BA}
\definecolor{color155}{HTML}{0000BA}
\definecolor{color156}{HTML}{0000BA}
\definecolor{color157}{HTML}{0000BC}
\definecolor{color158}{HTML}{0000BC}
\definecolor{color159}{HTML}{0000BC}
\definecolor{color160}{HTML}{0000BC}
\definecolor{color161}{HTML}{0000BF}
\definecolor{color162}{HTML}{0000BF}
\definecolor{color163}{HTML}{0000BF}
\definecolor{color164}{HTML}{0000BF}
\definecolor{color165}{HTML}{0000C2}
\definecolor{color166}{HTML}{0000C2}
\definecolor{color167}{HTML}{0000C2}
\definecolor{color168}{HTML}{0000C5}
\definecolor{color169}{HTML}{0000C5}
\definecolor{color170}{HTML}{0000C5}
\definecolor{color171}{HTML}{0000C5}
\definecolor{color172}{HTML}{0000C8}
\definecolor{color173}{HTML}{0000C8}
\definecolor{color174}{HTML}{0000C8}
\definecolor{color175}{HTML}{0000C8}
\definecolor{color176}{HTML}{0000CA}
\definecolor{color177}{HTML}{0000CA}
\definecolor{color178}{HTML}{0000CA}
\definecolor{color179}{HTML}{0000CA}
\definecolor{color180}{HTML}{0000CD}
\definecolor{color181}{HTML}{0000CD}
\definecolor{color182}{HTML}{0000CD}
\definecolor{color183}{HTML}{0000CD}
\definecolor{color184}{HTML}{0000D0}
\definecolor{color185}{HTML}{0000D0}
\definecolor{color186}{HTML}{0000D0}
\definecolor{color187}{HTML}{0000D0}
\definecolor{color188}{HTML}{0000D3}
\definecolor{color189}{HTML}{0000D3}
\definecolor{color190}{HTML}{0000D3}
\definecolor{color191}{HTML}{0000D3}
\definecolor{color192}{HTML}{0000D6}
\definecolor{color193}{HTML}{0000D6}
\definecolor{color194}{HTML}{0000D6}
\definecolor{color195}{HTML}{0000D6}
\definecolor{color196}{HTML}{0000D8}
\definecolor{color197}{HTML}{0000D8}
\definecolor{color198}{HTML}{0000D8}
\definecolor{color199}{HTML}{0000D8}
\definecolor{color200}{HTML}{0000DB}
\definecolor{color201}{HTML}{0000DB}
\definecolor{color202}{HTML}{0000DB}
\definecolor{color203}{HTML}{0000DB}
\definecolor{color204}{HTML}{0000DE}
\definecolor{color205}{HTML}{0000DE}
\definecolor{color206}{HTML}{0000DE}
\definecolor{color207}{HTML}{0000DE}
\definecolor{color208}{HTML}{0000E1}
\definecolor{color209}{HTML}{0000E1}
\definecolor{color210}{HTML}{0000E1}
\definecolor{color211}{HTML}{0000E4}
\definecolor{color212}{HTML}{0000E4}
\definecolor{color213}{HTML}{0000E4}
\definecolor{color214}{HTML}{0000E4}
\definecolor{color215}{HTML}{0000E6}
\definecolor{color216}{HTML}{0000E6}
\definecolor{color217}{HTML}{0000E6}
\definecolor{color218}{HTML}{0000E6}
\definecolor{color219}{HTML}{0000E9}
\definecolor{color220}{HTML}{0000E9}
\definecolor{color221}{HTML}{0000E9}
\definecolor{color222}{HTML}{0000E9}
\definecolor{color223}{HTML}{0000EC}
\definecolor{color224}{HTML}{0000EC}
\definecolor{color225}{HTML}{0000EC}
\definecolor{color226}{HTML}{0000EC}
\definecolor{color227}{HTML}{0000EF}
\definecolor{color228}{HTML}{0000EF}
\definecolor{color229}{HTML}{0000EF}
\definecolor{color230}{HTML}{0000EF}
\definecolor{color231}{HTML}{0000F2}
\definecolor{color232}{HTML}{0000F2}
\definecolor{color233}{HTML}{0000F2}
\definecolor{color234}{HTML}{0000F2}
\definecolor{color235}{HTML}{0000F4}
\definecolor{color236}{HTML}{0000F4}
\definecolor{color237}{HTML}{0000F4}
\definecolor{color238}{HTML}{0000F4}
\definecolor{color239}{HTML}{0000F7}
\definecolor{color240}{HTML}{0000F7}
\definecolor{color241}{HTML}{0000F7}
\definecolor{color242}{HTML}{0000F7}
\definecolor{color243}{HTML}{0000FA}
\definecolor{color244}{HTML}{0000FA}
\definecolor{color245}{HTML}{0000FA}
\definecolor{color246}{HTML}{0000FA}
\definecolor{color247}{HTML}{0000FD}
\definecolor{color248}{HTML}{0000FD}
\definecolor{color249}{HTML}{0000FD}
\definecolor{color250}{HTML}{0101FF}
\definecolor{color251}{HTML}{0101FF}
\definecolor{color252}{HTML}{0101FF}
\definecolor{color253}{HTML}{0101FF}
\definecolor{color254}{HTML}{0505FF}
\definecolor{color255}{HTML}{0505FF}
\definecolor{color256}{HTML}{0505FF}
\definecolor{color257}{HTML}{0505FF}
\definecolor{color258}{HTML}{0909FF}
\definecolor{color259}{HTML}{0909FF}
\definecolor{color260}{HTML}{0909FF}
\definecolor{color261}{HTML}{0909FF}
\definecolor{color262}{HTML}{0D0DFF}
\definecolor{color263}{HTML}{0D0DFF}
\definecolor{color264}{HTML}{0D0DFF}
\definecolor{color265}{HTML}{0D0DFF}
\definecolor{color266}{HTML}{1111FF}
\definecolor{color267}{HTML}{1111FF}
\definecolor{color268}{HTML}{1111FF}
\definecolor{color269}{HTML}{1111FF}
\definecolor{color270}{HTML}{1515FF}
\definecolor{color271}{HTML}{1515FF}
\definecolor{color272}{HTML}{1515FF}
\definecolor{color273}{HTML}{1515FF}
\definecolor{color274}{HTML}{1919FF}
\definecolor{color275}{HTML}{1919FF}
\definecolor{color276}{HTML}{1919FF}
\definecolor{color277}{HTML}{1919FF}
\definecolor{color278}{HTML}{1D1DFF}
\definecolor{color279}{HTML}{1D1DFF}
\definecolor{color280}{HTML}{1D1DFF}
\definecolor{color281}{HTML}{1D1DFF}
\definecolor{color282}{HTML}{2121FF}
\definecolor{color283}{HTML}{2121FF}
\definecolor{color284}{HTML}{2121FF}
\definecolor{color285}{HTML}{2121FF}
\definecolor{color286}{HTML}{2525FF}
\definecolor{color287}{HTML}{2525FF}
\definecolor{color288}{HTML}{2525FF}
\definecolor{color289}{HTML}{2525FF}
\definecolor{color290}{HTML}{2929FF}
\definecolor{color291}{HTML}{2929FF}
\definecolor{color292}{HTML}{2929FF}
\definecolor{color293}{HTML}{2D2DFF}
\definecolor{color294}{HTML}{2D2DFF}
\definecolor{color295}{HTML}{2D2DFF}
\definecolor{color296}{HTML}{2D2DFF}
\definecolor{color297}{HTML}{3131FF}
\definecolor{color298}{HTML}{3131FF}
\definecolor{color299}{HTML}{3131FF}
\definecolor{color300}{HTML}{3131FF}
\definecolor{color301}{HTML}{3535FF}
\definecolor{color302}{HTML}{3535FF}
\definecolor{color303}{HTML}{3535FF}
\definecolor{color304}{HTML}{3535FF}
\definecolor{color305}{HTML}{3939FF}
\definecolor{color306}{HTML}{3939FF}
\definecolor{color307}{HTML}{3939FF}
\definecolor{color308}{HTML}{3939FF}
\definecolor{color309}{HTML}{3D3DFF}
\definecolor{color310}{HTML}{3D3DFF}
\definecolor{color311}{HTML}{3D3DFF}
\definecolor{color312}{HTML}{3D3DFF}
\definecolor{color313}{HTML}{4141FF}
\definecolor{color314}{HTML}{4141FF}
\definecolor{color315}{HTML}{4141FF}
\definecolor{color316}{HTML}{4141FF}
\definecolor{color317}{HTML}{4545FF}
\definecolor{color318}{HTML}{4545FF}
\definecolor{color319}{HTML}{4545FF}
\definecolor{color320}{HTML}{4545FF}
\definecolor{color321}{HTML}{4949FF}
\definecolor{color322}{HTML}{4949FF}
\definecolor{color323}{HTML}{4949FF}
\definecolor{color324}{HTML}{4949FF}
\definecolor{color325}{HTML}{4D4DFF}
\definecolor{color326}{HTML}{4D4DFF}
\definecolor{color327}{HTML}{4D4DFF}
\definecolor{color328}{HTML}{4D4DFF}
\definecolor{color329}{HTML}{5151FF}
\definecolor{color330}{HTML}{5151FF}
\definecolor{color331}{HTML}{5151FF}
\definecolor{color332}{HTML}{5151FF}
\definecolor{color333}{HTML}{5555FF}
\definecolor{color334}{HTML}{5555FF}
\definecolor{color335}{HTML}{5555FF}
\definecolor{color336}{HTML}{5959FF}
\definecolor{color337}{HTML}{5959FF}
\definecolor{color338}{HTML}{5959FF}
\definecolor{color339}{HTML}{5959FF}
\definecolor{color340}{HTML}{5D5DFF}
\definecolor{color341}{HTML}{5D5DFF}
\definecolor{color342}{HTML}{5D5DFF}
\definecolor{color343}{HTML}{5D5DFF}
\definecolor{color344}{HTML}{6161FF}
\definecolor{color345}{HTML}{6161FF}
\definecolor{color346}{HTML}{6161FF}
\definecolor{color347}{HTML}{6161FF}
\definecolor{color348}{HTML}{6565FF}
\definecolor{color349}{HTML}{6565FF}
\definecolor{color350}{HTML}{6565FF}
\definecolor{color351}{HTML}{6565FF}
\definecolor{color352}{HTML}{6969FF}
\definecolor{color353}{HTML}{6969FF}
\definecolor{color354}{HTML}{6969FF}
\definecolor{color355}{HTML}{6969FF}
\definecolor{color356}{HTML}{6D6DFF}
\definecolor{color357}{HTML}{6D6DFF}
\definecolor{color358}{HTML}{6D6DFF}
\definecolor{color359}{HTML}{6D6DFF}
\definecolor{color360}{HTML}{7171FF}
\definecolor{color361}{HTML}{7171FF}
\definecolor{color362}{HTML}{7171FF}
\definecolor{color363}{HTML}{7171FF}
\definecolor{color364}{HTML}{7575FF}
\definecolor{color365}{HTML}{7575FF}
\definecolor{color366}{HTML}{7575FF}
\definecolor{color367}{HTML}{7575FF}
\definecolor{color368}{HTML}{7979FF}
\definecolor{color369}{HTML}{7979FF}
\definecolor{color370}{HTML}{7979FF}
\definecolor{color371}{HTML}{7979FF}
\definecolor{color372}{HTML}{7D7DFF}
\definecolor{color373}{HTML}{7D7DFF}
\definecolor{color374}{HTML}{7D7DFF}
\definecolor{color375}{HTML}{8181FF}
\definecolor{color376}{HTML}{8181FF}
\definecolor{color377}{HTML}{8181FF}
\definecolor{color378}{HTML}{8181FF}
\definecolor{color379}{HTML}{8585FF}
\definecolor{color380}{HTML}{8585FF}
\definecolor{color381}{HTML}{8585FF}
\definecolor{color382}{HTML}{8585FF}
\definecolor{color383}{HTML}{8989FF}
\definecolor{color384}{HTML}{8989FF}
\definecolor{color385}{HTML}{8989FF}
\definecolor{color386}{HTML}{8989FF}
\definecolor{color387}{HTML}{8D8DFF}
\definecolor{color388}{HTML}{8D8DFF}
\definecolor{color389}{HTML}{8D8DFF}
\definecolor{color390}{HTML}{8D8DFF}
\definecolor{color391}{HTML}{9191FF}
\definecolor{color392}{HTML}{9191FF}
\definecolor{color393}{HTML}{9191FF}
\definecolor{color394}{HTML}{9191FF}
\definecolor{color395}{HTML}{9595FF}
\definecolor{color396}{HTML}{9595FF}
\definecolor{color397}{HTML}{9595FF}
\definecolor{color398}{HTML}{9595FF}
\definecolor{color399}{HTML}{9999FF}
\definecolor{color400}{HTML}{9999FF}
\definecolor{color401}{HTML}{9999FF}
\definecolor{color402}{HTML}{9999FF}
\definecolor{color403}{HTML}{9D9DFF}
\definecolor{color404}{HTML}{9D9DFF}
\definecolor{color405}{HTML}{9D9DFF}
\definecolor{color406}{HTML}{9D9DFF}
\definecolor{color407}{HTML}{A1A1FF}
\definecolor{color408}{HTML}{A1A1FF}
\definecolor{color409}{HTML}{A1A1FF}
\definecolor{color410}{HTML}{A1A1FF}
\definecolor{color411}{HTML}{A5A5FF}
\definecolor{color412}{HTML}{A5A5FF}
\definecolor{color413}{HTML}{A5A5FF}
\definecolor{color414}{HTML}{A5A5FF}
\definecolor{color415}{HTML}{A9A9FF}
\definecolor{color416}{HTML}{A9A9FF}
\definecolor{color417}{HTML}{A9A9FF}
\definecolor{color418}{HTML}{ADADFF}
\definecolor{color419}{HTML}{ADADFF}
\definecolor{color420}{HTML}{ADADFF}
\definecolor{color421}{HTML}{ADADFF}
\definecolor{color422}{HTML}{B1B1FF}
\definecolor{color423}{HTML}{B1B1FF}
\definecolor{color424}{HTML}{B1B1FF}
\definecolor{color425}{HTML}{B1B1FF}
\definecolor{color426}{HTML}{B5B5FF}
\definecolor{color427}{HTML}{B5B5FF}
\definecolor{color428}{HTML}{B5B5FF}
\definecolor{color429}{HTML}{B5B5FF}
\definecolor{color430}{HTML}{B9B9FF}
\definecolor{color431}{HTML}{B9B9FF}
\definecolor{color432}{HTML}{B9B9FF}
\definecolor{color433}{HTML}{B9B9FF}
\definecolor{color434}{HTML}{BDBDFF}
\definecolor{color435}{HTML}{BDBDFF}
\definecolor{color436}{HTML}{BDBDFF}
\definecolor{color437}{HTML}{BDBDFF}
\definecolor{color438}{HTML}{C1C1FF}
\definecolor{color439}{HTML}{C1C1FF}
\definecolor{color440}{HTML}{C1C1FF}
\definecolor{color441}{HTML}{C1C1FF}
\definecolor{color442}{HTML}{C5C5FF}
\definecolor{color443}{HTML}{C5C5FF}
\definecolor{color444}{HTML}{C5C5FF}
\definecolor{color445}{HTML}{C5C5FF}
\definecolor{color446}{HTML}{C9C9FF}
\definecolor{color447}{HTML}{C9C9FF}
\definecolor{color448}{HTML}{C9C9FF}
\definecolor{color449}{HTML}{C9C9FF}
\definecolor{color450}{HTML}{CDCDFF}
\definecolor{color451}{HTML}{CDCDFF}
\definecolor{color452}{HTML}{CDCDFF}
\definecolor{color453}{HTML}{CDCDFF}
\definecolor{color454}{HTML}{D1D1FF}
\definecolor{color455}{HTML}{D1D1FF}
\definecolor{color456}{HTML}{D1D1FF}
\definecolor{color457}{HTML}{D1D1FF}
\definecolor{color458}{HTML}{D5D5FF}
\definecolor{color459}{HTML}{D5D5FF}
\definecolor{color460}{HTML}{D5D5FF}
\definecolor{color461}{HTML}{D9D9FF}
\definecolor{color462}{HTML}{D9D9FF}
\definecolor{color463}{HTML}{D9D9FF}
\definecolor{color464}{HTML}{D9D9FF}
\definecolor{color465}{HTML}{DDDDFF}
\definecolor{color466}{HTML}{DDDDFF}
\definecolor{color467}{HTML}{DDDDFF}
\definecolor{color468}{HTML}{DDDDFF}
\definecolor{color469}{HTML}{E1E1FF}
\definecolor{color470}{HTML}{E1E1FF}
\definecolor{color471}{HTML}{E1E1FF}
\definecolor{color472}{HTML}{E1E1FF}
\definecolor{color473}{HTML}{E5E5FF}
\definecolor{color474}{HTML}{E5E5FF}
\definecolor{color475}{HTML}{E5E5FF}
\definecolor{color476}{HTML}{E5E5FF}
\definecolor{color477}{HTML}{E9E9FF}
\definecolor{color478}{HTML}{E9E9FF}
\definecolor{color479}{HTML}{E9E9FF}
\definecolor{color480}{HTML}{E9E9FF}
\definecolor{color481}{HTML}{EDEDFF}
\definecolor{color482}{HTML}{EDEDFF}
\definecolor{color483}{HTML}{EDEDFF}
\definecolor{color484}{HTML}{EDEDFF}
\definecolor{color485}{HTML}{F1F1FF}
\definecolor{color486}{HTML}{F1F1FF}
\definecolor{color487}{HTML}{F1F1FF}
\definecolor{color488}{HTML}{F1F1FF}
\definecolor{color489}{HTML}{F5F5FF}
\definecolor{color490}{HTML}{F5F5FF}
\definecolor{color491}{HTML}{F5F5FF}
\definecolor{color492}{HTML}{F5F5FF}
\definecolor{color493}{HTML}{F9F9FF}
\definecolor{color494}{HTML}{F9F9FF}
\definecolor{color495}{HTML}{F9F9FF}
\definecolor{color496}{HTML}{F9F9FF}
\definecolor{color497}{HTML}{FDFDFF}
\definecolor{color498}{HTML}{FDFDFF}
\definecolor{color499}{HTML}{FDFDFF}
\definecolor{color500}{HTML}{FFFDFD}
\definecolor{color501}{HTML}{FFFDFD}
\definecolor{color502}{HTML}{FFFDFD}
\definecolor{color503}{HTML}{FFFDFD}
\definecolor{color504}{HTML}{FFF9F9}
\definecolor{color505}{HTML}{FFF9F9}
\definecolor{color506}{HTML}{FFF9F9}
\definecolor{color507}{HTML}{FFF9F9}
\definecolor{color508}{HTML}{FFF5F5}
\definecolor{color509}{HTML}{FFF5F5}
\definecolor{color510}{HTML}{FFF5F5}
\definecolor{color511}{HTML}{FFF5F5}
\definecolor{color512}{HTML}{FFF1F1}
\definecolor{color513}{HTML}{FFF1F1}
\definecolor{color514}{HTML}{FFF1F1}
\definecolor{color515}{HTML}{FFF1F1}
\definecolor{color516}{HTML}{FFEDED}
\definecolor{color517}{HTML}{FFEDED}
\definecolor{color518}{HTML}{FFEDED}
\definecolor{color519}{HTML}{FFEDED}
\definecolor{color520}{HTML}{FFE9E9}
\definecolor{color521}{HTML}{FFE9E9}
\definecolor{color522}{HTML}{FFE9E9}
\definecolor{color523}{HTML}{FFE9E9}
\definecolor{color524}{HTML}{FFE5E5}
\definecolor{color525}{HTML}{FFE5E5}
\definecolor{color526}{HTML}{FFE5E5}
\definecolor{color527}{HTML}{FFE5E5}
\definecolor{color528}{HTML}{FFE1E1}
\definecolor{color529}{HTML}{FFE1E1}
\definecolor{color530}{HTML}{FFE1E1}
\definecolor{color531}{HTML}{FFE1E1}
\definecolor{color532}{HTML}{FFDDDD}
\definecolor{color533}{HTML}{FFDDDD}
\definecolor{color534}{HTML}{FFDDDD}
\definecolor{color535}{HTML}{FFDDDD}
\definecolor{color536}{HTML}{FFD9D9}
\definecolor{color537}{HTML}{FFD9D9}
\definecolor{color538}{HTML}{FFD9D9}
\definecolor{color539}{HTML}{FFD9D9}
\definecolor{color540}{HTML}{FFD5D5}
\definecolor{color541}{HTML}{FFD5D5}
\definecolor{color542}{HTML}{FFD5D5}
\definecolor{color543}{HTML}{FFD1D1}
\definecolor{color544}{HTML}{FFD1D1}
\definecolor{color545}{HTML}{FFD1D1}
\definecolor{color546}{HTML}{FFD1D1}
\definecolor{color547}{HTML}{FFCDCD}
\definecolor{color548}{HTML}{FFCDCD}
\definecolor{color549}{HTML}{FFCDCD}
\definecolor{color550}{HTML}{FFCDCD}
\definecolor{color551}{HTML}{FFC9C9}
\definecolor{color552}{HTML}{FFC9C9}
\definecolor{color553}{HTML}{FFC9C9}
\definecolor{color554}{HTML}{FFC9C9}
\definecolor{color555}{HTML}{FFC5C5}
\definecolor{color556}{HTML}{FFC5C5}
\definecolor{color557}{HTML}{FFC5C5}
\definecolor{color558}{HTML}{FFC5C5}
\definecolor{color559}{HTML}{FFC1C1}
\definecolor{color560}{HTML}{FFC1C1}
\definecolor{color561}{HTML}{FFC1C1}
\definecolor{color562}{HTML}{FFC1C1}
\definecolor{color563}{HTML}{FFBDBD}
\definecolor{color564}{HTML}{FFBDBD}
\definecolor{color565}{HTML}{FFBDBD}
\definecolor{color566}{HTML}{FFBDBD}
\definecolor{color567}{HTML}{FFB9B9}
\definecolor{color568}{HTML}{FFB9B9}
\definecolor{color569}{HTML}{FFB9B9}
\definecolor{color570}{HTML}{FFB9B9}
\definecolor{color571}{HTML}{FFB5B5}
\definecolor{color572}{HTML}{FFB5B5}
\definecolor{color573}{HTML}{FFB5B5}
\definecolor{color574}{HTML}{FFB5B5}
\definecolor{color575}{HTML}{FFB1B1}
\definecolor{color576}{HTML}{FFB1B1}
\definecolor{color577}{HTML}{FFB1B1}
\definecolor{color578}{HTML}{FFB1B1}
\definecolor{color579}{HTML}{FFADAD}
\definecolor{color580}{HTML}{FFADAD}
\definecolor{color581}{HTML}{FFADAD}
\definecolor{color582}{HTML}{FFADAD}
\definecolor{color583}{HTML}{FFA9A9}
\definecolor{color584}{HTML}{FFA9A9}
\definecolor{color585}{HTML}{FFA9A9}
\definecolor{color586}{HTML}{FFA5A5}
\definecolor{color587}{HTML}{FFA5A5}
\definecolor{color588}{HTML}{FFA5A5}
\definecolor{color589}{HTML}{FFA5A5}
\definecolor{color590}{HTML}{FFA1A1}
\definecolor{color591}{HTML}{FFA1A1}
\definecolor{color592}{HTML}{FFA1A1}
\definecolor{color593}{HTML}{FFA1A1}
\definecolor{color594}{HTML}{FF9D9D}
\definecolor{color595}{HTML}{FF9D9D}
\definecolor{color596}{HTML}{FF9D9D}
\definecolor{color597}{HTML}{FF9D9D}
\definecolor{color598}{HTML}{FF9999}
\definecolor{color599}{HTML}{FF9999}
\definecolor{color600}{HTML}{FF9999}
\definecolor{color601}{HTML}{FF9999}
\definecolor{color602}{HTML}{FF9595}
\definecolor{color603}{HTML}{FF9595}
\definecolor{color604}{HTML}{FF9595}
\definecolor{color605}{HTML}{FF9595}
\definecolor{color606}{HTML}{FF9191}
\definecolor{color607}{HTML}{FF9191}
\definecolor{color608}{HTML}{FF9191}
\definecolor{color609}{HTML}{FF9191}
\definecolor{color610}{HTML}{FF8D8D}
\definecolor{color611}{HTML}{FF8D8D}
\definecolor{color612}{HTML}{FF8D8D}
\definecolor{color613}{HTML}{FF8D8D}
\definecolor{color614}{HTML}{FF8989}
\definecolor{color615}{HTML}{FF8989}
\definecolor{color616}{HTML}{FF8989}
\definecolor{color617}{HTML}{FF8989}
\definecolor{color618}{HTML}{FF8585}
\definecolor{color619}{HTML}{FF8585}
\definecolor{color620}{HTML}{FF8585}
\definecolor{color621}{HTML}{FF8585}
\definecolor{color622}{HTML}{FF8181}
\definecolor{color623}{HTML}{FF8181}
\definecolor{color624}{HTML}{FF8181}
\definecolor{color625}{HTML}{FF7D7D}
\definecolor{color626}{HTML}{FF7D7D}
\definecolor{color627}{HTML}{FF7D7D}
\definecolor{color628}{HTML}{FF7D7D}
\definecolor{color629}{HTML}{FF7979}
\definecolor{color630}{HTML}{FF7979}
\definecolor{color631}{HTML}{FF7979}
\definecolor{color632}{HTML}{FF7979}
\definecolor{color633}{HTML}{FF7575}
\definecolor{color634}{HTML}{FF7575}
\definecolor{color635}{HTML}{FF7575}
\definecolor{color636}{HTML}{FF7575}
\definecolor{color637}{HTML}{FF7171}
\definecolor{color638}{HTML}{FF7171}
\definecolor{color639}{HTML}{FF7171}
\definecolor{color640}{HTML}{FF7171}
\definecolor{color641}{HTML}{FF6D6D}
\definecolor{color642}{HTML}{FF6D6D}
\definecolor{color643}{HTML}{FF6D6D}
\definecolor{color644}{HTML}{FF6D6D}
\definecolor{color645}{HTML}{FF6969}
\definecolor{color646}{HTML}{FF6969}
\definecolor{color647}{HTML}{FF6969}
\definecolor{color648}{HTML}{FF6969}
\definecolor{color649}{HTML}{FF6565}
\definecolor{color650}{HTML}{FF6565}
\definecolor{color651}{HTML}{FF6565}
\definecolor{color652}{HTML}{FF6565}
\definecolor{color653}{HTML}{FF6161}
\definecolor{color654}{HTML}{FF6161}
\definecolor{color655}{HTML}{FF6161}
\definecolor{color656}{HTML}{FF6161}
\definecolor{color657}{HTML}{FF5D5D}
\definecolor{color658}{HTML}{FF5D5D}
\definecolor{color659}{HTML}{FF5D5D}
\definecolor{color660}{HTML}{FF5D5D}
\definecolor{color661}{HTML}{FF5959}
\definecolor{color662}{HTML}{FF5959}
\definecolor{color663}{HTML}{FF5959}
\definecolor{color664}{HTML}{FF5959}
\definecolor{color665}{HTML}{FF5555}
\definecolor{color666}{HTML}{FF5555}
\definecolor{color667}{HTML}{FF5555}
\definecolor{color668}{HTML}{FF5151}
\definecolor{color669}{HTML}{FF5151}
\definecolor{color670}{HTML}{FF5151}
\definecolor{color671}{HTML}{FF5151}
\definecolor{color672}{HTML}{FF4D4D}
\definecolor{color673}{HTML}{FF4D4D}
\definecolor{color674}{HTML}{FF4D4D}
\definecolor{color675}{HTML}{FF4D4D}
\definecolor{color676}{HTML}{FF4949}
\definecolor{color677}{HTML}{FF4949}
\definecolor{color678}{HTML}{FF4949}
\definecolor{color679}{HTML}{FF4949}
\definecolor{color680}{HTML}{FF4545}
\definecolor{color681}{HTML}{FF4545}
\definecolor{color682}{HTML}{FF4545}
\definecolor{color683}{HTML}{FF4545}
\definecolor{color684}{HTML}{FF4141}
\definecolor{color685}{HTML}{FF4141}
\definecolor{color686}{HTML}{FF4141}
\definecolor{color687}{HTML}{FF4141}
\definecolor{color688}{HTML}{FF3D3D}
\definecolor{color689}{HTML}{FF3D3D}
\definecolor{color690}{HTML}{FF3D3D}
\definecolor{color691}{HTML}{FF3D3D}
\definecolor{color692}{HTML}{FF3939}
\definecolor{color693}{HTML}{FF3939}
\definecolor{color694}{HTML}{FF3939}
\definecolor{color695}{HTML}{FF3939}
\definecolor{color696}{HTML}{FF3535}
\definecolor{color697}{HTML}{FF3535}
\definecolor{color698}{HTML}{FF3535}
\definecolor{color699}{HTML}{FF3535}
\definecolor{color700}{HTML}{FF3131}
\definecolor{color701}{HTML}{FF3131}
\definecolor{color702}{HTML}{FF3131}
\definecolor{color703}{HTML}{FF3131}
\definecolor{color704}{HTML}{FF2D2D}
\definecolor{color705}{HTML}{FF2D2D}
\definecolor{color706}{HTML}{FF2D2D}
\definecolor{color707}{HTML}{FF2D2D}
\definecolor{color708}{HTML}{FF2929}
\definecolor{color709}{HTML}{FF2929}
\definecolor{color710}{HTML}{FF2929}
\definecolor{color711}{HTML}{FF2525}
\definecolor{color712}{HTML}{FF2525}
\definecolor{color713}{HTML}{FF2525}
\definecolor{color714}{HTML}{FF2525}
\definecolor{color715}{HTML}{FF2121}
\definecolor{color716}{HTML}{FF2121}
\definecolor{color717}{HTML}{FF2121}
\definecolor{color718}{HTML}{FF2121}
\definecolor{color719}{HTML}{FF1D1D}
\definecolor{color720}{HTML}{FF1D1D}
\definecolor{color721}{HTML}{FF1D1D}
\definecolor{color722}{HTML}{FF1D1D}
\definecolor{color723}{HTML}{FF1919}
\definecolor{color724}{HTML}{FF1919}
\definecolor{color725}{HTML}{FF1919}
\definecolor{color726}{HTML}{FF1919}
\definecolor{color727}{HTML}{FF1515}
\definecolor{color728}{HTML}{FF1515}
\definecolor{color729}{HTML}{FF1515}
\definecolor{color730}{HTML}{FF1515}
\definecolor{color731}{HTML}{FF1111}
\definecolor{color732}{HTML}{FF1111}
\definecolor{color733}{HTML}{FF1111}
\definecolor{color734}{HTML}{FF1111}
\definecolor{color735}{HTML}{FF0D0D}
\definecolor{color736}{HTML}{FF0D0D}
\definecolor{color737}{HTML}{FF0D0D}
\definecolor{color738}{HTML}{FF0D0D}
\definecolor{color739}{HTML}{FF0909}
\definecolor{color740}{HTML}{FF0909}
\definecolor{color741}{HTML}{FF0909}
\definecolor{color742}{HTML}{FF0909}
\definecolor{color743}{HTML}{FF0505}
\definecolor{color744}{HTML}{FF0505}
\definecolor{color745}{HTML}{FF0505}
\definecolor{color746}{HTML}{FF0505}
\definecolor{color747}{HTML}{FF0101}
\definecolor{color748}{HTML}{FF0101}
\definecolor{color749}{HTML}{FF0101}
\definecolor{color750}{HTML}{FE0000}
\definecolor{color751}{HTML}{FE0000}
\definecolor{color752}{HTML}{FE0000}
\definecolor{color753}{HTML}{FE0000}
\definecolor{color754}{HTML}{FC0000}
\definecolor{color755}{HTML}{FC0000}
\definecolor{color756}{HTML}{FC0000}
\definecolor{color757}{HTML}{FC0000}
\definecolor{color758}{HTML}{FA0000}
\definecolor{color759}{HTML}{FA0000}
\definecolor{color760}{HTML}{FA0000}
\definecolor{color761}{HTML}{FA0000}
\definecolor{color762}{HTML}{F80000}
\definecolor{color763}{HTML}{F80000}
\definecolor{color764}{HTML}{F80000}
\definecolor{color765}{HTML}{F80000}
\definecolor{color766}{HTML}{F60000}
\definecolor{color767}{HTML}{F60000}
\definecolor{color768}{HTML}{F60000}
\definecolor{color769}{HTML}{F60000}
\definecolor{color770}{HTML}{F40000}
\definecolor{color771}{HTML}{F40000}
\definecolor{color772}{HTML}{F40000}
\definecolor{color773}{HTML}{F40000}
\definecolor{color774}{HTML}{F20000}
\definecolor{color775}{HTML}{F20000}
\definecolor{color776}{HTML}{F20000}
\definecolor{color777}{HTML}{F20000}
\definecolor{color778}{HTML}{F00000}
\definecolor{color779}{HTML}{F00000}
\definecolor{color780}{HTML}{F00000}
\definecolor{color781}{HTML}{F00000}
\definecolor{color782}{HTML}{EE0000}
\definecolor{color783}{HTML}{EE0000}
\definecolor{color784}{HTML}{EE0000}
\definecolor{color785}{HTML}{EE0000}
\definecolor{color786}{HTML}{EC0000}
\definecolor{color787}{HTML}{EC0000}
\definecolor{color788}{HTML}{EC0000}
\definecolor{color789}{HTML}{EC0000}
\definecolor{color790}{HTML}{EA0000}
\definecolor{color791}{HTML}{EA0000}
\definecolor{color792}{HTML}{EA0000}
\definecolor{color793}{HTML}{E80000}
\definecolor{color794}{HTML}{E80000}
\definecolor{color795}{HTML}{E80000}
\definecolor{color796}{HTML}{E80000}
\definecolor{color797}{HTML}{E60000}
\definecolor{color798}{HTML}{E60000}
\definecolor{color799}{HTML}{E60000}
\definecolor{color800}{HTML}{E60000}
\definecolor{color801}{HTML}{E30000}
\definecolor{color802}{HTML}{E30000}
\definecolor{color803}{HTML}{E30000}
\definecolor{color804}{HTML}{E30000}
\definecolor{color805}{HTML}{E20000}
\definecolor{color806}{HTML}{E20000}
\definecolor{color807}{HTML}{E20000}
\definecolor{color808}{HTML}{E20000}
\definecolor{color809}{HTML}{E00000}
\definecolor{color810}{HTML}{E00000}
\definecolor{color811}{HTML}{E00000}
\definecolor{color812}{HTML}{E00000}
\definecolor{color813}{HTML}{DE0000}
\definecolor{color814}{HTML}{DE0000}
\definecolor{color815}{HTML}{DE0000}
\definecolor{color816}{HTML}{DE0000}
\definecolor{color817}{HTML}{DC0000}
\definecolor{color818}{HTML}{DC0000}
\definecolor{color819}{HTML}{DC0000}
\definecolor{color820}{HTML}{DC0000}
\definecolor{color821}{HTML}{DA0000}
\definecolor{color822}{HTML}{DA0000}
\definecolor{color823}{HTML}{DA0000}
\definecolor{color824}{HTML}{DA0000}
\definecolor{color825}{HTML}{D80000}
\definecolor{color826}{HTML}{D80000}
\definecolor{color827}{HTML}{D80000}
\definecolor{color828}{HTML}{D80000}
\definecolor{color829}{HTML}{D60000}
\definecolor{color830}{HTML}{D60000}
\definecolor{color831}{HTML}{D60000}
\definecolor{color832}{HTML}{D60000}
\definecolor{color833}{HTML}{D30000}
\definecolor{color834}{HTML}{D30000}
\definecolor{color835}{HTML}{D30000}
\definecolor{color836}{HTML}{D20000}
\definecolor{color837}{HTML}{D20000}
\definecolor{color838}{HTML}{D20000}
\definecolor{color839}{HTML}{D20000}
\definecolor{color840}{HTML}{D00000}
\definecolor{color841}{HTML}{D00000}
\definecolor{color842}{HTML}{D00000}
\definecolor{color843}{HTML}{D00000}
\definecolor{color844}{HTML}{CE0000}
\definecolor{color845}{HTML}{CE0000}
\definecolor{color846}{HTML}{CE0000}
\definecolor{color847}{HTML}{CE0000}
\definecolor{color848}{HTML}{CC0000}
\definecolor{color849}{HTML}{CC0000}
\definecolor{color850}{HTML}{CC0000}
\definecolor{color851}{HTML}{CC0000}
\definecolor{color852}{HTML}{CA0000}
\definecolor{color853}{HTML}{CA0000}
\definecolor{color854}{HTML}{CA0000}
\definecolor{color855}{HTML}{CA0000}
\definecolor{color856}{HTML}{C80000}
\definecolor{color857}{HTML}{C80000}
\definecolor{color858}{HTML}{C80000}
\definecolor{color859}{HTML}{C80000}
\definecolor{color860}{HTML}{C60000}
\definecolor{color861}{HTML}{C60000}
\definecolor{color862}{HTML}{C60000}
\definecolor{color863}{HTML}{C60000}
\definecolor{color864}{HTML}{C30000}
\definecolor{color865}{HTML}{C30000}
\definecolor{color866}{HTML}{C30000}
\definecolor{color867}{HTML}{C30000}
\definecolor{color868}{HTML}{C20000}
\definecolor{color869}{HTML}{C20000}
\definecolor{color870}{HTML}{C20000}
\definecolor{color871}{HTML}{C20000}
\definecolor{color872}{HTML}{C00000}
\definecolor{color873}{HTML}{C00000}
\definecolor{color874}{HTML}{C00000}
\definecolor{color875}{HTML}{BE0000}
\definecolor{color876}{HTML}{BE0000}
\definecolor{color877}{HTML}{BE0000}
\definecolor{color878}{HTML}{BE0000}
\definecolor{color879}{HTML}{BC0000}
\definecolor{color880}{HTML}{BC0000}
\definecolor{color881}{HTML}{BC0000}
\definecolor{color882}{HTML}{BC0000}
\definecolor{color883}{HTML}{BA0000}
\definecolor{color884}{HTML}{BA0000}
\definecolor{color885}{HTML}{BA0000}
\definecolor{color886}{HTML}{BA0000}
\definecolor{color887}{HTML}{B80000}
\definecolor{color888}{HTML}{B80000}
\definecolor{color889}{HTML}{B80000}
\definecolor{color890}{HTML}{B80000}
\definecolor{color891}{HTML}{B60000}
\definecolor{color892}{HTML}{B60000}
\definecolor{color893}{HTML}{B60000}
\definecolor{color894}{HTML}{B60000}
\definecolor{color895}{HTML}{B30000}
\definecolor{color896}{HTML}{B30000}
\definecolor{color897}{HTML}{B30000}
\definecolor{color898}{HTML}{B30000}
\definecolor{color899}{HTML}{B20000}
\definecolor{color900}{HTML}{B20000}
\definecolor{color901}{HTML}{B20000}
\definecolor{color902}{HTML}{B20000}
\definecolor{color903}{HTML}{B00000}
\definecolor{color904}{HTML}{B00000}
\definecolor{color905}{HTML}{B00000}
\definecolor{color906}{HTML}{B00000}
\definecolor{color907}{HTML}{AE0000}
\definecolor{color908}{HTML}{AE0000}
\definecolor{color909}{HTML}{AE0000}
\definecolor{color910}{HTML}{AE0000}
\definecolor{color911}{HTML}{AC0000}
\definecolor{color912}{HTML}{AC0000}
\definecolor{color913}{HTML}{AC0000}
\definecolor{color914}{HTML}{AC0000}
\definecolor{color915}{HTML}{AA0000}
\definecolor{color916}{HTML}{AA0000}
\definecolor{color917}{HTML}{AA0000}
\definecolor{color918}{HTML}{A80000}
\definecolor{color919}{HTML}{A80000}
\definecolor{color920}{HTML}{A80000}
\definecolor{color921}{HTML}{A80000}
\definecolor{color922}{HTML}{A60000}
\definecolor{color923}{HTML}{A60000}
\definecolor{color924}{HTML}{A60000}
\definecolor{color925}{HTML}{A60000}
\definecolor{color926}{HTML}{A30000}
\definecolor{color927}{HTML}{A30000}
\definecolor{color928}{HTML}{A30000}
\definecolor{color929}{HTML}{A30000}
\definecolor{color930}{HTML}{A20000}
\definecolor{color931}{HTML}{A20000}
\definecolor{color932}{HTML}{A20000}
\definecolor{color933}{HTML}{A20000}
\definecolor{color934}{HTML}{A00000}
\definecolor{color935}{HTML}{A00000}
\definecolor{color936}{HTML}{A00000}
\definecolor{color937}{HTML}{A00000}
\definecolor{color938}{HTML}{9E0000}
\definecolor{color939}{HTML}{9E0000}
\definecolor{color940}{HTML}{9E0000}
\definecolor{color941}{HTML}{9E0000}
\definecolor{color942}{HTML}{9C0000}
\definecolor{color943}{HTML}{9C0000}
\definecolor{color944}{HTML}{9C0000}
\definecolor{color945}{HTML}{9C0000}
\definecolor{color946}{HTML}{9A0000}
\definecolor{color947}{HTML}{9A0000}
\definecolor{color948}{HTML}{9A0000}
\definecolor{color949}{HTML}{9A0000}
\definecolor{color950}{HTML}{980000}
\definecolor{color951}{HTML}{980000}
\definecolor{color952}{HTML}{980000}
\definecolor{color953}{HTML}{980000}
\definecolor{color954}{HTML}{960000}
\definecolor{color955}{HTML}{960000}
\definecolor{color956}{HTML}{960000}
\definecolor{color957}{HTML}{960000}
\definecolor{color958}{HTML}{930000}
\definecolor{color959}{HTML}{930000}
\definecolor{color960}{HTML}{930000}
\definecolor{color961}{HTML}{920000}
\definecolor{color962}{HTML}{920000}
\definecolor{color963}{HTML}{920000}
\definecolor{color964}{HTML}{920000}
\definecolor{color965}{HTML}{900000}
\definecolor{color966}{HTML}{900000}
\definecolor{color967}{HTML}{900000}
\definecolor{color968}{HTML}{900000}
\definecolor{color969}{HTML}{8E0000}
\definecolor{color970}{HTML}{8E0000}
\definecolor{color971}{HTML}{8E0000}
\definecolor{color972}{HTML}{8E0000}
\definecolor{color973}{HTML}{8C0000}
\definecolor{color974}{HTML}{8C0000}
\definecolor{color975}{HTML}{8C0000}
\definecolor{color976}{HTML}{8C0000}
\definecolor{color977}{HTML}{8A0000}
\definecolor{color978}{HTML}{8A0000}
\definecolor{color979}{HTML}{8A0000}
\definecolor{color980}{HTML}{8A0000}
\definecolor{color981}{HTML}{880000}
\definecolor{color982}{HTML}{880000}
\definecolor{color983}{HTML}{880000}
\definecolor{color984}{HTML}{880000}
\definecolor{color985}{HTML}{860000}
\definecolor{color986}{HTML}{860000}
\definecolor{color987}{HTML}{860000}
\definecolor{color988}{HTML}{860000}
\definecolor{color989}{HTML}{840000}
\definecolor{color990}{HTML}{840000}
\definecolor{color991}{HTML}{840000}
\definecolor{color992}{HTML}{840000}
\definecolor{color993}{HTML}{820000}
\definecolor{color994}{HTML}{820000}
\definecolor{color995}{HTML}{820000}
\definecolor{color996}{HTML}{820000}
\definecolor{color997}{HTML}{800000}
\definecolor{color998}{HTML}{800000}
\definecolor{color999}{HTML}{800000}
\definecolor{color1000}{HTML}{800000}

\newcommand*{\mybox}[2]{{\setlength{\fboxsep}{0.15pt}\colorbox{#1}{\strut #2}}}

\usepackage{nameref}
\usepackage{cleveref}

\crefformat{section}{\S#2#1#3}
\crefformat{subsection}{\S#2#1#3}
\crefformat{subsubsection}{\S#2#1#3}
\crefformat{paragraph}{\P#2#1#3}
\crefformat{subparagraph}{\P#2#1#3}
\crefmultiformat{section}{\S#2#1#3}{ and~\S#2#1#3}{, \S#2#1#3}{, and~\S#2#1#3}
\crefmultiformat{subsection}{\S#2#1#3}{ and~\S#2#1#3}{, \S#2#1#3}{, and~\S#2#1#3}
\crefmultiformat{subsubsection}{\S#2#1#3}{ and~\S#2#1#3}{, \S#2#1#3}{, and~\S#2#1#3}
\crefmultiformat{paragraph}{\P\P#2#1#3}{ and~#2#1#3}{, #2#1#3}{, and~#2#1#3}
\crefmultiformat{subparagraph}{\P\P#2#1#3}{ and~#2#1#3}{, #2#1#3}{, and~#2#1#3}
\crefrangeformat{section}{\mbox{\S\S#3#1#4--#5#2#6}}
\crefrangeformat{subsection}{\mbox{\S\S#3#1#4--#5#2#6}}
\crefrangeformat{subsubsection}{\mbox{\S\S#3#1#4--#5#2#6}}
\crefrangeformat{paragraph}{\mbox{\P\P#3#1#4--#5#2#6}}
\crefrangeformat{subparagraph}{\mbox{\P\P#3#1#4--#5#2#6}}
\crefname{part}{Part}{Parts}
\Crefname{part}{Part}{Parts}
\crefname{chapter}{ch.}{ch.}
\Crefname{chapter}{Ch.}{Ch.}
\crefname{footnote}{fn.}{fn.}
\Crefname{footnote}{Fn.}{Fn.}
\crefname{figure}{figure}{figures}
\crefname{subfigure}{figure}{figures}
\Crefname{subfigure}{Figure}{Figures}
\crefname{appsec}{appendix}{appendices}
\Crefname{appsec}{Appendix}{Appendices}
\crefname{algocf}{algorithm}{algorithms}
\Crefname{algocf}{Algorithm}{Algorithms}

\crefname{ExNo}{example}{examples}
\Crefname{ExNo}{Example}{Examples}
\crefname{SubExNo}{example}{examples}
\Crefname{SubExNo}{Example}{Examples}
\crefname{SubSubExNo}{example}{examples}
\Crefname{SubSubExNo}{Example}{Examples}
\crefformat{ExNo}{(#2#1#3)}
\crefformat{SubExNo}{(#2#1#3)}
\crefformat{SubSubExNo}{(#2#1#3)}
\crefrangeformat{ExNo}{\mbox{(#3#1#4--#5#2#6)}}
\crefrangeformat{SubExNo}{\mbox{(#3#1#4--#5#2#6)}}
\crefrangeformat{SubSubExNo}{\mbox{(#3#1#4--#5#2#6)}}
\crefmultiformat{ExNo}{(#2#1#3}{, #2#1#3)}{, #2#1#3}{, #2#1#3)}
\crefmultiformat{SubExNo}{(#2#1#3}{, #2#1#3)}{, #2#1#3}{, #2#1#3)}
\crefmultiformat{SubSubExNo}{(#2#1#3}{, #2#1#3)}{, #2#1#3}{, #2#1#3)}
\crefrangemultiformat{ExNo}{(#3#1#4--#5#2#6}{, #3#1#4--#5#2#6)}{, #3#1#4--#5#2#6}{, #3#1#4--#5#2#6)}
\crefrangemultiformat{SubExNo}{(#3#1#4--#5#2#6}{, #3#1#4--#5#2#6)}{, #3#1#4--#5#2#6}{, #3#1#4--#5#2#6)}
\crefrangemultiformat{SubSubExNo}{(#3#1#4--#5#2#6}{, #3#1#4--#5#2#6)}{, #3#1#4--#5#2#6}{, #3#1#4--#5#2#6)}

\newcommand{\gradinput}{G$\times$I}

\newcommand\norm[1]{\left\lVert#1\right\rVert}

\title{Explaining How Transformers Use Context to Build Predictions}

\author{Javier Ferrando$^1$, Gerard I. Gállego$^1$, Ioannis Tsiamas$^1$, Marta R. Costa-jussà$^2$\\
         $^1$TALP Research Center, Universitat Politècnica de Catalunya \\
         $^2$Meta AI \\
         \texttt{\{javier.ferrando.monsonis,gerard.ion.gallego,ioannis.tsiamas\}@upc.edu}\\
         \texttt{costajussa@meta.com}
         }

\begin{document}
\maketitle
\begin{abstract}
Language Generation Models produce words based on the previous context. Although existing methods offer input attributions as explanations for a model's prediction, it is still unclear how prior words affect the model's decision throughout the layers. In this work, we leverage recent advances in explainability of the Transformer and present a procedure to analyze models for language generation. Using contrastive examples, we compare the alignment of our explanations with evidence of the linguistic phenomena, and show that our method consistently aligns better than gradient-based and perturbation-based baselines. Then, we investigate the role of MLPs inside the Transformer and show that they learn features that help the model predict words that are grammatically acceptable. Lastly, we apply our method to Neural Machine Translation models, and demonstrate that they generate human-like source-target alignments for building predictions.
\end{abstract}

\section{Introduction}
Language Generation Models, like Transformer-based Language Models \citep{NEURIPS2020_1457c0d6, opt_lm} have recently revolutionized the field of Natural Language Processing (NLP). Despite this, there is still a gap in our understanding of how they are able to produce language that closely resembles that of humans. This means that we are unable to determine the cause of a model's failure in specific
instances, which can result in the generation of hallucinated content or toxic output.

\begin{table}[t]
\small
\centering
\begin{tabular}{l}\toprule
Logits Difference: \mybox{color700}{\strut{Increase}}\;\;\;\mybox{color420}{\strut{Decrease}}
\\
\midrule
Model Prediction: has (2.2\%), have (0.1\%)\\
Logits Difference: $\text{logit}_{\text{has}-\text{have}}= 3.1$\\
\midrule
L12\;| \mybox{color633}{\strut{A}} \mybox{color688}{\strut{report}} \mybox{color514}{\strut{about}} \mybox{color507}{\strut{the}} \mybox{color519}{\strut{Impressionists}} \mybox{color501}{\strut{\textbf{has}}}\\
\addlinespace
L11\;| \mybox{color846}{\strut{A}} \mybox{color1000}{\strut{report}} \mybox{color531}{\strut{about}} \mybox{color515}{\strut{the}} \mybox{color553}{\strut{Impressionists}} \mybox{color501}{\strut{\textbf{has}}}\\
\addlinespace
L10\;| \mybox{color528}{\strut{A}} \mybox{color546}{\strut{report}} \mybox{color533}{\strut{about}} \mybox{color509}{\strut{the}} \mybox{color499}{\strut{Impressionists}} \mybox{color501}{\strut{\textbf{has}}}\\
\addlinespace
L9\,\,\,\;| \mybox{color610}{\strut{A}} \mybox{color674}{\strut{report}} \mybox{color546}{\strut{about}} \mybox{color507}{\strut{the}} \mybox{color513}{\strut{Impressionists}} \mybox{color501}{\strut{\textbf{has}}}\\
\addlinespace
L8\,\,\,\;| \mybox{color584}{\strut{A}} \mybox{color592}{\strut{report}} \mybox{color589}{\strut{about}} \mybox{color524}{\strut{the}} \mybox{color498}{\strut{Impressionists}} \mybox{color501}{\strut{\textbf{has}}}\\
\addlinespace
L7\,\,\,\;| \mybox{color504}{\strut{A}} \mybox{color505}{\strut{report}} \mybox{color505}{\strut{about}} \mybox{color499}{\strut{the}} \mybox{color485}{\strut{Impressionists}} \mybox{color501}{\strut{\textbf{has}}}\\
\addlinespace
L6\,\,\,\;| \mybox{color501}{\strut{A}} \mybox{color500}{\strut{report}} \mybox{color501}{\strut{about}} \mybox{color496}{\strut{the}} \mybox{color471}{\strut{Impressionists}} \mybox{color501}{\strut{\textbf{has}}}\\
\addlinespace
L5\,\,\,\;| \mybox{color485}{\strut{A}} \mybox{color493}{\strut{report}} \mybox{color489}{\strut{about}} \mybox{color498}{\strut{the}} \mybox{color510}{\strut{Impressionists}} \mybox{color501}{\strut{\textbf{has}}}\\
\addlinespace
L4\,\,\,\;| \mybox{color502}{\strut{A}} \mybox{color494}{\strut{report}} \mybox{color484}{\strut{about}} \mybox{color493}{\strut{the}} \mybox{color492}{\strut{Impressionists}} \mybox{color501}{\strut{\textbf{has}}}\\
\addlinespace
L3\,\,\,\;| \mybox{color496}{\strut{A}} \mybox{color501}{\strut{report}} \mybox{color496}{\strut{about}} \mybox{color504}{\strut{the}} \mybox{color477}{\strut{Impressionists}} \mybox{color501}{\strut{\textbf{has}}}\\
\addlinespace
L2\,\,\,\;| \mybox{color524}{\strut{A}} \mybox{color495}{\strut{report}} \mybox{color489}{\strut{about}} \mybox{color493}{\strut{the}} \mybox{color464}{\strut{Impressionists}} \mybox{color501}{\strut{\textbf{has}}}\\
\addlinespace
L1\,\,\,\;| \mybox{color531}{\strut{A}} \mybox{color515}{\strut{report}} \mybox{color499}{\strut{about}} \mybox{color501}{\strut{the}} \mybox{color379}{\strut{Impressionists}} \mybox{color501}{\strut{\textbf{has}}}\\
\midrule
$\sum$ \;\,\,|  \mybox{color842}{\strut{A}} \mybox{color962}{\strut{report}} \mybox{color579}{\strut{about}} \mybox{color519}{\strut{the}} \mybox{color436}{\strut{Impressionists}} \mybox{color501}{\strut{\textbf{has}}}\\
\bottomrule
\end{tabular}
\caption{Updates to the (logits) prediction difference between \textbf{has} and \textbf{have} in different layers produced by input tokens. Red indicates an increase in the difference in logits between both predictions. At the bottom, we show the final logit contributions. The contrastive extension of our proposed method, ALTI-Logit, shows that the model relies on the head of the subject (report) to correctly solve the subject-verb agreement. See explanations from other methods in \Cref{table:distractor_15vsothers}. GPT-2 Small shown here, see GPT-2 XL ALTI-Logit explanation in \Cref{table:distractor_15_gpt2xl}.}
\label{tab:main}
\end{table}
The majority of previous work in explainability of NLP model predictions has focused on analyzing them on downstream tasks, generally with a small output space, such as text classification or Natural Language Inference \citep{atanasova-etal-2020-diagnostic,bastings-etal-2022-will,zaman2022multilingual}. This line of research includes a large body of work focusing on the analysis of the attention mechanism \citep{jain-wallace-2019-attention,serrano-smith-2019-attention,pruthi-etal-2020-learning}, and on applying gradient-based methods \citep{li-etal-2016-visualizing, pmlr-v70-sundararajan17a} to obtain input attribution scores.

Recently, several works have tackled the interpretability of Transformers \citep{NIPS2017_3f5ee243} on the Language Modeling task. \citet{elhage2021mathematical} studied the Transformer from the \textit{residual stream} perspective, depicted in \Cref{fig:transformer_diagram}, where different components (MLPs, attention heads...) read and write to subspaces of the residual stream. This approach has aided in explaining certain behaviours of language models, like induction heads \citep{olsson2022context}, where attention heads search over the context for previous repetitions of the same token and copy the next token, or even specialized heads solving the Indirect Object Identification (IOI) task \citep{interpretability_in_the_wild}. Similarly, MLPs inside the Transformer have also been studied as elements writing into the residual stream. \citet{geva-etal-2022-transformer} observed that MLP blocks can act as key-value memories, where values add to the residual, thus promoting the prediction of words that convey similar semantic meaning. 

Furthermore, the \textit{attention mechanism} in the Transformer, composed of attention heads, an output weight matrix, and a layer normalization, can be decomposed into an interpretable operation \citep{kobayashi-etal-2020-attention,kobayashi-etal-2021-incorporating}, providing layer-wise explanations which have proven to be highly faithful \citep{ferrando2022measuring, alti_plus}.

In this work, we propose explaining the predictions of Transformers language generators by combining the residual stream analysis perspective with the attention decomposition. Our approach measures the amount of logit (pre-activation of the softmax) added or subtracted by each token representation at each layer. We then track the logit contributions back to the model's input by aggregating across layers (\textit{Logit} explanation). Additionally, we consider the mixing of information in intermediate layers by using ALTI \citep{ferrando2022measuring} (\textit{ALTI-Logit} explanation).

To evaluate the proposed interpretability methods, we follow the recently introduced contrastive explanations framework \citep{kayo_interpreting_lms}, which aims to explain why the model predicted one token instead of a foil token, \textit{a priori} explained by some linguistic phenomena evidence. Then, we analyze the role of MLPs and show that they aid the model in determining predictions that follow grammar rules. Finally, we demonstrate that NMT models generate human-like source-target alignments for building translations.\footnote{The code accompanying the paper is available at\\ \url{https://github.com/mt-upc/logit-explanations}.}
\section{Approach}
\subsection{Residual Stream}
Given a language generation timestep $t$, the output of the last layer,\footnote{We refer to it as a row vector.} $\vx^{L}_t \in \mathbb{R}^{d}$, is projected to the token embedding space by applying the unembedding matrix $\mU \in \mathbb{R}^{d \times |V|}$ to get the logits of the next token prediction. Then, a softmax function is applied to obtain a probability distribution over the vocabulary:
\begin{equation}
P(\vx^{L}_{t}) = \text{softmax}(\vx^{L}_{t} \mU)
\end{equation}
\begin{figure}[t]\begin{centering}\includegraphics[width=0.23\textwidth]{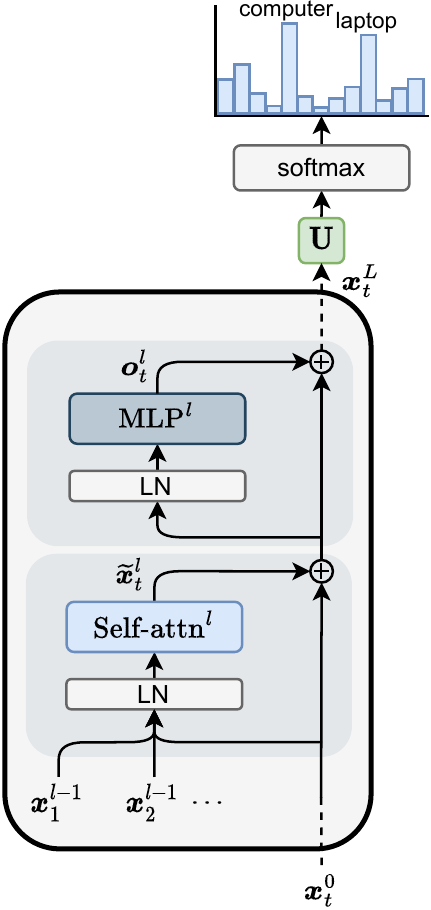}
	\caption{A Transformer Language Model, represented as modules writing into the residual stream.}
	\label{fig:transformer_diagram}
	\end{centering}
\end{figure}
The residual connection in the Transformer can be seen as an information stream \citep{nostalgebraist,elhage2021mathematical, 10.1162/tacl_a_00501} that gets updated after each block. Let's call $\vo^l_t$ and $\widetilde{\vx}_{t}^{l}$ the output of the MLP and self-attention blocks at layer $l$ respectively, `writing' into the residual stream at position $t$ (\Cref{fig:transformer_diagram}). The last state of the residual stream can be represented as
\begin{equation}
\vx^{L}_t = \sum_{l}^{L}\vo^l_t + \sum_{l}^{L} \widetilde{\vx}_{t}^{l} + \vx^{0}_t
\end{equation}
The final logit of a particular next token prediction $w$ can be computed by multiplying the last state of the residual stream with the $w$-th column\footnote{Note that we refer to the $j$-th column of a matrix $\mB$ as $\mB_{j}$, instead of $\mB_{:,j}$.} of $\mU$:
\begin{equation}
\begin{aligned}
\text{logit}_w&= \vx^{L}_t\mU_{w}\\
&= \Big(\sum_{l}^{L}\vo^l_t + \sum_{l}^{L} \widetilde{\vx}_{t}^{l} + \vx^{0}_t\Big)\mU_{w}
\end{aligned}
\end{equation}
By linearity:
\begin{equation}\label{eq:logits_w}
    \text{logit}_w = \sum_{l}^{L}\vo^l_t\mU_{w} + \sum_{l}^{L} \widetilde{\vx}_{t}^{l}\mU_{w} + \vx^{0}_t\mU_{w}
\end{equation}

\begin{figure}[t]
	\begin{centering}
	\includegraphics[width=0.48\textwidth]{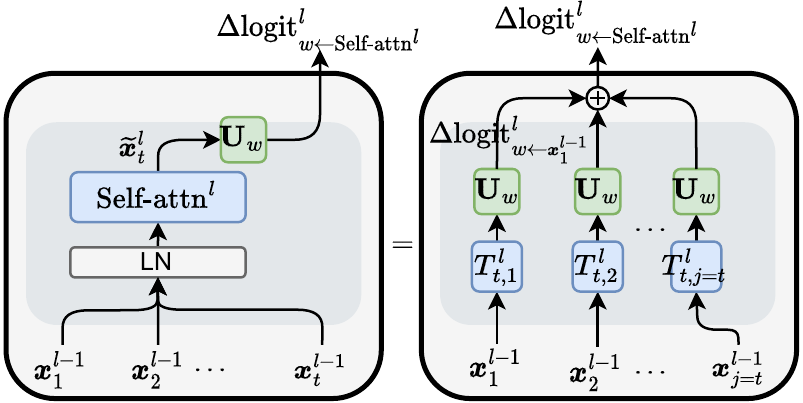}
	\caption{The output of the self-attention block at each layer updates the logit of $w$ (left). The logit's update can be decomposed per input token (right).}
	\label{fig:attention_embed_space_proj}
	\end{centering}
\end{figure}

\subsection{Multi-head Attention as a Sum of Vectors}
Inspired by the decomposition of the Post-LN self-attention block done by \citet{kobayashi-etal-2021-incorporating}, we apply a similar approach to the Pre-LN setting, common in current LMs (see full derivation in \Cref{appx:self_attn_decomp}). The output of the self-attention block at each generation step $t$ can be expressed  as
\begin{equation}\label{eq:post_layer_transformed_vectors}
\widetilde{\vx}_{t}^{l} = \sum_{j}^{t} T^l_{t,j}(\vx^{l-1}_j) + \vb_O^l
\end{equation}
where $T^l_{t,j}: \mathbb{R}^{d}\mapsto\mathbb{R}^{d}$ is an affine transformation applied to each layer's input token representation (or residual stream) $\vx^{l-1}_{j} \in \mathbb{R}^{d}$:
\begin{equation}\label{eq:transformed_vectors}
\resizebox{0.48\textwidth}{!}{$\displaystyle{
T^l_{t,j}(\bm{x}^{l-1}_j) = \sum_{h}^{H}\biggl( \vx^{l-1}_j\mL^l\mW^{l,h}_{V}\mA_{t,j}^{l,h}\mW_{O}^{l,h} + \mA_{t,j}^{l,h}\theta^{l,h}\biggl)
}$}
\end{equation}
with $\mW_V^{l,h} \in \mathbb{R}^{d \times d_h}$ the matrix generating the values, $\mW_O^{l,h} \in \mathbb{R}^{d_h \times d}$ the attention output matrix (per head) and $\vb^{l}_{O} \in \mathbb{R}^{d}$ its associated bias. $\mA^{l,h} \in \mathbb{R}^{t \times t}$ is the attention weight matrix of each head, $\theta^{l,h} \in \mathbb{R}^{d}$ remaining terms originated from biases, and $\mL^l \in \mathbb{R}^{d \times d}$ combines centering, normalizing, and scaling operations of the layer normalization (see \Cref{appx:self_attn_decomp}).
\subsection{Layer-wise Contributions to the Logits}
Combining \Cref{eq:logits_w} and \Cref{eq:post_layer_transformed_vectors} we get\footnote{Biases are removed to save space.}:
\begin{equation}\label{eq:decomposition_final_logits}
\resizebox{0.48\textwidth}{!}{$\displaystyle{
    \text{logit}_w = \sum_{l}^{L}\underbrace{\vphantom{\sum_{j}^{L}T^l}\vo^l_t\mU_{w}}_{\mathclap{\Delta \text{logit}^l_{w \gets \text{MLP}^{l}}}} + \sum_{l}^{L}\underbrace{\sum_{j}^{t} T_{t,j}^l(\vx^{l-1}_j)\mU_{w}}_{\Delta \text{logit}^l_{w \gets \text{Self-attn}^{l}}} + \vx^{0}_t\mU_{w}
}$}
\end{equation}

The logit's update of each self-attention, $\Delta \text{logit}^l_{w \gets \text{Self-attn}^{l}}$, can be expanded into individual updates by each $\vx^{l-1}_j$ (\Cref{fig:attention_embed_space_proj}). Therefore, the contribution of each layer's input token representation $\vx^{l-1}_j$ to an output token $w$ can be defined as its update to the logit of $w$:
\begin{equation}\label{eq:indiv_contrib}
\Delta \text{logit}^{l}_{w\gets{\boldsymbol{x}^{l-1}_j}} = T_{t,j}^l(\vx^{l-1}_j) \mU_{w}
\end{equation}

Similarly, logit updates can be computed at the head level ($\Delta 
 \text{logit}^{l,h}_{w\gets{\boldsymbol{x}^{l-1}_j}}$) by multiplying the unembedding matrix with the head-wise affine transformation in \Cref{eq:transformed_vectors}.

\begin{figure}[t]\begin{centering}\includegraphics[width=0.3\textwidth]{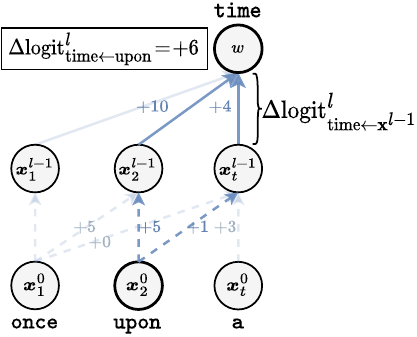}
	\caption{$\vx_2^{l-1}$ and $\vx_3^{l-1}$ contribute 10 and 4 logits respectively to the next token prediction $w=\texttt{time}$. Due to the mixing of contextual information across layers, \texttt{upon} contributes $\frac{1}{2}$ to $\vx_2^{l-1}$ and $\frac{1}{4}$ to $\vx_3^{l-1}$, which results in \texttt{upon} contributing $10\cdot\frac{1}{2} + 4\cdot\frac{1}{4} = 5 + 1 = +6$ logits.}
	\label{fig:logits_input_example}
	\end{centering}
\end{figure}

\subsection{Tracking Logit Updates to the Input Tokens}

If we assume each residual stream preserves its token identity throughout the layers, the total logit update to $w$ produced by input token $s$ can be computed as
\begin{equation}\label{eq:input_model_contrib_no_mix}
\Delta \text{logit}_{w \gets s} = \sum_{l}^{L} \Delta \text{logit}^l_{w\gets{\boldsymbol{x}^{l-1}_{j=s}}}
\end{equation}
that is, the sum of the logit updates performed by the $s$-th token intermediate representations at every layer. Henceforth, we refer to this as the \textit{Logit} explanation.

However, in intermediate layers, each residual stream represents a mixture of input tokens \citep{Brunner2020On}. Therefore, $\Delta \text{logit}^{l}_{w\gets{\boldsymbol{x}^{l-1}_j}}$ can't be directly interpreted as the logit update \textit{caused by} the model's input token $s=j$. We propose to track the logit update back to the model inputs by measuring the mixing of contextual information in the residual streams. For that purpose, we use ALTI \citep{ferrando2022measuring}. ALTI, as well as other methods relying on the \textit{rollout} method \citep{abnar-zuidema-2020-quantifying, mohebbi-etal-2023-quantifying} assume that token representations are formed by linearly combining the representations from the preceding layer, i.e. ${\vx_i^{l} = \sum_j c^{l}_{i,j} \vx^{l-1}_j}$, with ${\sum_j c^{l}_{i,j} = 1}$. Each $c^{l}_{i,j}$ refers to the contribution of $\vx^{l-1}_j$ to $\vx^{l}_i$. By multiplying the layer-wise coefficient matrices, ${\mM^l = \mC^l \cdot \mC^2 \cdots \mC^1}$, one can describe each intermediate layer representation as a linear combination of the model input tokens, ${\vx_i^{l} = \sum_s m^{l}_{i,s} \vx^{0}_s}$.

Column $s$ of $\mM^{l-1}$ contains the proportion of the $s$-th input token’s contribution encoded in each token representation \textit{entering} layer $l$. We can obtain the update performed by each model input token (\Cref{fig:logits_input_example}, right) to the logit of a next prediction token $w$ as
\begin{equation}\label{eq:input_model_contrib_layer}
 \Delta \text{logit}^l_{w \gets s} = \Delta \text{logit}^l_{w\gets{\mathbf{x}^{l-1}}}\; \mM^{l-1}_{s}
\end{equation}

We refer to \Cref{apx:track_logits_rollout} for a more detailed explanation. The final contribution of the $s$-th input token to the prediction of token $w$ can be obtained as the sum of its logit updates at each layer:
\begin{equation}\label{eq:input_model_contrib}
\Delta \text{logit}_{w \gets s} = \sum_{l}^{L} \Delta \text{logit}^l_{w \gets s}
\end{equation}
We denote this method the \textit{ALTI-Logit} explanation. Note that if we don't consider mixing of contextual information, $\mM^{l-1}$ becomes the identity matrix, and we get the Logit explanation (\Cref{eq:input_model_contrib_no_mix}).
\subsection{Contrastive Explanations}
Contrastive explanations \citep{kayo_interpreting_lms} aim to explain why the model predicted one target token $w$ instead of another foil token $f$. We can explain this decision by determining how much each token contributed to the final logit difference between $w$ and $f$: $\text{logit}_{(w-f)}$. Following \Cref{eq:input_model_contrib_no_mix} and \Cref{eq:input_model_contrib}, we can define the Contrastive Logit and Contrastive ALTI-Logit \footnote{Throughout the paper we use Logit and ALTI-Logit to refer also to their contrastive variant.} saliency scores of input tokens as their update to the logit difference:
\begin{equation}\label{eq:contrastive_logit_saliency}
\Delta \text{logit}_{(w - f) \gets s} = \Delta \text{logit}_{w \gets s} - \Delta \text{logit}_{f \gets s}
\end{equation}

\section{Experimental Setup}

We evaluate the quality of our proposed method through contrastive explanations. Following \citet{kayo_interpreting_lms} we use a subset of BLiMP dataset \citep{warstadt-etal-2020-blimp-benchmark}, which contains sentence pairs with small variations in grammatical correctness. The 11 subsets belong to 5 linguistic phenomena: anaphor agreement, argument structure, determiner-noun agreement, NPI licensing, and subject-verb agreement.

\begin{table}[t]
\centering
\setlength{\tabcolsep}{2.3pt}
\resizebox{\linewidth}{!}{ 
\begin{tabular}{ccl}
\toprule
\textbf{Phenomena} & \textbf{ID}  & \textbf{Example} (\textcolor{custom_green}{Acceptable}/\textcolor{custom_red}{Unacceptable})\\ 
\midrule
\multirow{2}{*}{Anaphor Agreement} & aga & \ul{Karla} could listen to \textcolor{custom_green}{herself}/\textcolor{custom_red}{himself}.\\
& ana & \ul{Eva} approached \textcolor{custom_green}{herself}/\textcolor{custom_red}{themselves}.\\
\midrule
Argument Structure & asp & Gerald is \ul{hated} by the \textcolor{custom_green}{teachers}/\textcolor{custom_red}{pie}.\\
\midrule
\multirow{4}{*}{Determiner-Noun Agreement} & dna & Eva has scared \ul{these} \textcolor{custom_green}{children}/\textcolor{custom_red}{child}.\\
& dnai & Tammy was observing \ul{that} \textcolor{custom_green}{man}/\textcolor{custom_red}{men}.\\
& dnaa & The driver sees \ul{that} unlucky \textcolor{custom_green}{person}/\textcolor{custom_red}{people}.\\
& dnaai & Phillip liked \ul{that} smooth \textcolor{custom_green}{horse}/\textcolor{custom_red}{horses}.\\
\midrule
NPI Licensing & npi & \ul{Even} Danielle \textcolor{custom_green}{also}/\textcolor{custom_red}{ever} leaves.\\
\midrule
\multirow{3}{*}{Subject-Verb Agreement} & darn & The \ul{grandfathers} of Diana \textcolor{custom_green}{drink}/\textcolor{custom_red}{drinks}.\\
& ipsv & Many \ul{people} \textcolor{custom_green}{have}/\textcolor{custom_red}{has} hidden away.\\
& rpsv & Most \ul{associations} \textcolor{custom_green}{buy}/\textcolor{custom_red}{buys} those libraries.\\
\bottomrule
\end{tabular}}
\caption{Examples: in \Cref{tab:gpt2xl_mrr} of BLiMP phenomenons\protect\footnotemark\ used by \citet{kayo_interpreting_lms}, with acceptable and unacceptable continuations in bold. Underlined words represent the linguistic evidence to resolve the phenomena (extracted by the rules).} 
\label{tab:blimp_examples}
\end{table}
\footnotetext{BLiMP IDs. aga: anaphor\_gender\_agreement; ana: anaphor\_number\_agreement; asp: animate\_subject\_passive; dna: determiner\_noun\_agreement\_1; dnai: determiner\_noun\_agreement\_irregular\_1; dnaa: determiner\_noun\_agreement\_with\_adj\_1; dnaai: determiner\_noun\_agreement\_with\_adj\_irregular\_1; npi: npi\_present\_1; darn: distractor\_agreement\_relational\_noun; ipsv: irregular\_plural\_subject\_verb\_agreement\_1; rpsv: regular\_plural\_subject\_verb\_agreement\_1}

For each linguistic phenomena, we use spaCy \citep{spacy2} and follow \citet{kayo_interpreting_lms} rules to find the evidence (in previous tokens), that is enforcing grammatical acceptability (\Cref{tab:blimp_examples}). For anaphor agreement, we obtain all context tokens that are coreferent with the target token. For argument structure, we extract the main verb of the sentence. Determiner-noun agreement's evidence is found in the determiner of the target noun. In NPI licensing, "even" word can appear in the acceptable target, but not in the unacceptable. Finally, in the subject-verb agreement phenomenon, the form of the verb has to agree in number with the head of the subject, which we use as evidence. We differ from \citet{kayo_interpreting_lms} in that we discard \texttt{ipsv} and \texttt{rpsv} subsets, due to the large fraction of sentences with a `quantifier + head of subject + verb' structure, where both the quantifier (many, most...) and the head of the subject could be used by the model to solve the agreement.

We also add to the analysis \texttt{SVA} (subject-verb agreement) \citep{linzen2016assessing} and the Indirect Object Identification (\texttt{IOI}) \citep{interpretability_in_the_wild,fahamu_2022} datasets. The \texttt{SVA} dataset includes nouns with an opposite number to that of the main subject, which makes this dataset well-suited for evaluating saliency methods. Indirect object identification (\texttt{IOI}) is a feature present in sentences that have an initial dependent clause, like "After Lee and Evelyn went to the lake", followed by a main clause, like "Lee gave a grape to Evelyn". The indirect object "Evelyn" and the subject "Lee" are found in the initial clause. In all examples of \texttt{IOI} dataset, the main clause refers to the subject again, which gives an object to the IO. The goal of the \texttt{IOI} task is to predict the final word in the sentence to be the IO. In \texttt{IOI} examples, the rule for predicting the IO is the IO itself being in the first clause.

We use GPT-2 XL (1.5B) model \citep{Radford2019LanguageMA}, as in \citep{kayo_interpreting_lms}, as well as other autoregressive Transformer language models, such as GPT-2 Small (124M), and GPT-2 Large models (774M), OPT 125M \citep{opt}, and BLOOM's 560M and 1.1B variants \citep{bloom}, through HuggingFace library \cite{wolf-etal-2020-transformers}.

\paragraph{Alignment Metrics.}
Following \citet{kayo_interpreting_lms}, we define the \textit{evidence} as a binary vector $\vb \in \mathbb{R}^t$ (with as many components as the number of previous tokens), with all zeros except in the position of the tokens inside the evidence, i.e. the tokens which the prediction depends on, extracted by the rule. Explanations are vectors, also $\in \mathbb{R}^t$. To measure the alignment between an explanation and the evidence we use MRR (Mean Reciprocal Analysis). Sorting the tokens in descending order, MRR evaluates the average of the inverse of the rank of the first token that is part of $\vb$.
Although \citet{kayo_interpreting_lms} use also dot-product and Probes Needed metrics for measuring alignments, dot-product favors Grad Norm explanations since it gives positive scores only, and Probes Needed is closely related to MRR, giving redundant results.

\section{Contrastive Methods}\label{sec:contrastive_methods}
\citet{kayo_interpreting_lms} proposed extending different common input attribution methods to the contrastive setting. In \Cref{sec:results} we compare their explanations with the ones obtained with our proposed contrastive methods (\Cref{eq:contrastive_logit_saliency}).

\subsection{Input Erasure}
\label{sec:erasure}
Erasure-based methods remove parts of the input and measure the change in the model's prediction \cite{li2016understanding}, where the higher the prediction change, the higher the attribution of that particular token. Specifically, we take the difference between the model's output with the entire input $\mathbf{x}$, and after removing from $\mathbf{x}$ the $s$-th token, i.e. $m_w(\mathbf{x}) - m_w{(\mathbf{x}_{\neg s})}$. \citet{kayo_interpreting_lms} define the Contrastive Input Erasure as
\begin{equation}
\resizebox{0.48\textwidth}{!}{$\displaystyle{
\vc^e_{(w, \neg f) \gets s} = \left(m_w(\mathbf{x}) - m_w(\mathbf{x}_{\neg s})\right) - \left(m_f(\mathbf{x}) - m_f(\mathbf{x}_{\neg s})\right)
}$}
\end{equation}
This metric evaluates the extent to which removing $x_s$ from the input increases the likelihood of the foil, and decreases the likelihood of the target in the model's output.

\subsection{Gradient Norm}\label{sec:norm}
The Transformer model can be approximated by the linear part of the Taylor-expansion at a baseline point \cite{simonyan_grads}, ${m(\mX^{0}) \approx \nabla m(\mX^{0}) \cdot \mX^{0}}$, where $\mX^0 \in \mathbb{R}^{t \times d}$ is the sequence of input embeddings. Therefore, $\nabla m_w(\mX^{0})$ represents the sensitivity of the model to each input dimension when predicting $w$. Following, saliency scores for each token can be computed by taking the norm of the gradient vector corresponding to the token embedding, 
$\norm{\nabla_{\vx_{s}^{0}}m(\mX^{0})}_1$.

\citet{kayo_interpreting_lms} extend this method to the Contrastive Gradient Norm and define it as
\begin{equation}
\vc^g_{(w, \neg f) \gets s} = \norm{\nabla_{\vx_{s}^{0}} \left(m_w(\mX^{0}) - m_f(\mX^{0}) \right)}_1
\end{equation}

\begin{figure}[t]\begin{centering}\includegraphics[width=0.49\textwidth]{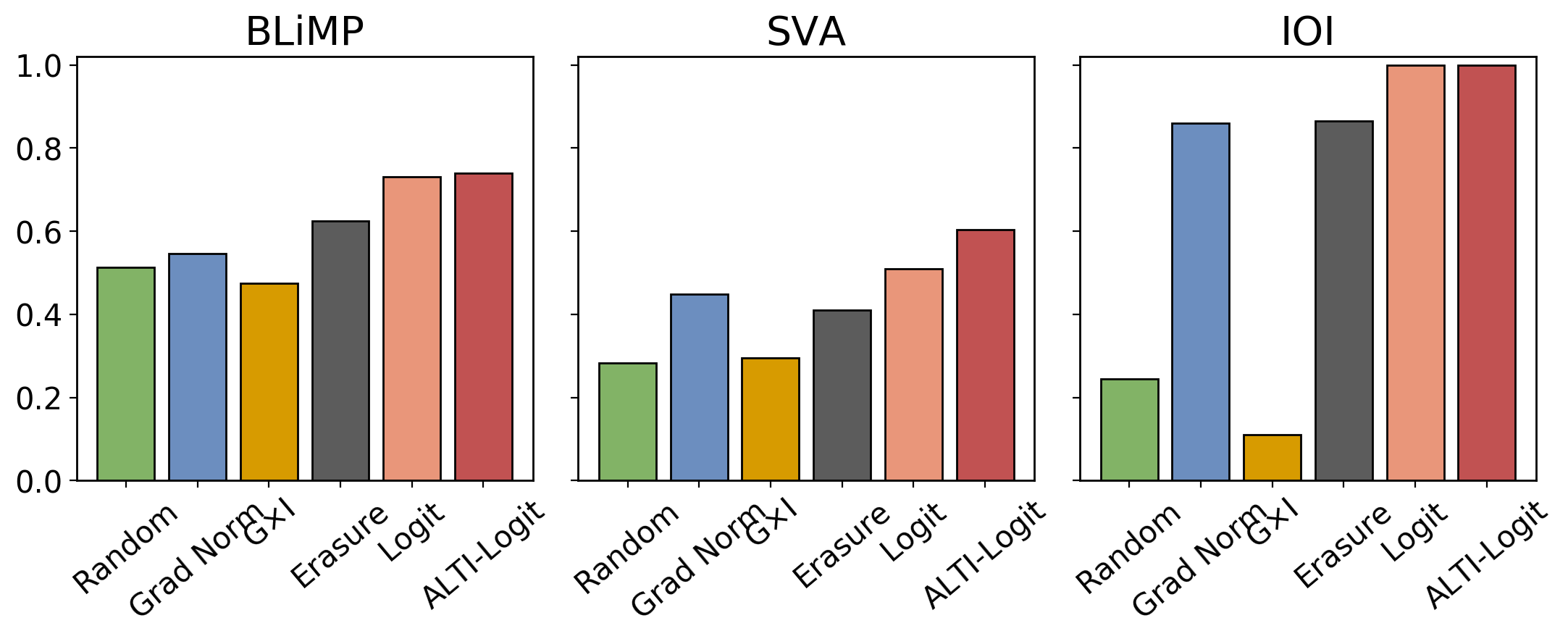}
	\caption{Alignment (MRR $\uparrow$) of different explanation methods of GPT-2 Small model predictions with BLiMP, \texttt{SVA}, and \texttt{IOI} datasets.}
	\label{fig:mrr_gpt2xl}
	\end{centering}
\end{figure}
\subsection{Gradient $\times$ Input}
\label{sec:input}
The gradient $\times$ input method \cite{grad_input2, grad_input} calculates the dot product between the gradient and the input token embedding. \citet{kayo_interpreting_lms} define the Contrastive Gradient $\times$ Input as
\begin{equation}
\vc^{g \times i}_{(w, \neg f) \gets s} = \nabla_{\vx_{s}^{0}} \left( m_w(\mX^{0}) - m_f(\mX^{0}) \right) \cdot \vx^0_s
\end{equation}

\begin{table}[!t]
    \footnotesize
    \centering
    \resizebox{0.272\textwidth}{!}{%
    \begin{tabular}{l} \toprule
    \multicolumn{1}{l}{Logit}\\ 
     \mybox{color517}{\strut{A}} \mybox{color1000}{\strut{report}} \mybox{color651}{\strut{about}} \mybox{color568}{\strut{the}} \mybox{color394}{\strut{Impressionists}} \mybox{color501}{\strut{\textbf{has}}}\\
    \addlinespace
    \multicolumn{1}{l}{ALTI-Logit}\\
  \mybox{color842}{\strut{A}} \mybox{color962}{\strut{report}} \mybox{color579}{\strut{about}} \mybox{color519}{\strut{the}} \mybox{color436}{\strut{Impressionists}} \mybox{color501}{\strut{\textbf{has}}}\\
    \addlinespace
    \multicolumn{1}{l}{Erasure}\\ 
     \mybox{color627}{\strut{A}} \mybox{color623}{\strut{report}} \mybox{color1000}{\strut{about}} \mybox{color485}{\strut{the}} \mybox{color593}{\strut{Impressionists}} \mybox{color501}{\strut{\textbf{has}}}\\
    \addlinespace
    \multicolumn{1}{l}{Grad Norm}\\ 
     \mybox{color717}{\strut{A}} \mybox{color674}{\strut{report}} \mybox{color587}{\strut{about}} \mybox{color562}{\strut{the}} \mybox{color963}{\strut{Impressionists}} \mybox{color501}{\strut{\textbf{has}}}\\
    \addlinespace
    \multicolumn{1}{l}{\gradinput}\\ 
     \mybox{color741}{\strut{A}} \mybox{color319}{\strut{report}} \mybox{color536}{\strut{about}} \mybox{color461}{\strut{the}} \mybox{color144}{\strut{Impressionists}} \mybox{color501}{\strut{\textbf{has}}}\\
    \bottomrule
    \end{tabular}}
    \caption{Comparison of different contrastive explanation methods described in \Cref{sec:contrastive_methods} and ALTI-Logit (\textbf{has} vs. \textbf{have}). Same example as in \Cref{tab:main}.}
    \label{table:distractor_15vsothers}
\end{table}

\section{Results}\label{sec:results}
In the following sections we provide results on the alignment between the explanations of different methods and linguistic evidence, as well as an analysis of observed model behaviours through the lens of ALTI-Logit.
\subsection{Alignment Results}
In \Cref{fig:mrr_gpt2xl} we present the MRR results of GPT-2 Small averaged across dataset categories, while the extended results for every subset can be found at \Cref{sec:apx_results}, \Cref{tab:gpt2small_mrr}. In \Cref{sec:apx_results}, \Cref{fig:mrr_all} we expand \Cref{fig:mrr_gpt2xl} across different models. We can observe that Logit and ALTI-Logit explanations consistently align better with the evidence of linguistic phenomena than common gradient-based and erasure-based baselines. Note that for BLiMP the average we show in \Cref{fig:mrr_gpt2xl} is across 9 different subsets. In \Cref{table:distractor_15vsothers} we show an example comparing different contrastive explanations, where Grad Norm, \gradinput\, and Erasure explanations don't align with the evidence to solve the subject-verb agreement (report), and disagree between each other.

We find similar alignment results for Logit and ALTI-Logit methods. However, we observe that ALTI-Logit aligns better at tasks where the tokens of the linguistic evidence are far from the prediction. This is especially noticeable in Subject-verb agreement datasets (including \texttt{SVA} and \texttt{darn}), where ALTI-Logit shows higher alignments than any other method across all models. This might indicate that incorporating information about contextual mixing is advantageous for dealing with large contexts.

\begin{figure}[!t]
	\begin{centering}
	\includegraphics[width=0.495\textwidth]{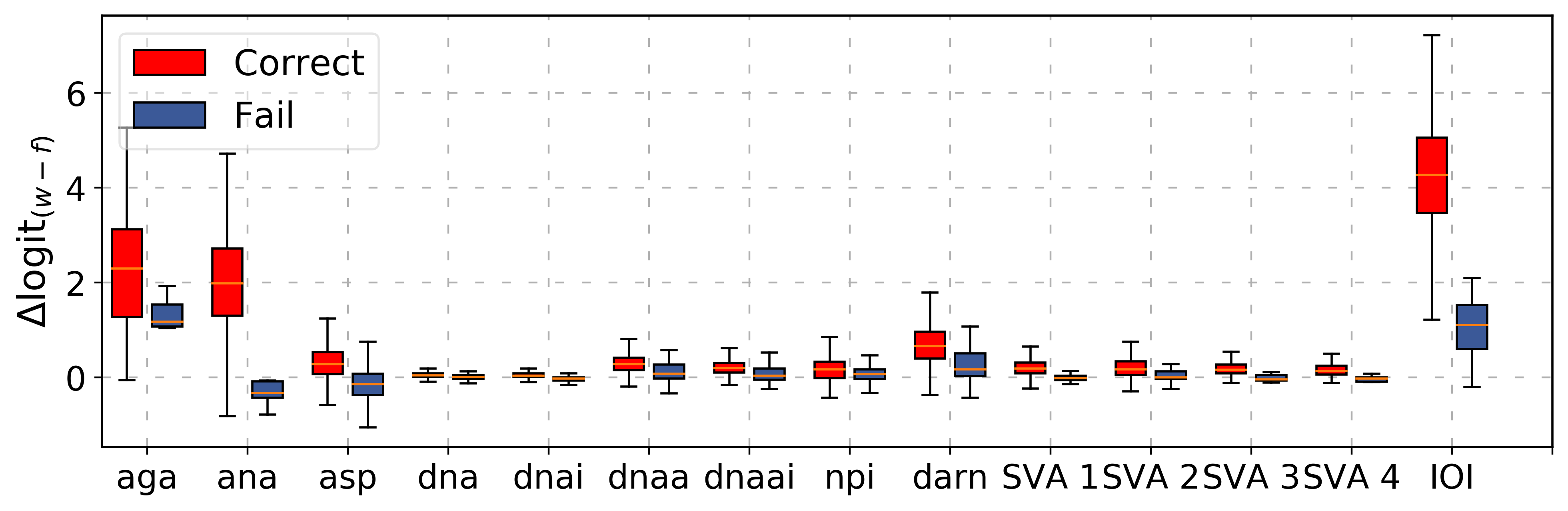}
	\caption{Update to the logit difference between the acceptable and the unacceptable predictions produced by the input tokens inside the linguistic evidence (GPT-2 XL).}
	\label{fig:logit_attn_full_diff_corrects_fails}
	\end{centering}
\end{figure}

\begin{figure}[!t]
\centering     
\subfigure{\label{fig:update_logit_attn_small_c}\includegraphics[width=0.49\textwidth]{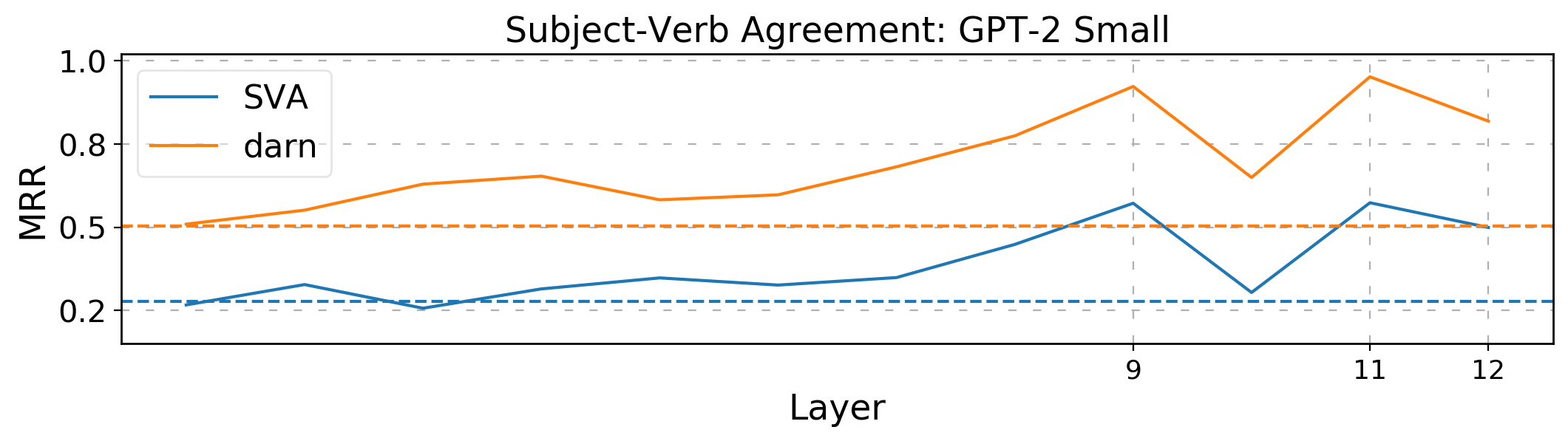}}
\subfigure{\label{fig:update_logit_attn_small_darn}\includegraphics[width=0.49\textwidth]{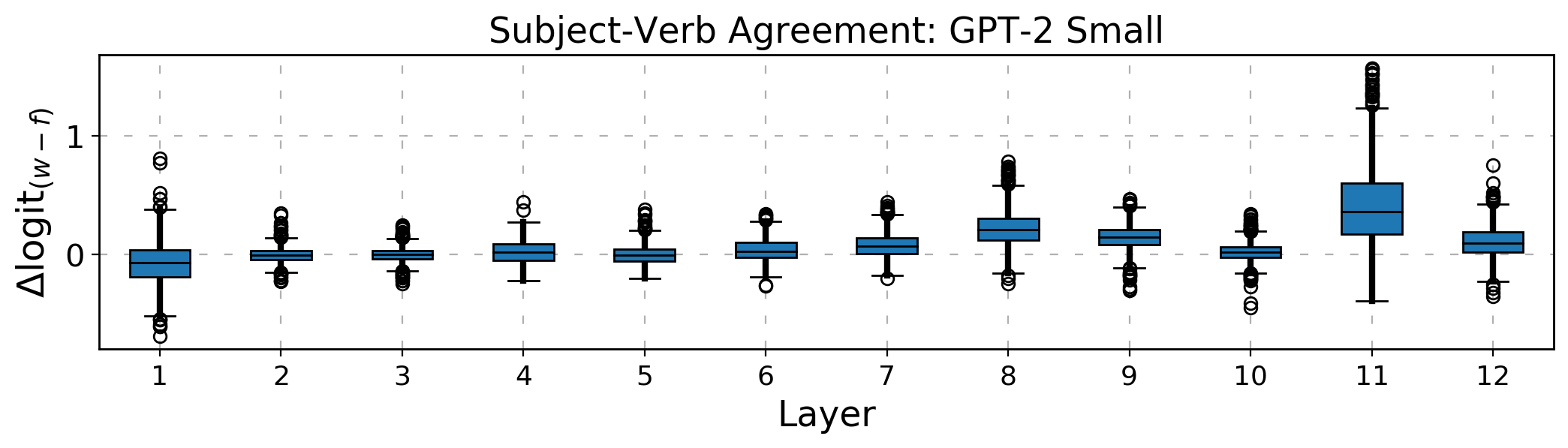}}
\subfigure{\label{fig:update_logit_attn_small_e}\includegraphics[width=0.49\textwidth]{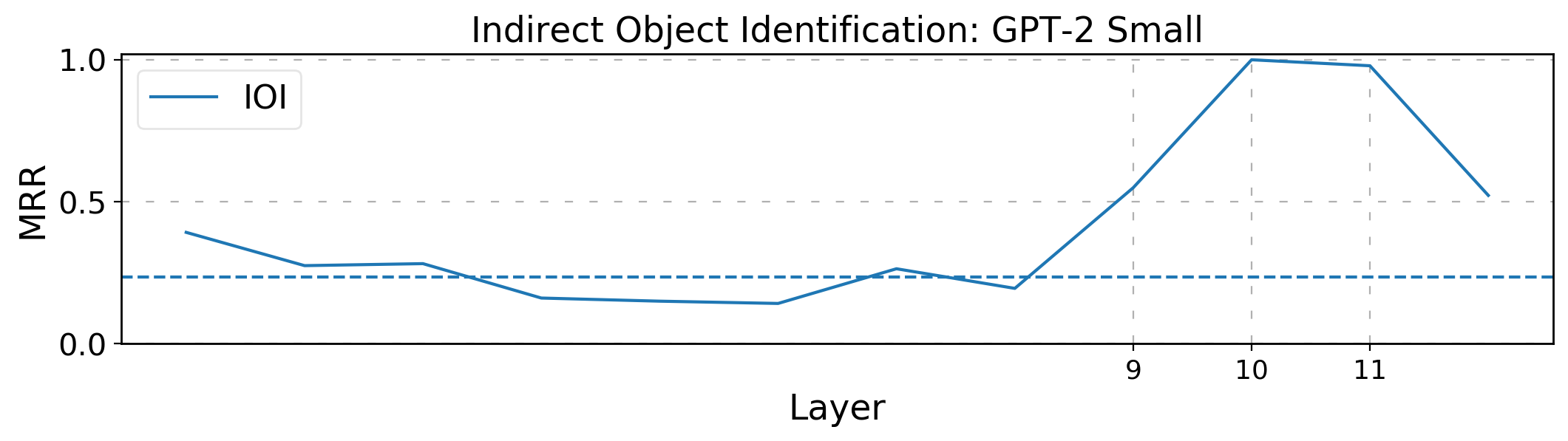}}
\subfigure{\label{fig:update_logit_attn_small_ioi}\includegraphics[width=0.49\textwidth]{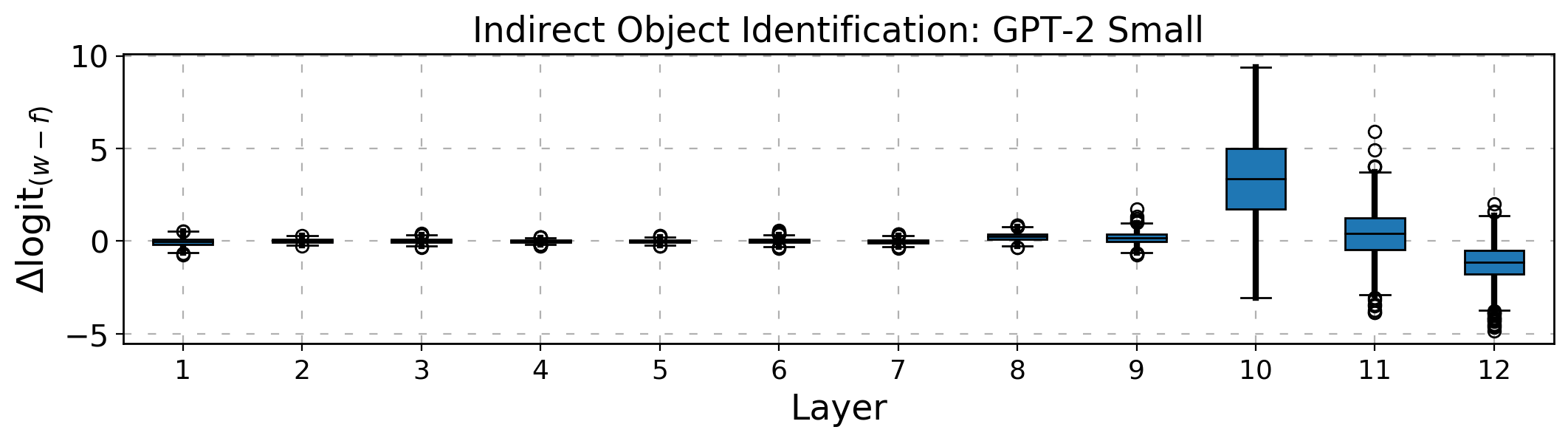}}

\caption{ALTI-Logit MRR alignment scores (line plots) and updates in logit difference by every input token ($\Delta \text{logit}^l_{(w-f)\gets \text{Self-attn}^l}$) between acceptable and unacceptable predictions (box plots) per layer (GPT-2 Small). Horizontal dashed lines refer to random alignment.}
\label{fig:update_logit_attn_small}
\end{figure}

Despite the generally accurate performance of the models examined in this study (\Cref{fig:logit_difference_gpt2_small_blimp} and \Cref{fig:logit_difference_gpt2_xl_blimp}, \Cref{apx:model_predictions}), there are cases where the unacceptable token gets predicted with a higher probability. In order to gain a deeper understanding of the variations in model behavior between correct and incorrect predictions, we analyze the logit update generated by the input tokens associated with the linguistic evidence. This analysis, conducted using ALTI-Logit (\Cref{fig:logit_attn_full_diff_corrects_fails}), reveals differences in the distributions. These findings suggest that the tokens representing the linguistic evidence play a crucial role in achieving accurate predictions, and if their contribution is only marginal, the likelihood of failure increases considerably.

\begin{figure}[!t]
\centering     
\subfigure{\label{fig:update_logit_attn_small_d}\includegraphics[width=0.49\textwidth]{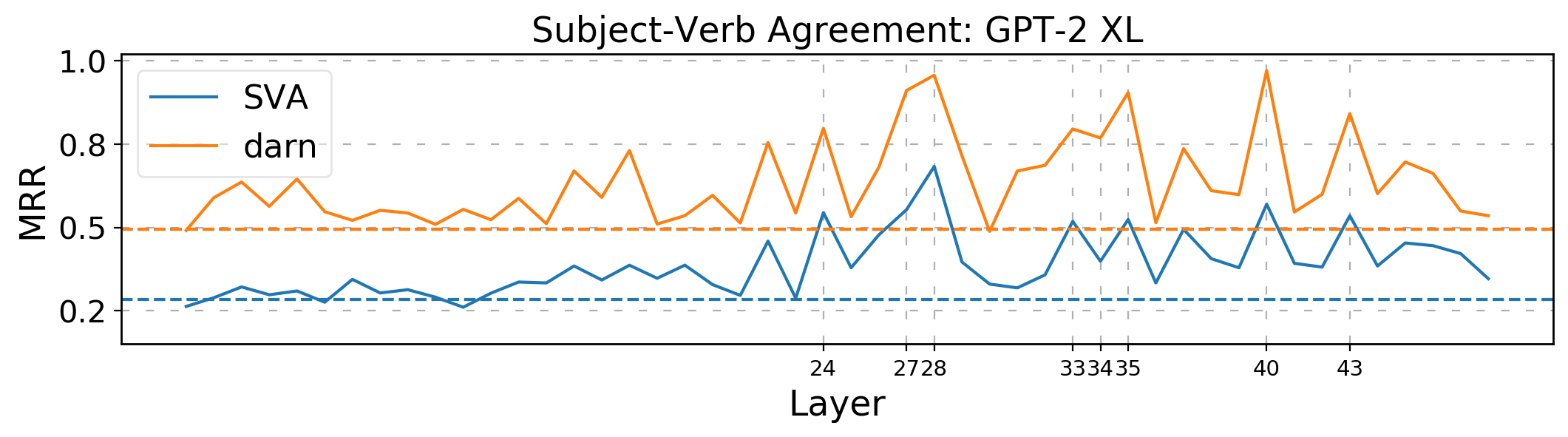}}
\subfigure{\label{fig:update_logit_attn_small_f}\includegraphics[width=0.49\textwidth]{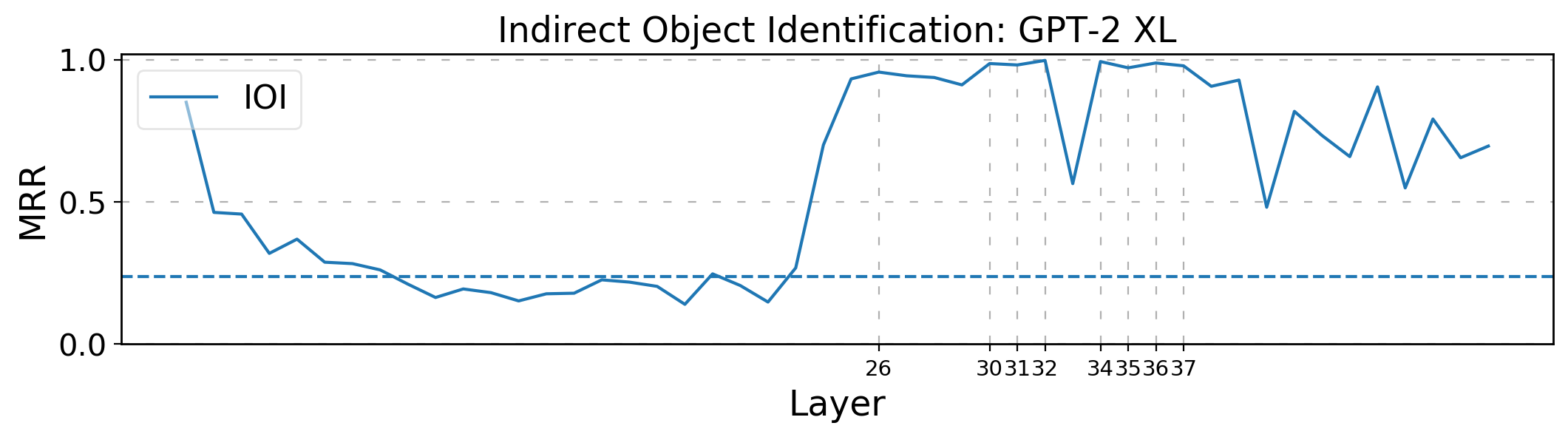}}
\caption{ALTI-Logit MRR alignment scores across layers (GPT-2 XL). Horizontal dashed lines refer to random alignment.}
\label{fig:update_logit_attn_xl}
\end{figure}

\subsection{Layer-wise Analysis with ALTI-Logit}
In the line plots in \Cref{fig:update_logit_attn_small,fig:update_logit_attn_xl} we provide the MRR alignment results across layers of GPT2-Small and GPT2-XL for two different linguistic phenomena. Models behave similarly across subsets inside the same phenomena, like in Subject-Verb Agreement (\texttt{SVA} and \texttt{darn}), and Anaphor Agreement (\texttt{aga} and \texttt{ana}) in \Cref{apx:alignment_gpt2_datasets}. The model’s alignment trend also stays similar, even though the distance between the prediction and the evidence is different across subsets (\texttt{SVA}'s distance is 4 times \texttt{darn}'s).

In the boxplots in \Cref{fig:update_logit_attn_small}, we show the distribution of self-attention updates to the logit difference between the acceptable and the unacceptable predictions, $\Delta \text{logit}^l_{(w-f)\gets \text{Self-attn}^l}$. As a general pattern, we observe that models tend to update more heavily on the layers where the alignment with linguistic phenomena is higher. This conclusion holds for larger models too, see the \texttt{darn} example in \Cref{table:distractor_15_gpt2xl}, where large logit updates are found in layers 28, 35, and 40, matching the layers where alignment peaks (\Cref{fig:update_logit_attn_xl} Top). In \texttt{IOI} and \texttt{SVA} tasks both models align with the evidence and increase their logit update towards the last layers. This indicates that models solve these phenomena once they have acquired sufficient contextual information.

Our findings in the \texttt{IOI} task support those by \citet{interpretability_in_the_wild}. In GPT-2 Small we observe high logit difference updates coming from the Indirect Object (IO) in layers 10 and 11. We further study the heads in those layers (\Cref{table:head_level_IOI}), where \citet{interpretability_in_the_wild} found `Name Mover Heads' and `Negative Mover Heads'. These heads rely on the IO to increase (Name Mover Heads) and decrease (Negative Mover Heads) respectively the logit of the correct prediction. In \Cref{apx:ioi_models} we provide an example of how every model solves the task across layers.

\begin{table}[!t]
    \footnotesize
    \centering
    \resizebox{0.472\textwidth}{!}{%
    \begin{tabular}{l} \toprule
    \multicolumn{1}{l}{Name Mover Head L10 H7}\\ 
 \mybox{color502}{\strut{Then,}} \mybox{color481}{\strut{Yvette}} \mybox{color500}{\strut{and}} \mybox{color1000}{\strut{Angie}} \mybox{color500}{\strut{were}} \mybox{color500}{\strut{working}} \mybox{color500}{\strut{at}} \mybox{color501}{\strut{the}} \mybox{color501}{\strut{mountain.}} \mybox{color498}{\strut{Yvette}} \mybox{color500}{\strut{decided}} \mybox{color501}{\strut{to}} \mybox{color501}{\strut{give}} \mybox{color501}{\strut{a}} \mybox{color498}{\strut{banana}} \mybox{color500}{\strut{to}} \mybox{color501}{\strut{\textbf{Angie}}}\\
\addlinespace
\multicolumn{1}{l}{Name Mover Head L10 H10}\\ 
 \mybox{color501}{\strut{Then,}} \mybox{color449}{\strut{Yvette}} \mybox{color499}{\strut{and}} \mybox{color1000}{\strut{Angie}} \mybox{color500}{\strut{were}} \mybox{color501}{\strut{working}} \mybox{color501}{\strut{at}} \mybox{color501}{\strut{the}} \mybox{color501}{\strut{mountain.}} \mybox{color493}{\strut{Yvette}} \mybox{color500}{\strut{decided}} \mybox{color501}{\strut{to}} \mybox{color500}{\strut{give}} \mybox{color500}{\strut{a}} \mybox{color502}{\strut{banana}} \mybox{color500}{\strut{to}} \mybox{color501}{\strut{\textbf{Angie}}}\\
\addlinespace
\multicolumn{1}{l}{Name Mover Head L11 H1}\\ 
 \mybox{color503}{\strut{Then,}} \mybox{color488}{\strut{Yvette}} \mybox{color499}{\strut{and}} \mybox{color1000}{\strut{Angie}} \mybox{color497}{\strut{were}} \mybox{color501}{\strut{working}} \mybox{color500}{\strut{at}} \mybox{color501}{\strut{the}} \mybox{color541}{\strut{mountain.}} \mybox{color496}{\strut{Yvette}} \mybox{color500}{\strut{decided}} \mybox{color499}{\strut{to}} \mybox{color501}{\strut{give}} \mybox{color499}{\strut{a}} \mybox{color469}{\strut{banana}} \mybox{color501}{\strut{to}} \mybox{color501}{\strut{\textbf{Angie}}}\\
\addlinespace
\multicolumn{1}{l}{Negative Name Mover Head L11 H8}\\ 
 \mybox{color499}{\strut{Then,}} \mybox{color523}{\strut{Yvette}} \mybox{color501}{\strut{and}} \mybox{color1}{\strut{Angie}} \mybox{color500}{\strut{were}} \mybox{color501}{\strut{working}} \mybox{color501}{\strut{at}} \mybox{color500}{\strut{the}} \mybox{color499}{\strut{mountain.}} \mybox{color515}{\strut{Yvette}} \mybox{color500}{\strut{decided}} \mybox{color501}{\strut{to}} \mybox{color501}{\strut{give}} \mybox{color500}{\strut{a}} \mybox{color500}{\strut{banana}} \mybox{color501}{\strut{to}} \mybox{color501}{\strut{\textbf{Angie}}}\\
     \addlinespace
     \multicolumn{1}{l}{Negative Name Mover Head L12 H11}\\ 
 \mybox{color509}{\strut{Then,}} \mybox{color629}{\strut{Yvette}} \mybox{color512}{\strut{and}} \mybox{color1}{\strut{Angie}} \mybox{color500}{\strut{were}} \mybox{color500}{\strut{working}} \mybox{color500}{\strut{at}} \mybox{color501}{\strut{the}} \mybox{color524}{\strut{mountain.}} \mybox{color555}{\strut{Yvette}} \mybox{color500}{\strut{decided}} \mybox{color500}{\strut{to}} \mybox{color504}{\strut{give}} \mybox{color500}{\strut{a}} \mybox{color500}{\strut{banana}} \mybox{color506}{\strut{to}} \mybox{color501}{\strut{\textbf{Angie}}}\\
\bottomrule
    \bottomrule
    \end{tabular}}
    \caption{GPT-2 Small updates to the logit prediction difference between \textbf{Angie} and \textbf{Yvette} in different heads produced by layer input token representations ($\Delta 
 \text{logit}^{l,h}_{(w-f)\gets{\boldsymbol{x}^{l-1}_j}}$).}
    \label{table:head_level_IOI}
\end{table}

\section{Analysis of MLPs}
The MLP block in the Transformer contains two learnable weight matrices\footnote{We omit bias terms.}: $\mW^l_1 \in \mathbb{R}^{d \times d_{mlp}}$ and $\mW^l_2 \in \mathbb{R}^{d_{mlp} \times d}$, and an element-wise non-linear activation function $\alpha$. It takes as input the state of the residual stream at timestep $t$ ($\tilde{\vx}^{l}_t$) and computes:
\begin{equation}\label{eq:mlp}
\vo^{l}_{t} = \alpha(\text{LN}(\tilde{\vx}^{l}_t)\mW^l_1) \mW^l_2
\end{equation}
Following, $\vo_t^{l}$ is added back to the residual stream (\Cref{fig:transformer_diagram}). \Cref{eq:mlp} can be seen as key-value memories \citep{geva-etal-2021-transformer}, where keys are stored in components of $\vk^l = \alpha(\text{LN}(\vx^{l}_t)\mW^l_1) \in \mathbb{R}^{d_{mlp}}$, and values ($\vv^l$) are rows of $\mW_2$. Following the key-value perspective, \Cref{eq:mlp} can be rewritten as
\begin{equation}
    \vo^{l}_{t} = \sum_i^{d_{mlp}} k_i^l \vv^l_i
\end{equation}
where $\vv_i^l$ represents the $i$-th row of $\mW_2$. Recalling how the final logit of a token $w$ is decomposed by layer-wise updates in \Cref{eq:decomposition_final_logits}, the $\text{MLP}^{l}$ updates the logit of $w$ as follows:
\begin{equation}
\begin{aligned}
    \Delta \text{logit}^l_{w \gets \text{MLP}^{l}} &=\vo^{l}_{t} \mU_w^{\intercal}\\
    &= \sum_i^{d_{mlp}} k_i^l \vv^l_i \mU_w^{\intercal}\\
    &= \sum _i^{d_{mlp}} \Delta \text{logit}^l_{w \gets k_i^l \vv^l_i}
\end{aligned}
\end{equation}
Thus, the update of the MLP can be decomposed into sub-updates \citep{geva-etal-2022-transformer} performed by each $k_i^l\vv^l_i$ (weighted row in $\mW^l_2$). The update in the logit's difference between the target and foil tokens by each value $i$ is therefore:
\begin{equation}
\Delta \text{logit}^l_{(w - f) \gets k_i^l \vv^l_i} = \Delta \text{logit}^l_{w \gets k_i^l \vv^l_i} - \Delta \text{logit}^l_{f \gets k_i^l \vv^l_i}
\end{equation}

\begin{figure}
\centering     
\subfigure[]{\label{fig:mlp_subupdates_results_a}\includegraphics[width=0.245\textwidth]{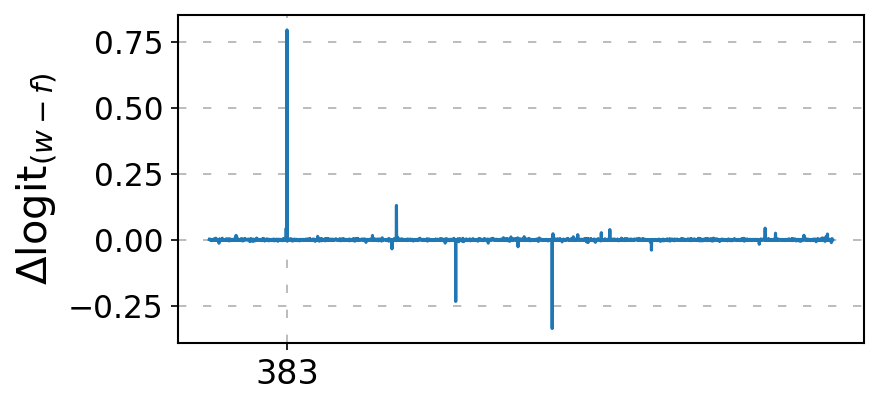}}
\subfigure[]{\label{fig:mlp_subupdates_results_b}\includegraphics[width=0.229\textwidth]{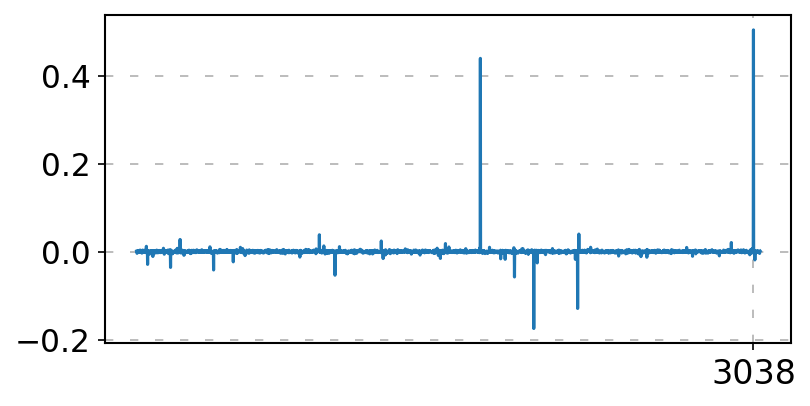}}\\
\subfigure[]{\label{fig:gpt2_small_l12_darn_singular}\includegraphics[width=0.245\textwidth]{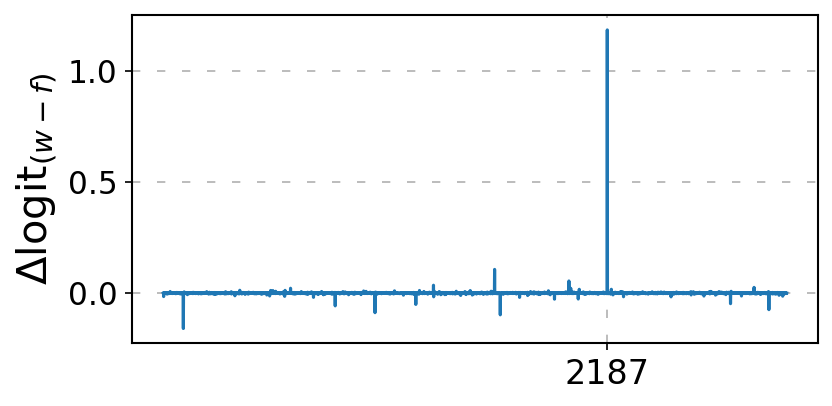}}
\subfigure[]{\label{fig:gpt2_small_l12_sva_4_singular}\includegraphics[width=0.229\textwidth]{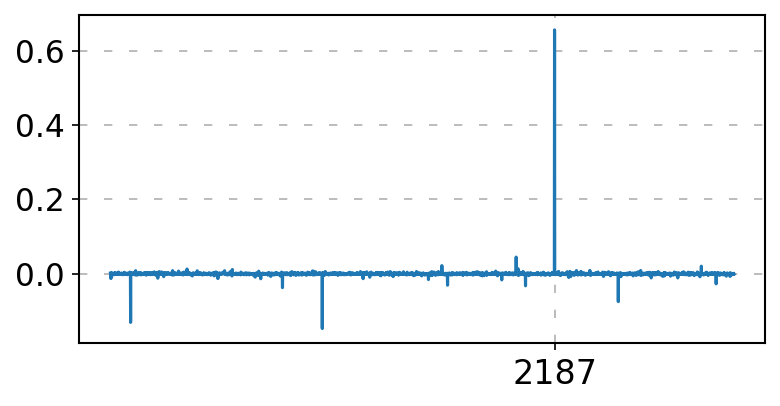}}
\caption{Average (across the dataset) of the updates to the logit difference caused by the weighted values in the MLP (each row $i$ in $\mW_2^l$), $\Delta \text{logit}^l_{(w - f)_\gets{k_i^l \vv^l_i}}$. a) \texttt{dna}: dimension $i\!=\!\texttt{383}$ (L11) promotes singular nouns (increases the logit difference between singular and plural nouns) after \texttt{this}/\texttt{that}, b) \texttt{dna}: dimension $i\!=\!3038$ (L11) promotes plural nouns after \texttt{these}/\texttt{those}. Dimension $i\!=\!\texttt{2187}$ (L12) pushes the prediction of singular verbs in different Subject-Verb Agreement datasets c) \texttt{darn} and d) \texttt{SVA}.}
\label{fig:mlp_subupdates_results}
\end{figure}

\begin{figure}[t]
	\begin{centering}
	\includegraphics[width=0.49\textwidth]{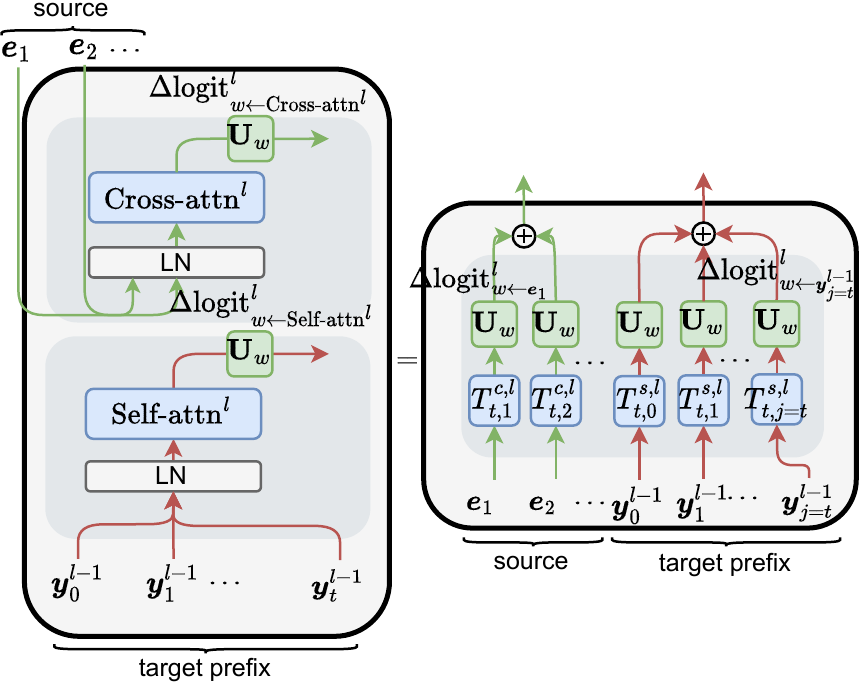}
	\caption{Cross-attention block in the Transformer's decoder (left) and its equivalent using vector transformations (right). Depicted in green and red it's shown the information coming from the encoder and the decoder (target prefix) respectively.}
	\label{fig:cross_attn_embed_space_proj}
	\end{centering}
\end{figure}

In \Cref{fig:mlp_subupdates_results}, we show some examples of the contribution of each weighted value $k_i^l \vv^l_i$ to the logit difference between the acceptable target token and the unacceptable one, at different layers and datasets. We can observe that there is a small subset of values that consistently increase the difference in logits helping to solve the linguistic task. Some of them include the value $i\!=\!\texttt{383}$ in layer 10 (\Cref{fig:mlp_subupdates_results} (a)), which increases the logit of singular nouns and reduces the plural ones when the determiner is \texttt{this} or \texttt{that}. For instance, in the sentence “William described this \_\_\_”, value $i\!=\!\texttt{383}$ increases the logit difference between \texttt{movie} and \texttt{movies}. In dimension \texttt{3038} we find a value upweighting the logits of the plural nouns over the singular ones when the determiner is \texttt{these} or \texttt{those} (\Cref{fig:mlp_subupdates_results} (b)). These values help solve the linguistic task at hand across different subsets, for instance, the value in dimension $i\!=\!\texttt{2187}$ is in charge of promoting the singular form of the verb when the head of the subject is singular too. This occurs in both \texttt{darn} and \texttt{SVA} subsets.

\section{Neural Machine Translation}
An NMT system estimates the likelihood of a target sequence of tokens, $\mathbf{y} = (y_1, \ldots, y_t)$, given a source sequence of tokens, $\mathbf{x} = (x_1, \ldots, x_I)$:
\begin{equation}
P(\mathbf{y}|\mathbf{x}) = \prod_{s}^{t} P(y_s|\mathbf{y}_{<s},\mathbf{x})
\end{equation}
where $\mathbf{y}_{<s} = (y_0, \ldots, y_{s-1})$ is the prefix of $y_{s}$, and $x_{I} = y_{0} = \eos$ is a special token used to mark the start and end of the sentence. The encoder processes the source sentence and generates a sequence of contextualized representations, $\mathbf{e} = (\ve_1, \ldots, \ve_i)$. At each decoding step $t$, the decoder uses the encoder outputs and the target prefix to compute a probability distribution over the target vocabulary.

\paragraph{Cross-attention.} Similar to \Cref{eq:transformed_vectors}, the output of the cross-attention ($\widetilde{\vy}_{t}^{c,l}$) and self-attention ($\widetilde{\vy}_{t}^{s,l}$) (\Cref{fig:cross_attn_embed_space_proj}) of a decoder layer in an encoder-decoder Transformer can be decomposed\footnote{Removing biases.} as
\begin{equation}\label{eq:transformed_vectors_cross_attn}
\widetilde{\vy}_{t}^{c,l} = \sum_{j}^t T^{c,l}_{t,i}(\ve_i),\;\;\; \widetilde{\vy}_{t}^{s,l} = \sum_{j}^t T^{s,l}_{t,j}(\vy_j^{l-1})
\end{equation}

As shown in \Cref{fig:cross_attn_embed_space_proj}, each transformed vector updates the logits of the token predictions by multiplying it with the corresponding column of $\mU$, as in \Cref{eq:indiv_contrib}:
\begin{equation}
\Delta \text{logit}^l_{w\gets{\ve_i}} = T^{c,l}_{t,i}(\ve_i) \mU_w
\end{equation}

\begin{table}[!t]
\centering
\resizebox{0.45\textwidth}{!}{%
\begin{tabular}{@{}lcc@{}}
\toprule
                                                   & \multicolumn{2}{c}{\textbf{AER} ($\downarrow$)} \\ \cmidrule(l){2-3} 
\multicolumn{1}{c}{\textbf{Method}}                         & \textbf{Bilingual}       & \textbf{M2M}         \\ \midrule
Attention weights                                  & $48.6$                   & 96.4                \\
SD-SmoothGrad \citep{ding-etal-2019-saliency}      & $36.4$                   & -                    \\
Vector Norms \citep{kobayashi-etal-2020-attention} & $41.4$                   & -                    \\
Distance Vectors-Output \citep{alti_plus}                                  & $38.8$                   & 36.4                \\
Proposed alignment extraction                                & \textbf{26.0}            & \textbf{27.3}       \\ \bottomrule
\end{tabular}
}
\caption{Mean AER of the cross-attention contributions in the best layer of the bilingual and M2M models. For the bilingual model, we show the average on five different seeds.}
\label{tab:aer_results}
\end{table}

\paragraph{Alignment.} Source-target alignments derived from attention weights in NMT systems can be unreliable \cite{Zenkel_2019,li-etal-2019-word,garg_jointly_2019}, with upper layers producing better alignments. A limitation of using this method to interpret model predictions is that the ground truth target word may not match the model's actual prediction. However, by measuring how the encoder token representations update the logits of the reference words, $\Delta \text{logit}^l_{w\gets{\ve_i}}$, we can more precisely explain which source word causes the final logit of the reference word, even if it is not one of the top predictions.

Following \citet{kobayashi-etal-2020-attention} and \citet{ding-etal-2019-saliency} setting, we train a 6-layer Transformer model for the German-English (De-En) translation task using Europarl v7 corpus\footnote{\url{http://www.statmt.org/europarl/v7}} \citet{koehn-2005-europarl}. We also evaluate on M2M, a 12 layer multilingual model \citep{m2m_100}. We use \citet{Vilar2006AERDW} dataset, consisting of 508 De-En human annotated sentence pairs with alignments, and compare them with our extracted alignments using Alignment Error Rate (AER). We also show results of other attention-based alignments extraction methods. Vector Norms take the norm of the transformed vectors in \Cref{eq:transformed_vectors_cross_attn}, Distance Vectors-Output measures the distance between the transformed vectors and the attention block output $\widetilde{\vy}_{t}^{c,l}$. SD-SmoothGrad relies on gradients to extract alignments. In \Cref{tab:aer_results} we show that our proposed method achieves lower AER values, which indicates that NMT models generate human-like alignments for building model predictions.

\begin{figure}[t]
	\begin{centering}
	\includegraphics[width=0.49\textwidth]{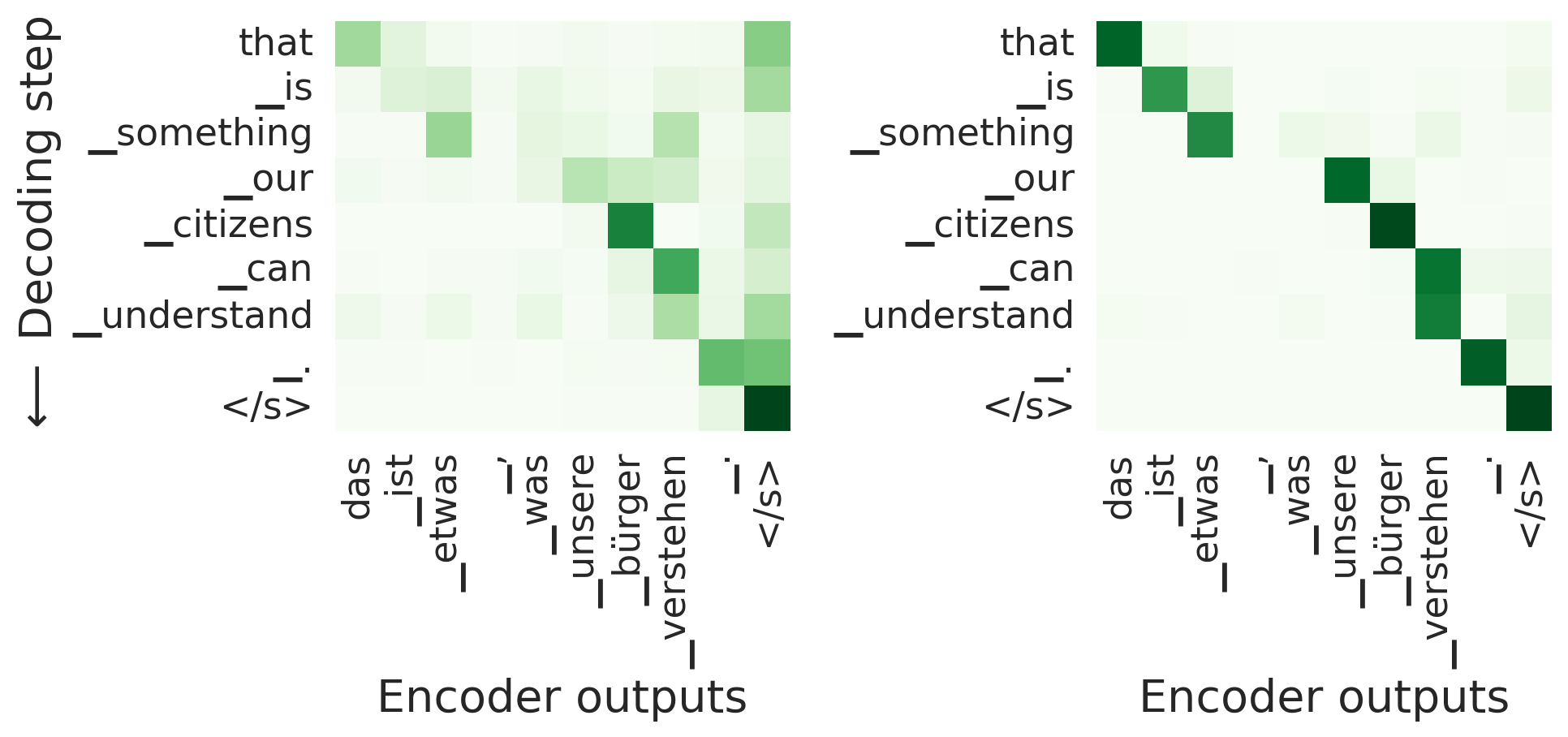}
	\caption{Left: attention weights in the cross-attention in the penultimate layer. Right: contributions obtained as logit updates to token predictions in the penultimate layer.}
	\label{fig:joint_attn_weights_logit_update}
	\end{centering}
\end{figure}

\section{Related Work}
The projection of LMs representations and model parameters to the vocabulary space has been a subject of previous research \citep{belrose2023eliciting, din2023jump}. \citet{geva-etal-2021-transformer, geva-etal-2022-transformer} view feed-forward layers as performing updates to the probability distribution of the token predictions. \citet{10.1162/tacl_a_00501} study how the different Transformer modules contribute to the hidden representations, and \citet{dar_embedding_space} directly interpret Transformer static parameters in the embedding space.
In this work, our focus lies in interpreting the influence of input tokens and its representations in the model predictions.

Furthermore, work on mechanistic interpretability \citep{mech_int} has discovered `circuits' within LMs in charge of solving tasks \citep{interpretability_in_the_wild, geva2023dissecting}. In contrast to their methods, our approach does not rely on causal interventions in the computations of Transformers. More broadly, our work can be related to those explaining the prediction process of LMs \citep{tenney-etal-2019-bert, voita-etal-2019-bottom,sarti-etal-2023-inseq}.

\section{Conclusions}
In this paper, we introduce a new procedure for analyzing language generation models by combining the residual stream perspective with interpretable attention decomposition, and tested our approach using contrastive examples in Transformer LMs. We found that the explanations provided by our proposed methods, Logit and ALTI-Logit, align better with available linguistic evidence in the context of the sentence, compared to common gradient-based and erasure-based baselines. We also analyzed the role of MLPs and showed that they assist the model in determining predictions that conform to the grammar rules. Additionally, we applied our method to a Machine Translation model and demonstrated that it generates human-like alignments for building predictions. Overall, our results suggest that decomposing the logit scores is an effective way to analyze language generation models.

\section{Limitations}


The experimental methodology employed in this study for both contrastive explanations and NMT is not directly extensible to languages other than English, due to the scarcity of resources such as models and annotations.

The datasets employed in this study to evaluate contrastive explanations across various linguistic paradigms are restricted to sentences that possess a well-defined structure. As a result, it is possible that the conclusions drawn may not be generalizable to the broader distribution of sentences.

Lastly, it should be noted that the method proposed in this study should not be used as a definitive explanation of model predictions in any other context. It is recommended to use the method as a debugging tool and should be employed in conjunction with other methods to gain a comprehensive understanding of model predictions.


\section{Ethics statement}

It is acknowledged that the experiments reported in this study are limited to high-resource languages. However, the methodology employed is language-independent and may be applied to other languages in the future, provided that adequate annotated data becomes available.

\section{Acknowledgements}
We would like to thank the anonymous reviewers for their useful comments. Javier Ferrando, Gerard I. Gállego and Ioannis Tsiamas are supported by the Spanish Ministerio de Ciencia e Innovación through the project PID2019-107579RB-I00 / AEI / 10.13039/501100011033.

\bibliography{anthology,custom}
\bibliographystyle{acl_natbib}
\newpage
\appendix
\clearpage
\onecolumn
\section{Pre-LN Self-attention Decomposition}\label{appx:self_attn_decomp}
\begin{table*}[!h]
\centering
\begin{tabular}{cc}
\toprule
$\vx_j^{l-1} \in \mathbb{R}^{d \times d_h}$         & Layer Input (Residual Stream position $j$) \\
$\mA^{l,h} \in \mathbb{R}^{t \times t}$             & Attention Matrix                           \\
$\mW_V^{l,h} \in \mathbb{R}^{d \times d_h}$         & Values Weight Matrix                        \\
$\mW_O^{l,h} \in \mathbb{R}^{d_h \times d}$         & Output Weight Matrix (per head)                      \\
$\vb^{l,h}_{V} \in \mathbb{R}^{d_h}$                    & Value bias                                 \\
$\vb^{l}_{O} \in \mathbb{R}^{d}$                    & Output bias             \\
$H \in \mathbb{R}$                    & Number of heads             \\
$\text{LN}^{l}: \mathbb{R}^{d} \mapsto \mathbb{R}^{d}$ & Layer Normalization                        \\
\bottomrule
\end{tabular}
\caption{Components of the self-attention module.}
\end{table*}
At position $t$, each head of a Pre-LN self-attention mechanism computes:
\begin{equation}
\vz^{l,h}_t = \sum_{j}^{t} \Bigl(\underbrace{\text{LN}^l(\vx^{l-1}_j)\mW^{l,h}_{V} + \vb^{l,h}_{V}}_{\text{$j$-th value}}\Bigr)\mA_{t,j}^{l,h}
\end{equation}

By representing attention heads as parallel independent components, we can express the output of the self-attention as
\begin{equation}
\widetilde{\vx}_{t}^{l} = \sum^H_h \vz^{l,h}_t\mW_O^{l,h} + \vb_O^{l}
\end{equation}
leading to:
\begin{align}\label{eq:complete_attn_decomposition}
\widetilde{\vx}_{t}^{l} = \sum_{j}^{t} \sum_h^H \left(\text{LN}^l(\vx^{l-1}_j)\mW^{l,h}_{V} + \vb^{l,h}_{V}\right)\mA_{t,j}^{l,h}\mW_{O}^{l,h} + \vb^{l}_{O}
\end{align}

The layer normalization computes:
\begin{equation}
\text{LN}^l(\vx^{l-1}_j)=\frac{\vx^{l-1}_j-\mu(\vx^{l-1}_j)}{\sigma(\vx^{l-1}_j)} \odot \mathbf{\gamma}^l+ \mathbf{\beta}^l
\end{equation}
with $\mu$ and $\sigma$ computing the mean and standard deviation, and $\gamma^l \in \mathbb{R}^{d}$ and $\beta^l \in \mathbb{R}^{d}$ refer to learned element-wise transformation and bias respectively. Considering $\sigma(\vx^{l-1}_j)$ as a constant, $\text{LN}$ can be treated as a constant affine transformation:
\begin{equation}\label{eq:ln_linearization}
\text{LN}(\vx^{l-1}_j) = \vx^{l-1}_j \mL^l + \mathbf{\beta}^l
\end{equation}
where $\mL^{l} \in \mathbb{R}^{d \times d}$ represents a matrix that combines centering, normalizing, and scaling operations together. 

Using \Cref{eq:ln_linearization} in \Cref{eq:complete_attn_decomposition}:
\begin{align*}
\widetilde{\vx}_{t}^{l} &= \sum_{j}^{t} \sum_h^H \biggl(\left((\vx^{l-1}_j \mL^l + \mathbf{\beta}^l)\mW^{l,h}_{V} + \vb^{l,h}_{V}\right)\mA_{t,j}^{l,h}\mW_{O}^{l,h}\biggl) + \vb^{l}_{O}\\
&= \sum_{j}^{t} \sum_h^H \biggl(\left(\vx^{l-1}_j\mL^l\mW^{l,h}_{V} + \mathbf{\beta}^l\mW^{l,h}_{V} + \vb^{l,h}_{V}\right)\mA_{t,j}^{l,h}\mW_{O}^{l,h}\biggl) + \vb^{l}_{O}\\
&= \sum_{j}^{t} \sum_h^H \biggl(\vx^{l-1}_j\mL^l\mW^{l,h}_{V}\mA_{t,j}^{l,h}\mW_{O}^{l,h} + \mathbf{\beta}^l\mW^{l,h}_{V}\mA_{t,j}^{l,h}\mW_{O}^{l,h} + \vb^{l,h}_{V}\mA_{t,j}^{l,h}\mW_{O}^{l,h}\biggl) + \vb^{l}_{O}\\
&= \sum_{j}^{t} \sum_h^H \biggl(\vx^{l-1}_j\mL^l\mW^{l,h}_{V}\mA_{t,j}^{l,h}\mW_{O}^{l,h} + \mA_{t,j}^{l,h}\left(\mathbf{\beta}^l\mW^{l,h}_{V}\mW_{O}^{l,h} + \vb^{l,h}_{V}\mW_{O}^{l,h}\right)\biggl) + \vb^{l}_{O}\numberthis \label{eqn}
\end{align*}

Considering $\theta^{l,h} = \left(\mathbf{\beta}^l\mW^{l,h}_{V} + \vb^{l,h}_{V}\right)\mW_{O}^{l,h}$

\begin{align*}
\widetilde{\vx}_{t}^{l} &= \sum_{j}^{t}\sum_{h}^{H}\biggl( \vx^{l-1}_j\mL^l\mW^{l,h}_{V}\mA_{t,j}^{l,h}\mW_{O}^{l,h} + \mA_{t,j}^{l,h}\theta^{l,h}\biggl) + \vb^{l}_{O}\numberthis
\end{align*}

For each $j$-th input term, $H$ affine transformations are applied to $\vx_j$. Furthermore, all heads' operations can be further grouped into a single affine transformation:
\begin{align}
\widetilde{\vx}_{t}^{l} &= \sum_{j}^{t}\biggl(\vx^{l-1}_j\mL^l\sum_{h}^{H}\mW^{l,h}_{V}\mA_{t,j}^{l,h}\mW_{O}^{l,h} + \sum_{h}^{H}\mA_{t,j}^{l,h}\theta^{l,h}\biggl) + \vb^{l}_{O}
\end{align}

So, we can write $\widetilde{\vx}_{t}^{l}$ as a sum of $t$ affine transformations, and the output bias:
\begin{equation}\label{eq:linearity_t}
\widetilde{\vx}_{t}^{l} =\sum_{j}^{t} T^l_{t,j}(\vx^{l-1}_j) + \vb^{l}_{O}
\end{equation}

\section{Tracking Logits to the Input with Rollout}\label{apx:track_logits_rollout}
The rollout method \citep{abnar-zuidema-2020-quantifying} assumes any intermediate representation is a linear combination of the model inputs, ${\vx_j^{l-1} = \sum_s m^{l-1}_{j,s} \vx^{0}_s}$, where $m^{l-1}_{j,s}$ is a score indicating the contribution of input token $s$ to the $l-1$ representation (or residual path) of token $j$. By dividing the logit update performed by $\vx_j^{l-1}$ among the model inputs ($\Delta \text{logit}^l_{w,j \gets {\vx^{0}_{s}}}$) based on their contributions to $\vx_j^{l-1}$, we obtain:
\begin{align*}
 \Delta \text{logit}^l_{w \gets {\vx^{l-1}_{j}}}  &= \sum_s \Delta \text{logit}^l_{w,j \gets {\vx^{0}_{s}}}\\
 &= \sum_s m^{l-1}_{j,s} \Delta \text{logit}^l_{w \gets {\vx^{l-1}_{j}}}\numberthis
 \end{align*}

Based on the total logit update produced in layer $l$, we have that:
\begin{align*}
 \Delta \text{logit}^l_{w \gets {\text{Self-attn}^l}} &= \sum_j \Delta \text{logit}^l_{w \gets {\vx^{l-1}_{j}}}\\
 &= \sum_j \sum_s \Delta \text{logit}^l_{w,j \gets {\vx^{0}_{s}}}\\
 &= \sum_j \sum_s m^{l-1}_{j,s} \Delta \text{logit}^l_{w \gets {\vx^{l-1}_{j}}}\\
 &= \sum_s \sum_j m^{l-1}_{j,s} \Delta \text{logit}^l_{w \gets {\vx^{l-1}_{j}}}\\
 &= \sum_s \Delta \text{logit}^l_{w \gets s}\numberthis
\end{align*}
So, we have obtained \Cref{eq:input_model_contrib_layer}: 
\begin{equation}
 \Delta \text{logit}^l_{w \gets s} = \Delta \text{logit}^l_{w\gets{\mathbf{x}^{l-1}}}\; \mM^{l-1}_{s}
\end{equation}
\clearpage
\twocolumn
\section{Results}\label{sec:apx_results}
    \vspace{3ex}

\noindent\begin{minipage}{\textwidth}
        \centering
        \includegraphics[width=0.90\linewidth]{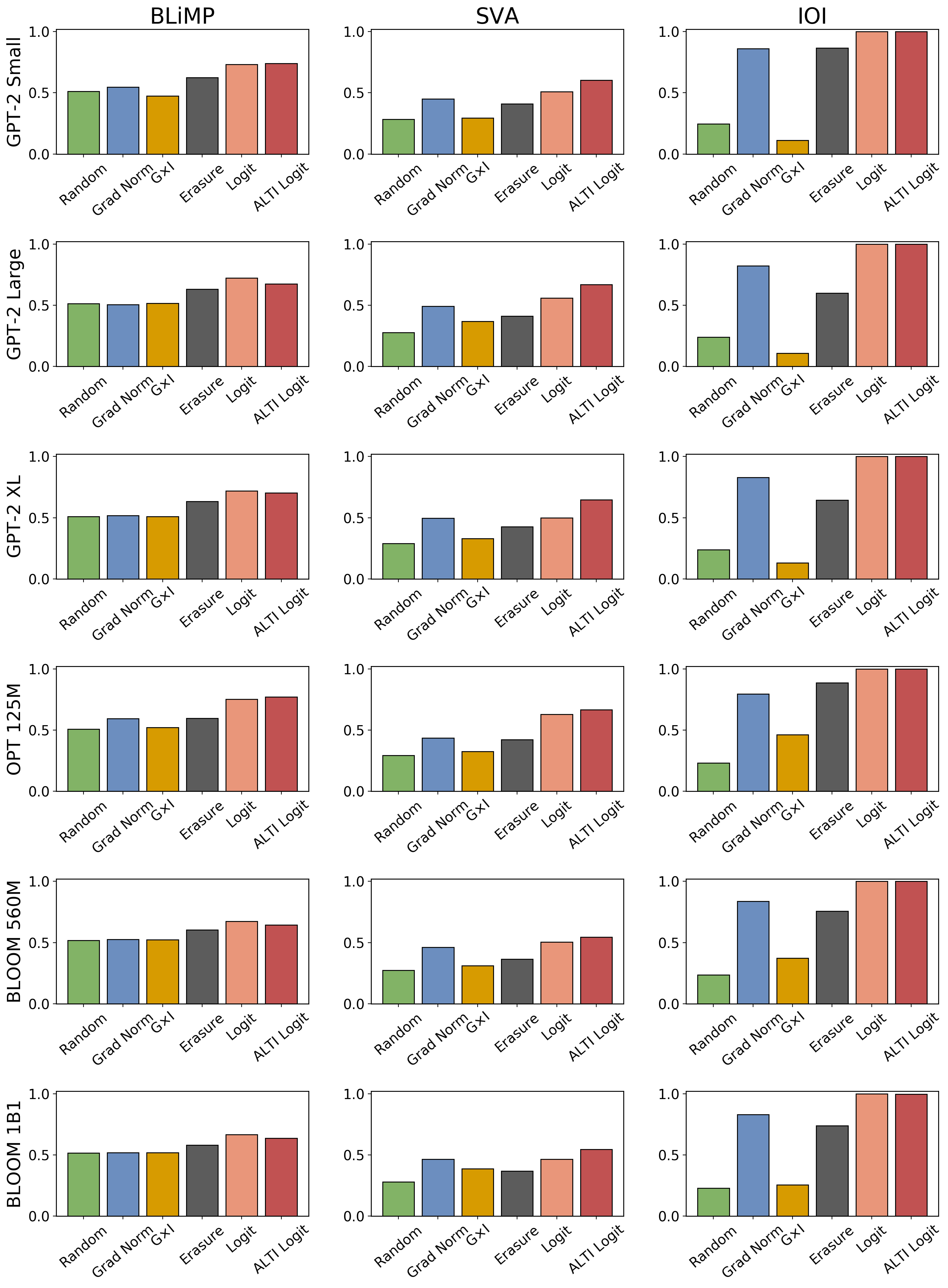}
        \captionof{figure}{Alignment (MRR $\uparrow$) of different explanation methods of GPT-2 Small, Large, and XL, OPT 125M, BLOOM 560M, and BLOOM 1B1 model predictions with BLiMP, SVA, and IOI datasets.}
        \label{fig:mrr_all}
    \end{minipage}

\clearpage

\subsection{GPT-2 Small Results}
\begin{table}[H]
\centering
\small
\resizebox{\linewidth}{!}{%
\begin{tabular}{cccccccc}
\toprule
\textbf{Dataset} & \textbf{Erasure} & \textbf{Logit} & \textbf{ALTI-Logit} & \textbf{Grad Norm} & \textbf{\gradinput} & \textbf{Random} & \textbf{Distance} \\ \midrule
aga   & 0.959   & 0.827             & 0.964                   & 0.793         & 0.791        & 0.699 & 3.2  \\
ana   & 0.963   & 0.817             & 0.976                   & 0.675         & 0.739        & 0.716 & 3.2  \\
asp   & 0.492   & 0.386             & 0.499                   & 0.751         & 0.409        & 0.381 & 3.3  \\
dna   & 0.35    & 0.737             & 0.646                   & 0.363         & 0.387        & 0.459 & 1  \\
dnai  & 0.374   & 0.711             & 0.637                   & 0.408         & 0.432        & 0.466 & 1  \\
dnaa  & 0.61    & 0.951             & 0.807                   & 0.263         & 0.321        & 0.397 & 2.1  \\
dnaai & 0.659   & 0.9               & 0.757                   & 0.263         & 0.339        & 0.406 & 2.1  \\
npi   & 0.663   & 0.445             & 0.417                   & 0.785         & 0.495        & 0.599 & 3.2  \\
darn  & 0.557   & 0.802             & 0.949                   & 0.617         & 0.363        & 0.488 & 3.9  \\
SVA 1 & 0.389   & 0.558             & 0.641                   & 0.432         & 0.298        & 0.333 & 8  \\
SVA 2 & 0.425   & 0.57              & 0.606                   & 0.421         & 0.303        & 0.292 & 11.6  \\
SVA 3 & 0.454   & 0.459             & 0.603                   & 0.51          & 0.356        & 0.259 & 12.9  \\
SVA 4 & 0.371   & 0.454             & 0.566                   & 0.433         & 0.222        & 0.249 & 16.4  \\
IOI   & 0.865   & 1.0               & 1.0                     & 0.86          & 0.111        & 0.245 & 14.9 \\
\bottomrule
\end{tabular}
}
\caption{MRR Alignment of different explanation methods on GPT-2 Small predictions on every dataset. The average distance to the linguistic evidence tokens is shown in the last column.}
\label{tab:gpt2small_mrr}
\end{table}

\subsection{GPT-2 XL Results}

\begin{table}[H]
\centering
\small
\resizebox{\linewidth}{!}{%
\begin{tabular}{cccccccc}
\toprule
\textbf{Dataset} & \textbf{Erasure} & \textbf{Logit} & \textbf{ALTI-Logit} & \textbf{Grad Norm} & \textbf{\gradinput} & \textbf{Random} & \textbf{Distance} \\ \midrule
aga   & 0.974   & 0.79              & 0.974                   & 0.778         & 0.713        & 0.681  & 3.2      \\
ana   & 0.945   & 0.777             & 0.964                   & 0.721         & 0.655        & 0.71   & 3.2      \\
asp   & 0.506   & 0.368             & 0.514                   & 0.721         & 0.44         & 0.369  & 3.3      \\
dna   & 0.326   & 0.655             & 0.539                   & 0.255         & 0.486        & 0.465  & 1        \\
dnai  & 0.366   & 0.598             & 0.524                   & 0.264         & 0.515        & 0.453  & 1        \\
dnaa  & 0.631   & 0.932             & 0.615                   & 0.205         & 0.352        & 0.413  & 2.1      \\
dnaai & 0.644   & 0.874             & 0.529                   & 0.205         & 0.359        & 0.393  & 2.1      \\
npi   & 0.735   & 0.602             & 0.711                   & 0.82          & 0.586        & 0.594  & 3.2      \\
darn  & 0.576   & 0.873             & 0.945                   & 0.686         & 0.477        & 0.51   & 3.9      \\
SVA 1 & 0.416   & 0.564             & 0.638                   & 0.467         & 0.365        & 0.352  & 8        \\
SVA 2 & 0.455   & 0.558             & 0.646                   & 0.489         & 0.353        & 0.269  & 11.6     \\
SVA 3 & 0.424   & 0.455             & 0.678                   & 0.535         & 0.343        & 0.31   & 12.9     \\
SVA 4 & 0.411   & 0.418             & 0.625                   & 0.489         & 0.256        & 0.226  & 16.4     \\
IOI   & 0.643   & 1.0               & 1.0                     & 0.829         & 0.131        & 0.239  & 14.9       \\
\bottomrule
\end{tabular}
}
\caption{MRR Alignment of different explanation methods on GPT-2 XL predictions on every dataset. The average distance to the linguistic evidence tokens is shown in the last column.}
\label{tab:gpt2xl_mrr}
\end{table}

\section{Model Predictions}\label{apx:model_predictions}

\begin{figure}[H]
\begin{centering}
\includegraphics[width=0.41\textwidth]{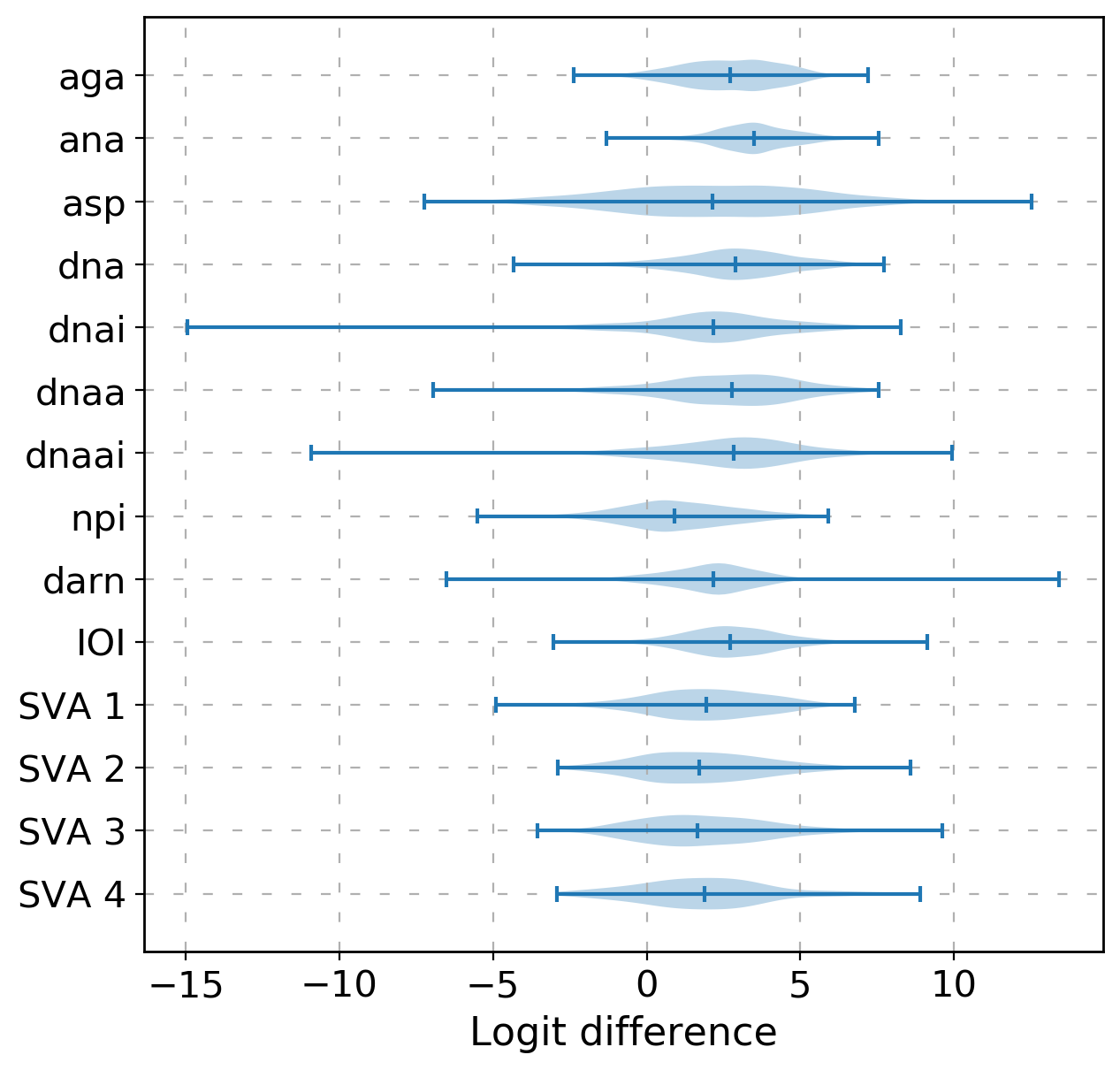}
\caption{Logit difference between the acceptable and the unacceptable predictions of a GPT-2 Small on every dataset.}
\label{fig:logit_difference_gpt2_small_blimp}
\end{centering}
\end{figure}

\begin{figure}[H]
    \begin{centering}
    \includegraphics[width=0.41\textwidth]{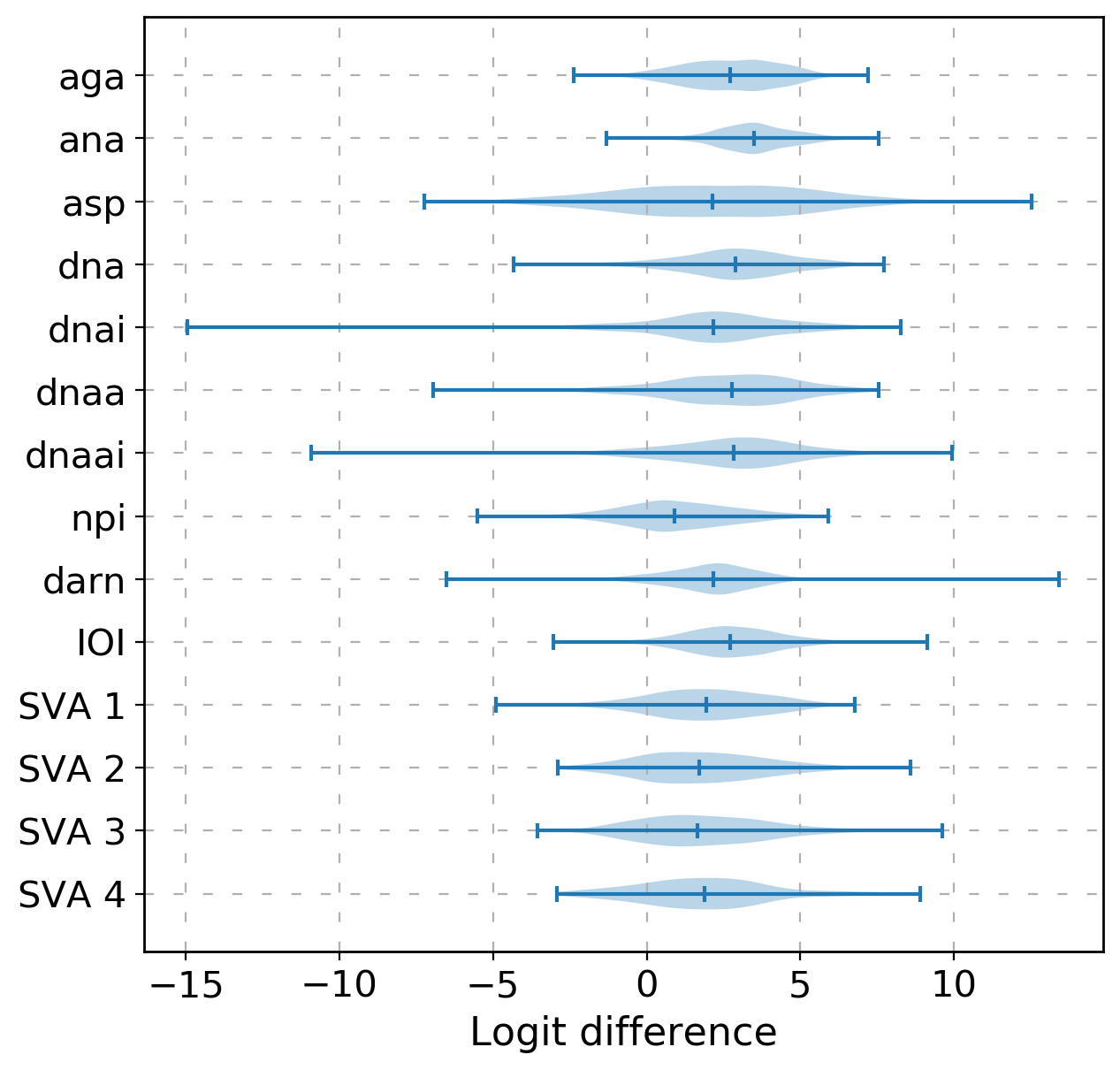}
    \caption{Logit difference between the acceptable and the unacceptable predictions of a GPT-2 XL on every dataset.}
    \label{fig:logit_difference_gpt2_xl_blimp}
    \end{centering}
\end{figure}

\section{MRR Alignment across layers}\label{apx:alignment_gpt2_datasets}
\begin{figure}[H]
\begin{centering}
\includegraphics[width=0.49\textwidth]{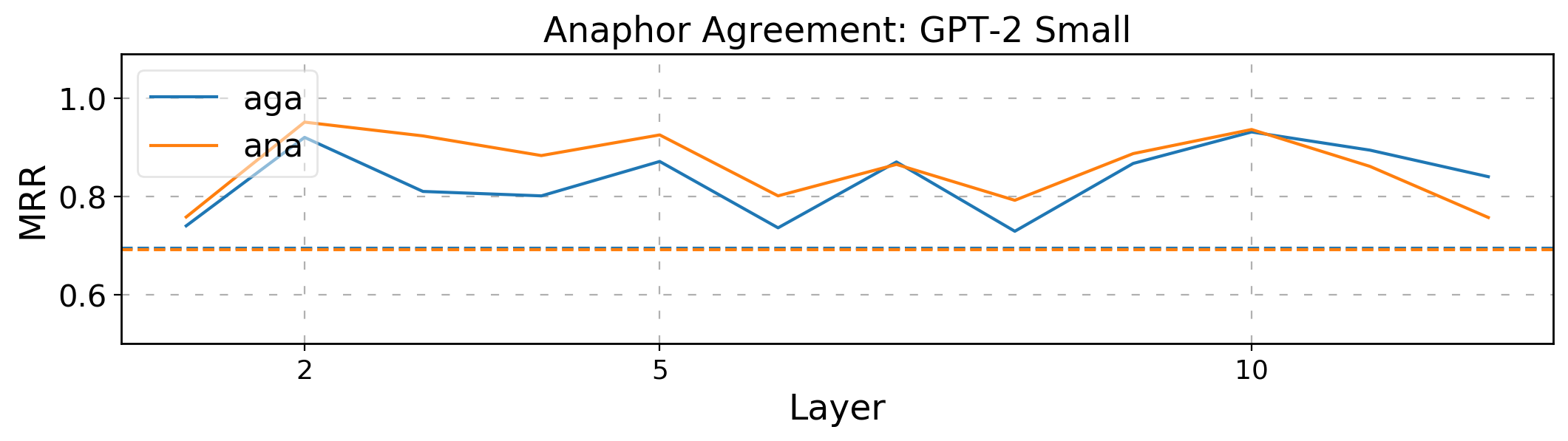}
\caption{ALTI-Logit MRR alignment scores across layers on Anaphor Agreement datasets (GPT-2 Small).}
\label{fig:layerwise_anaphor_gpt2_small}
\end{centering}
\end{figure}

\begin{figure}[H]
\begin{centering}
\includegraphics[width=0.49\textwidth]{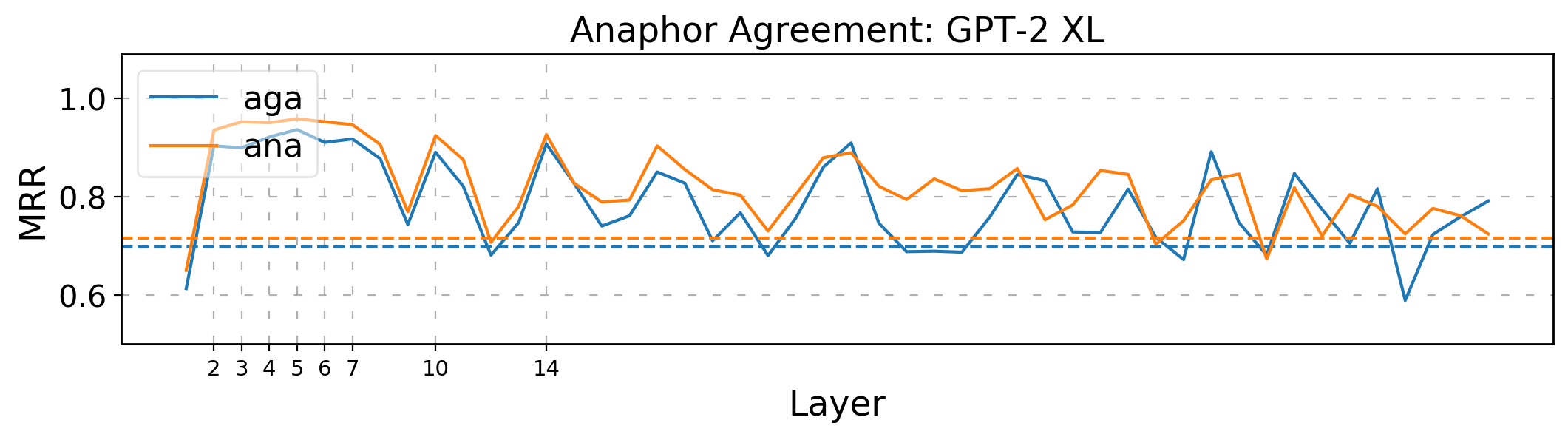}
\caption{ALTI-Logit MRR alignment scores across layers on Anaphor Agreement datasets (GPT-2 XL).}
\label{fig:layerwise_anaphor_gpt2_xl_layer}
\end{centering}
\end{figure}

\begin{figure}[H]
\begin{centering}
\includegraphics[width=0.49\textwidth]{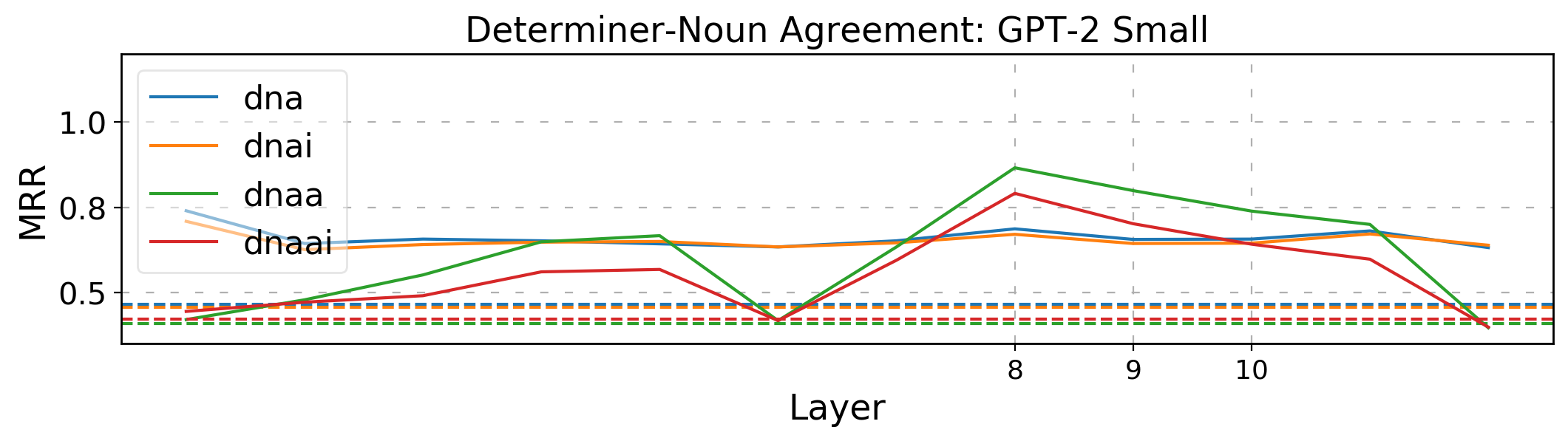}
\caption{ALTI-Logit MRR alignment scores across layers on Determiner-Noun Agreement datasets (GPT-2 Small).}
\label{fig:layerwise_determiner_gpt2_small_layer}
\end{centering}
\end{figure}

\begin{figure}[H]
\begin{centering}
\includegraphics[width=0.49\textwidth]{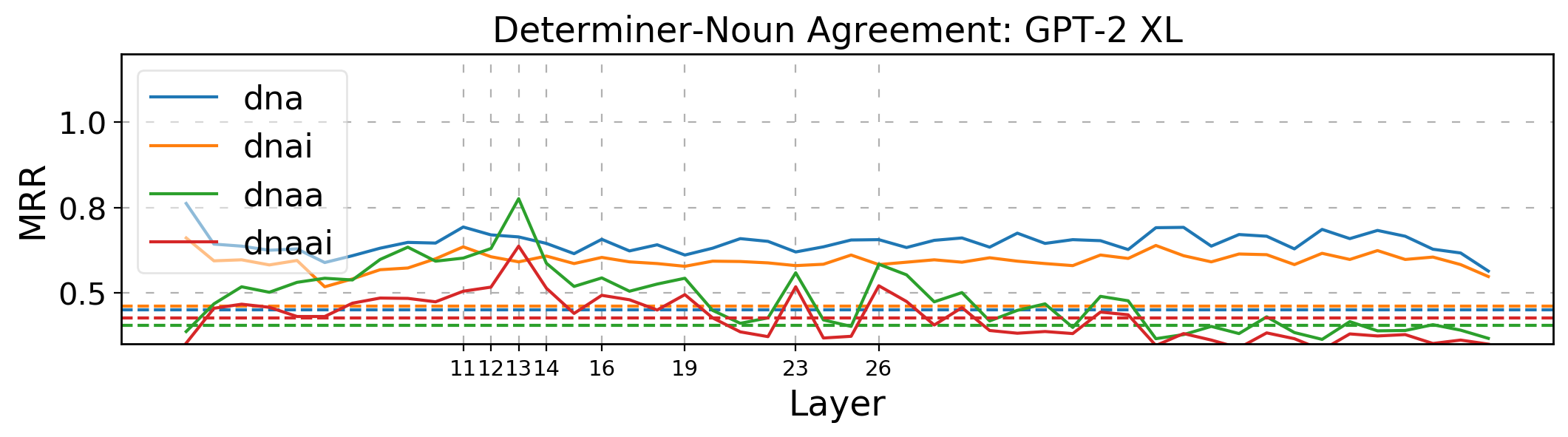}
\caption{ALTI-Logit MRR alignment scores across layers on Determiner-Noun Agreement datasets (GPT-2 XL).}
\label{fig:layerwise_determiner_gpt2_xl_layer}
\end{centering}
\end{figure}

\section{MLPs Logit Difference Update}\label{sec:update_logits_mlp_gpt2_small}
\begin{figure}[H]
\begin{centering}
\includegraphics[width=0.34\textwidth]{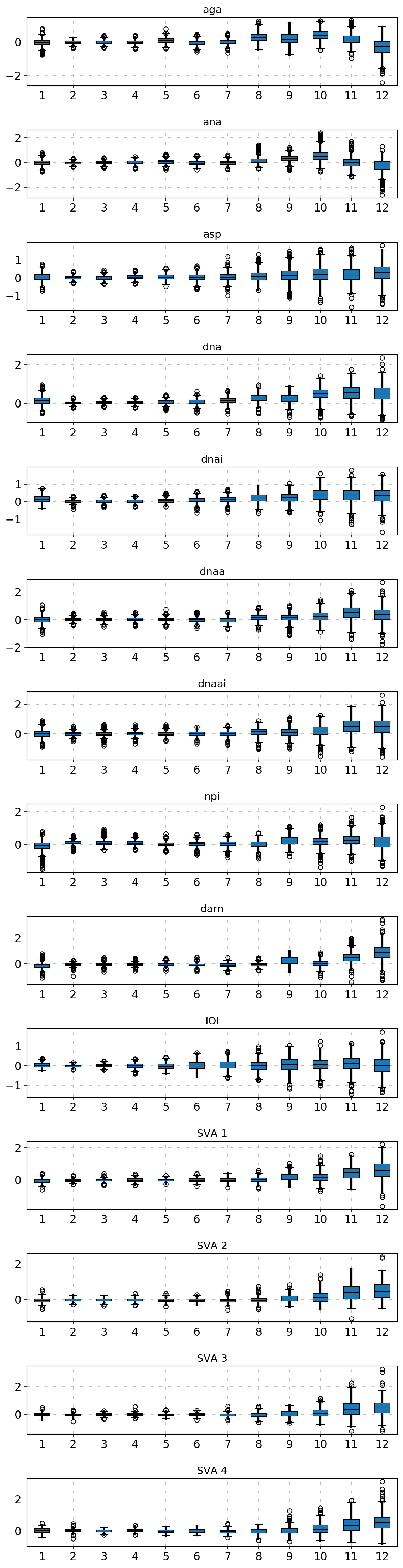}
\caption{MLPs update to the logit difference $\Delta \text{logit}^l_{(w-f) \gets \text{MLP}^{l}}$ across layers (GPT-2 Small).}
\label{fig:gpt2_small_mlps_layers_all_dataset}
\end{centering}
\end{figure}

\section{Self-attention Logit Difference Update}\label{sec:update_logits_attn_gpt2_small}
\begin{figure}[H]
\begin{centering}
\includegraphics[width=0.34\textwidth]{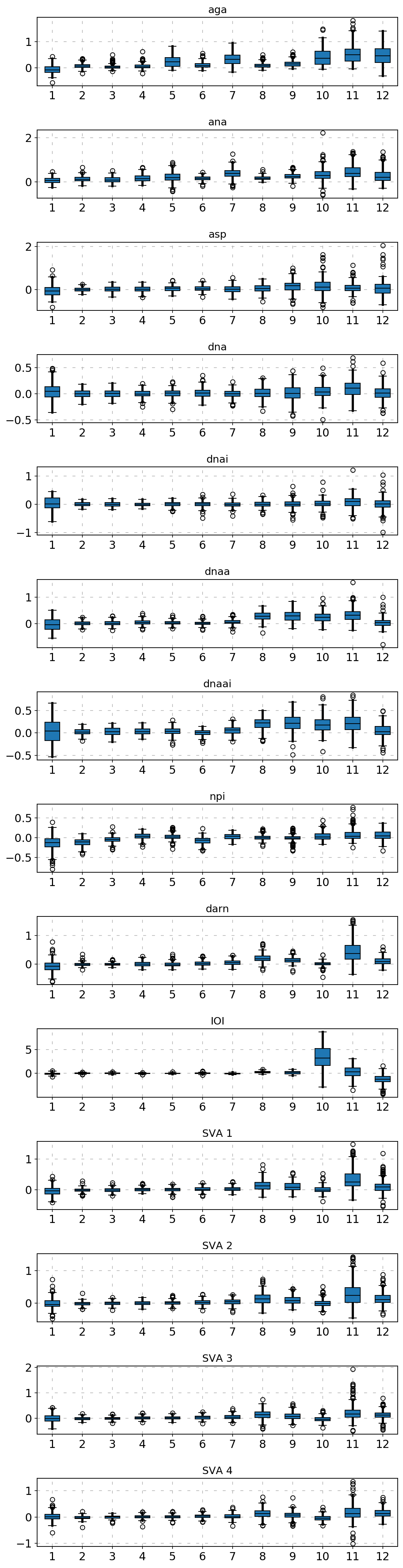}
\caption{Self-attention update to the logit difference $\Delta \text{logit}^l_{(w-f) \gets \text{Self-attn}^{l}}$ across layers (GPT-2 Small).}
\label{fig:gpt2_small_attn_layers_all_dataset}
\end{centering}
\end{figure}

\section{Qualitative Contrastive Exaplantions}
\subsection{Explanations of Different Contrastive Methods}
\begin{table}[H]
\centering
\resizebox{0.8\linewidth}{!}{ 
\begin{tabular}{l} \toprule
\multicolumn{1}{l}{Logit}\\ 
 \mybox{color451}{\strut{a}} \mybox{color567}{\strut{2006}} \mybox{color651}{\strut{guide}} \mybox{color478}{\strut{to}} \mybox{color507}{\strut{the}} \mybox{color666}{\strut{churches}} \mybox{color146}{\strut{of}} \mybox{color560}{\strut{anglesey}} \mybox{color501}{\strut{\textbf{says}}}\\

\addlinespace
\multicolumn{1}{l}{ALTI-Logit}\\ 
 \mybox{color489}{\strut{a}} \mybox{color658}{\strut{2006}} \mybox{color701}{\strut{guide}} \mybox{color371}{\strut{to}} \mybox{color441}{\strut{the}} \mybox{color514}{\strut{churches}} \mybox{color288}{\strut{of}} \mybox{color597}{\strut{anglesey}} \mybox{color501}{\strut{\textbf{says}}}\\
\addlinespace
\multicolumn{1}{l}{Erasure}\\ 
 \mybox{color599}{\strut{a}} \mybox{color516}{\strut{2006}} \mybox{color555}{\strut{guide}} \mybox{color1000}{\strut{to}} \mybox{color509}{\strut{the}} \mybox{color561}{\strut{churches}} \mybox{color381}{\strut{of}} \mybox{color541}{\strut{anglesey}} \mybox{color501}{\strut{\textbf{says}}}\\
\addlinespace
\multicolumn{1}{l}{Grad Norm}\\ 
 \mybox{color590}{\strut{a}} \mybox{color577}{\strut{2006}} \mybox{color587}{\strut{guide}} \mybox{color556}{\strut{to}} \mybox{color550}{\strut{the}} \mybox{color657}{\strut{churches}} \mybox{color553}{\strut{of}} \mybox{color933}{\strut{anglesey}} \mybox{color501}{\strut{\textbf{says}}}\\
\addlinespace
\multicolumn{1}{l}{\gradinput}\\ 
 \mybox{color454}{\strut{a}} \mybox{color355}{\strut{2006}} \mybox{color228}{\strut{guide}} \mybox{color524}{\strut{to}} \mybox{color496}{\strut{the}} \mybox{color354}{\strut{churches}} \mybox{color496}{\strut{of}} \mybox{color145}{\strut{anglesey}} \mybox{color501}{\strut{\textbf{says}}}\\
\bottomrule
\end{tabular}}
\caption{Comparison of different contrastive explanations on a GPT-2 Small \texttt{SVA} example (why \textbf{says} instead of \textbf{say}).}
\label{table:gpt_2_small_sva_example}
\end{table}

\begin{table}[H]
\centering
\footnotesize
\begin{tabular}{l} \toprule
\multicolumn{1}{l}{Logit}\\ 
 \mybox{color423}{\strut{Diane}} \mybox{color478}{\strut{should}} \mybox{color497}{\strut{complain}} \mybox{color551}{\strut{about}} \mybox{color1000}{\strut{these}} \mybox{color607}{\strut{unconvinced}} \mybox{color501}{\strut{\textbf{drivers}}}\\
\addlinespace
\multicolumn{1}{l}{ALTI-Logit}\\ 
 \mybox{color604}{\strut{Diane}} \mybox{color619}{\strut{should}} \mybox{color698}{\strut{complain}} \mybox{color652}{\strut{about}} \mybox{color847}{\strut{these}} \mybox{color584}{\strut{unconvinced}} \mybox{color501}{\strut{\textbf{drivers}}}\\
\addlinespace
\multicolumn{1}{l}{Erasure}\\ 
 \mybox{color416}{\strut{Diane}} \mybox{color458}{\strut{should}} \mybox{color616}{\strut{complain}} \mybox{color507}{\strut{about}} \mybox{color549}{\strut{these}} \mybox{color866}{\strut{unconvinced}} \mybox{color501}{\strut{\textbf{drivers}}}\\
\addlinespace
\multicolumn{1}{l}{Grad Norm}\\ 
 \mybox{color694}{\strut{Diane}} \mybox{color558}{\strut{should}} \mybox{color591}{\strut{complain}} \mybox{color554}{\strut{about}} \mybox{color562}{\strut{these}} \mybox{color1000}{\strut{unconvinced}} \mybox{color501}{\strut{\textbf{drivers}}}\\
\addlinespace
\multicolumn{1}{l}{\gradinput}\\ 
 \mybox{color355}{\strut{Diane}} \mybox{color404}{\strut{should}} \mybox{color226}{\strut{complain}} \mybox{color479}{\strut{about}} \mybox{color329}{\strut{these}} \mybox{color392}{\strut{unconvinced}} \mybox{color501}{\strut{\textbf{drivers}}}\\
\bottomrule
\end{tabular}
\caption{Comparison of different contrastive explanations on a GPT-2 Small \texttt{dnaa} example (why \textbf{drivers} instead of \textbf{driver}).}
\label{table:gpt_2_small_dnaa_6}
\end{table}

\begin{table}[H]
\centering
\footnotesize
\begin{tabular}{l} \toprule
\multicolumn{1}{l}{Logit}\\ 
 \mybox{color678}{\strut{Amanda}} \mybox{color352}{\strut{isn't}} \mybox{color830}{\strut{respected}} \mybox{color800}{\strut{by}} \mybox{color456}{\strut{the}} \mybox{color501}{\strut{\textbf{children}}}\\
\addlinespace
\multicolumn{1}{l}{ALTI-Logit}\\ 
 \mybox{color906}{\strut{Amanda}} \mybox{color609}{\strut{isn't}} \mybox{color886}{\strut{respected}} \mybox{color589}{\strut{by}} \mybox{color488}{\strut{the}} \mybox{color501}{\strut{\textbf{children}}}\\
\addlinespace
\multicolumn{1}{l}{Erasure}\\ 
 \mybox{color797}{\strut{Amanda}} \mybox{color191}{\strut{isn't}} \mybox{color665}{\strut{respected}} \mybox{color703}{\strut{by}} \mybox{color528}{\strut{the}} \mybox{color501}{\strut{\textbf{children}}}\\
\addlinespace
\multicolumn{1}{l}{Grad Norm}\\ 
 \mybox{color889}{\strut{Amanda}} \mybox{color629}{\strut{isn't}} \mybox{color744}{\strut{respected}} \mybox{color598}{\strut{by}} \mybox{color644}{\strut{the}} \mybox{color501}{\strut{\textbf{children}}}\\
\addlinespace
\multicolumn{1}{l}{\gradinput}\\ 
 \mybox{color660}{\strut{Amanda}} \mybox{color474}{\strut{isn't}} \mybox{color414}{\strut{respected}} \mybox{color109}{\strut{by}} \mybox{color164}{\strut{the}} \mybox{color501}{\strut{\textbf{children}}}\\
\bottomrule
\end{tabular}
\caption{Comparison of different contrastive explanations on a GPT-2 Small \texttt{asp} example (why \textbf{children} instead of \textbf{cups}).}
\label{table:gpt_2_small_asp_3}
\end{table}

\clearpage
\twocolumn
\subsection{GPT-2 XL ALTI-Logit across layers}
    \begin{table}[!h]
    \small
    \centering
    \resizebox{!}{.36\paperheight}{%
    \begin{tabular}{l} \toprule
    L48\;| \mybox{color534}{\strut{A}} \mybox{color550}{\strut{report}} \mybox{color513}{\strut{about}} \mybox{color498}{\strut{the}} \mybox{color511}{\strut{Impressionists}} \mybox{color501}{\strut{\textbf{has}}}\\
    \addlinespace
    L47\;| \mybox{color519}{\strut{A}} \mybox{color503}{\strut{report}} \mybox{color504}{\strut{about}} \mybox{color501}{\strut{the}} \mybox{color497}{\strut{Impressionists}} \mybox{color501}{\strut{\textbf{has}}}\\
    \addlinespace
    L46\;| \mybox{color515}{\strut{A}} \mybox{color498}{\strut{report}} \mybox{color500}{\strut{about}} \mybox{color501}{\strut{the}} \mybox{color487}{\strut{Impressionists}} \mybox{color501}{\strut{\textbf{has}}}\\
    \addlinespace
    L45\;| \mybox{color523}{\strut{A}} \mybox{color538}{\strut{report}} \mybox{color504}{\strut{about}} \mybox{color501}{\strut{the}} \mybox{color510}{\strut{Impressionists}} \mybox{color501}{\strut{\textbf{has}}}\\
    \addlinespace
    L44\;| \mybox{color506}{\strut{A}} \mybox{color511}{\strut{report}} \mybox{color502}{\strut{about}} \mybox{color501}{\strut{the}} \mybox{color509}{\strut{Impressionists}} \mybox{color501}{\strut{\textbf{has}}}\\
    \addlinespace
    L43\;| \mybox{color508}{\strut{A}} \mybox{color511}{\strut{report}} \mybox{color502}{\strut{about}} \mybox{color501}{\strut{the}} \mybox{color508}{\strut{Impressionists}} \mybox{color501}{\strut{\textbf{has}}}\\
    \addlinespace
    L42\;| \mybox{color507}{\strut{A}} \mybox{color511}{\strut{report}} \mybox{color498}{\strut{about}} \mybox{color500}{\strut{the}} \mybox{color481}{\strut{Impressionists}} \mybox{color501}{\strut{\textbf{has}}}\\
    \addlinespace
    L41\;| \mybox{color507}{\strut{A}} \mybox{color503}{\strut{report}} \mybox{color497}{\strut{about}} \mybox{color501}{\strut{the}} \mybox{color487}{\strut{Impressionists}} \mybox{color501}{\strut{\textbf{has}}}\\
\bottomrule
    \end{tabular}}
    \label{table:distractor_15_gpt2xl}
    \caption{GPT-2 XL \texttt{darn} (why \textbf{has} instead of \textbf{have}).}
    \end{table}

\begin{table}[H]
\vspace{15.25ex}
\small
\centering
\resizebox{!}{.36\paperheight}{%
\begin{tabular}{l} \toprule
L48\;| \mybox{color479}{\strut{Katherine}} \mybox{color490}{\strut{can't}} \mybox{color508}{\strut{help}} \mybox{color501}{\strut{\textbf{herself}}}\\
\addlinespace
L47\;| \mybox{color505}{\strut{Katherine}} \mybox{color502}{\strut{can't}} \mybox{color501}{\strut{help}} \mybox{color501}{\strut{\textbf{herself}}}\\
\addlinespace
L46\;| \mybox{color524}{\strut{Katherine}} \mybox{color509}{\strut{can't}} \mybox{color505}{\strut{help}} \mybox{color501}{\strut{\textbf{herself}}}\\
\addlinespace
L45\;| \mybox{color502}{\strut{Katherine}} \mybox{color510}{\strut{can't}} \mybox{color503}{\strut{help}} \mybox{color501}{\strut{\textbf{herself}}}\\
\addlinespace
L44\;| \mybox{color532}{\strut{Katherine}} \mybox{color500}{\strut{can't}} \mybox{color501}{\strut{help}} \mybox{color501}{\strut{\textbf{herself}}}\\
\addlinespace
L43\;| \mybox{color529}{\strut{Katherine}} \mybox{color510}{\strut{can't}} \mybox{color502}{\strut{help}} \mybox{color501}{\strut{\textbf{herself}}}\\
\addlinespace
L42\;| \mybox{color533}{\strut{Katherine}} \mybox{color504}{\strut{can't}} \mybox{color505}{\strut{help}} \mybox{color501}{\strut{\textbf{herself}}}\\
\addlinespace
L41\;| \mybox{color561}{\strut{Katherine}} \mybox{color522}{\strut{can't}} \mybox{color505}{\strut{help}} \mybox{color501}{\strut{\textbf{herself}}}\\
\bottomrule
\end{tabular}}
\label{table:aga_0_gpt2xl}
\caption{GPT-2 XL \texttt{aga} (why \textbf{herself} instead of \textbf{himself}).}
\end{table}

\clearpage
\subsection{ALTI-Logit (IOI) across Models}\label{apx:ioi_models}

    \vspace{15ex}
    \noindent\begin{minipage}{\textwidth}
    \resizebox{\linewidth}{!}{
    \centering
    \begin{tabular}{l} \toprule
    L12\;| \mybox{color498}{\strut{</s>}} \mybox{color493}{\strut{When}} \mybox{color479}{\strut{Paula}} \mybox{color494}{\strut{and}} \mybox{color468}{\strut{Martha}} \mybox{color497}{\strut{got}} \mybox{color499}{\strut{a}} \mybox{color503}{\strut{coconut}} \mybox{color501}{\strut{at}} \mybox{color500}{\strut{the}} \mybox{color497}{\strut{zoo,}} \mybox{color492}{\strut{Martha}} \mybox{color501}{\strut{decided}} \mybox{color501}{\strut{to}} \mybox{color501}{\strut{give}} \mybox{color501}{\strut{the}} \mybox{color501}{\strut{coconut}} \mybox{color501}{\strut{to}} \mybox{color501}{\strut{\textbf{Paula}}}\\
    \addlinespace
    L11\;| \mybox{color537}{\strut{</s>}} \mybox{color551}{\strut{When}} \mybox{color1000}{\strut{Paula}} \mybox{color496}{\strut{and}} \mybox{color459}{\strut{Martha}} \mybox{color515}{\strut{got}} \mybox{color504}{\strut{a}} \mybox{color504}{\strut{coconut}} \mybox{color503}{\strut{at}} \mybox{color502}{\strut{the}} \mybox{color517}{\strut{zoo,}} \mybox{color583}{\strut{Martha}} \mybox{color499}{\strut{decided}} \mybox{color501}{\strut{to}} \mybox{color500}{\strut{give}} \mybox{color500}{\strut{the}} \mybox{color500}{\strut{coconut}} \mybox{color500}{\strut{to}} \mybox{color501}{\strut{\textbf{Paula}}}\\
    \addlinespace
    L10\;| \mybox{color502}{\strut{</s>}} \mybox{color497}{\strut{When}} \mybox{color583}{\strut{Paula}} \mybox{color457}{\strut{and}} \mybox{color351}{\strut{Martha}} \mybox{color499}{\strut{got}} \mybox{color501}{\strut{a}} \mybox{color500}{\strut{coconut}} \mybox{color499}{\strut{at}} \mybox{color499}{\strut{the}} \mybox{color487}{\strut{zoo,}} \mybox{color494}{\strut{Martha}} \mybox{color501}{\strut{decided}} \mybox{color502}{\strut{to}} \mybox{color502}{\strut{give}} \mybox{color501}{\strut{the}} \mybox{color502}{\strut{coconut}} \mybox{color501}{\strut{to}} \mybox{color501}{\strut{\textbf{Paula}}}\\
\bottomrule
    \end{tabular}}
    \label{table:ioi_opt_40}
    \captionof{figure}{OPT 125M \texttt{IOI} (why \textbf{Paula} instead of \textbf{Martha}).}
    \end{minipage}

\begin{table*}[t]
\resizebox{\linewidth}{!}{
\centering
\begin{tabular}{l} \toprule
L24\;| \mybox{color494}{\strut{When}} \mybox{color494}{\strut{Paula}} \mybox{color499}{\strut{and}} \mybox{color498}{\strut{Martha}} \mybox{color497}{\strut{got}} \mybox{color498}{\strut{a}} \mybox{color499}{\strut{coconut}} \mybox{color500}{\strut{at}} \mybox{color498}{\strut{the}} \mybox{color502}{\strut{zoo,}} \mybox{color501}{\strut{Martha}} \mybox{color503}{\strut{decided}} \mybox{color500}{\strut{to}} \mybox{color501}{\strut{give}} \mybox{color501}{\strut{the}} \mybox{color502}{\strut{coconut}} \mybox{color501}{\strut{to}} \mybox{color501}{\strut{\textbf{Paula}}}\\
\addlinespace
L23\;| \mybox{color502}{\strut{When}} \mybox{color615}{\strut{Paula}} \mybox{color480}{\strut{and}} \mybox{color449}{\strut{Martha}} \mybox{color496}{\strut{got}} \mybox{color499}{\strut{a}} \mybox{color492}{\strut{coconut}} \mybox{color498}{\strut{at}} \mybox{color501}{\strut{the}} \mybox{color510}{\strut{zoo,}} \mybox{color523}{\strut{Martha}} \mybox{color490}{\strut{decided}} \mybox{color501}{\strut{to}} \mybox{color501}{\strut{give}} \mybox{color501}{\strut{the}} \mybox{color501}{\strut{coconut}} \mybox{color502}{\strut{to}} \mybox{color501}{\strut{\textbf{Paula}}}\\
\addlinespace
L22\;| \mybox{color501}{\strut{When}} \mybox{color531}{\strut{Paula}} \mybox{color482}{\strut{and}} \mybox{color438}{\strut{Martha}} \mybox{color501}{\strut{got}} \mybox{color499}{\strut{a}} \mybox{color494}{\strut{coconut}} \mybox{color497}{\strut{at}} \mybox{color498}{\strut{the}} \mybox{color489}{\strut{zoo,}} \mybox{color523}{\strut{Martha}} \mybox{color510}{\strut{decided}} \mybox{color503}{\strut{to}} \mybox{color507}{\strut{give}} \mybox{color500}{\strut{the}} \mybox{color505}{\strut{coconut}} \mybox{color506}{\strut{to}} \mybox{color501}{\strut{\textbf{Paula}}}\\
\bottomrule
\end{tabular}}
\label{table:ioi_bloom_560m_40}
\caption{BLOOM 560M \texttt{IOI} (why \textbf{Paula} instead of \textbf{Martha}).}
\end{table*}

\begin{table*}[t]
\resizebox{\linewidth}{!}{
\centering
\begin{tabular}{l} \toprule
L24\;| \mybox{color502}{\strut{When}} \mybox{color518}{\strut{Paula}} \mybox{color495}{\strut{and}} \mybox{color470}{\strut{Martha}} \mybox{color502}{\strut{got}} \mybox{color502}{\strut{a}} \mybox{color505}{\strut{coconut}} \mybox{color501}{\strut{at}} \mybox{color501}{\strut{the}} \mybox{color497}{\strut{zoo,}} \mybox{color494}{\strut{Martha}} \mybox{color494}{\strut{decided}} \mybox{color502}{\strut{to}} \mybox{color503}{\strut{give}} \mybox{color501}{\strut{the}} \mybox{color507}{\strut{coconut}} \mybox{color500}{\strut{to}} \mybox{color501}{\strut{\textbf{Paula}}}\\
\addlinespace
L23\;| \mybox{color500}{\strut{When}} \mybox{color513}{\strut{Paula}} \mybox{color502}{\strut{and}} \mybox{color469}{\strut{Martha}} \mybox{color511}{\strut{got}} \mybox{color502}{\strut{a}} \mybox{color503}{\strut{coconut}} \mybox{color506}{\strut{at}} \mybox{color502}{\strut{the}} \mybox{color509}{\strut{zoo,}} \mybox{color531}{\strut{Martha}} \mybox{color503}{\strut{decided}} \mybox{color501}{\strut{to}} \mybox{color504}{\strut{give}} \mybox{color500}{\strut{the}} \mybox{color502}{\strut{coconut}} \mybox{color503}{\strut{to}} \mybox{color501}{\strut{\textbf{Paula}}}\\
\addlinespace
L22\;| \mybox{color457}{\strut{When}} \mybox{color438}{\strut{Paula}} \mybox{color470}{\strut{and}} \mybox{color334}{\strut{Martha}} \mybox{color505}{\strut{got}} \mybox{color500}{\strut{a}} \mybox{color488}{\strut{coconut}} \mybox{color501}{\strut{at}} \mybox{color500}{\strut{the}} \mybox{color508}{\strut{zoo,}} \mybox{color549}{\strut{Martha}} \mybox{color497}{\strut{decided}} \mybox{color500}{\strut{to}} \mybox{color504}{\strut{give}} \mybox{color499}{\strut{the}} \mybox{color495}{\strut{coconut}} \mybox{color499}{\strut{to}} \mybox{color501}{\strut{\textbf{Paula}}}\\
\bottomrule
\end{tabular}}
\label{table:ioi_bloom_1b1_40}
\caption{BLOOM 1B1 \texttt{IOI} (why \textbf{Paula} instead of \textbf{Martha}).}
\end{table*}

\begin{table*}[t]
\resizebox{\linewidth}{!}{
\centering
\begin{tabular}{l} \toprule
L12\;| \mybox{color484}{\strut{When}} \mybox{color557}{\strut{Paula}} \mybox{color466}{\strut{and}} \mybox{color395}{\strut{Martha}} \mybox{color494}{\strut{got}} \mybox{color497}{\strut{a}} \mybox{color497}{\strut{coconut}} \mybox{color497}{\strut{at}} \mybox{color498}{\strut{the}} \mybox{color491}{\strut{zoo,}} \mybox{color489}{\strut{Martha}} \mybox{color498}{\strut{decided}} \mybox{color501}{\strut{to}} \mybox{color499}{\strut{give}} \mybox{color501}{\strut{the}} \mybox{color501}{\strut{coconut}} \mybox{color501}{\strut{to}} \mybox{color501}{\strut{\textbf{Paula}}}\\
\addlinespace
L11\;| \mybox{color702}{\strut{When}} \mybox{color1000}{\strut{Paula}} \mybox{color448}{\strut{and}} \mybox{color348}{\strut{Martha}} \mybox{color498}{\strut{got}} \mybox{color500}{\strut{a}} \mybox{color504}{\strut{coconut}} \mybox{color500}{\strut{at}} \mybox{color500}{\strut{the}} \mybox{color498}{\strut{zoo,}} \mybox{color490}{\strut{Martha}} \mybox{color502}{\strut{decided}} \mybox{color501}{\strut{to}} \mybox{color501}{\strut{give}} \mybox{color501}{\strut{the}} \mybox{color501}{\strut{coconut}} \mybox{color501}{\strut{to}} \mybox{color501}{\strut{\textbf{Paula}}}\\
\addlinespace
L10\;| \mybox{color516}{\strut{When}} \mybox{color829}{\strut{Paula}} \mybox{color376}{\strut{and}} \mybox{color112}{\strut{Martha}} \mybox{color493}{\strut{got}} \mybox{color498}{\strut{a}} \mybox{color499}{\strut{coconut}} \mybox{color498}{\strut{at}} \mybox{color498}{\strut{the}} \mybox{color493}{\strut{zoo,}} \mybox{color467}{\strut{Martha}} \mybox{color499}{\strut{decided}} \mybox{color501}{\strut{to}} \mybox{color502}{\strut{give}} \mybox{color501}{\strut{the}} \mybox{color502}{\strut{coconut}} \mybox{color502}{\strut{to}} \mybox{color501}{\strut{\textbf{Paula}}}\\
\bottomrule
\end{tabular}}
\label{table:ioi_gpt2_40}
\caption{GPT-2 Small \texttt{IOI} (why \textbf{Paula} instead of \textbf{Martha}).}
\end{table*}

\begin{table*}[t]
\resizebox{\linewidth}{!}{
\centering
\resizebox{!}{.36\paperheight}{%
\begin{tabular}{l} \toprule
L36\;| \mybox{color518}{\strut{When}} \mybox{color521}{\strut{Paula}} \mybox{color503}{\strut{and}} \mybox{color504}{\strut{Martha}} \mybox{color503}{\strut{got}} \mybox{color501}{\strut{a}} \mybox{color506}{\strut{coconut}} \mybox{color502}{\strut{at}} \mybox{color501}{\strut{the}} \mybox{color506}{\strut{zoo,}} \mybox{color503}{\strut{Martha}} \mybox{color502}{\strut{decided}} \mybox{color501}{\strut{to}} \mybox{color506}{\strut{give}} \mybox{color501}{\strut{the}} \mybox{color502}{\strut{coconut}} \mybox{color501}{\strut{to}} \mybox{color501}{\strut{\textbf{Paula}}}\\
\addlinespace
L35\;| \mybox{color501}{\strut{When}} \mybox{color472}{\strut{Paula}} \mybox{color499}{\strut{and}} \mybox{color456}{\strut{Martha}} \mybox{color501}{\strut{got}} \mybox{color501}{\strut{a}} \mybox{color504}{\strut{coconut}} \mybox{color501}{\strut{at}} \mybox{color501}{\strut{the}} \mybox{color506}{\strut{zoo,}} \mybox{color499}{\strut{Martha}} \mybox{color507}{\strut{decided}} \mybox{color501}{\strut{to}} \mybox{color507}{\strut{give}} \mybox{color501}{\strut{the}} \mybox{color504}{\strut{coconut}} \mybox{color501}{\strut{to}} \mybox{color501}{\strut{\textbf{Paula}}}\\
\addlinespace
L34\;| \mybox{color518}{\strut{When}} \mybox{color568}{\strut{Paula}} \mybox{color503}{\strut{and}} \mybox{color502}{\strut{Martha}} \mybox{color509}{\strut{got}} \mybox{color502}{\strut{a}} \mybox{color522}{\strut{coconut}} \mybox{color503}{\strut{at}} \mybox{color502}{\strut{the}} \mybox{color515}{\strut{zoo,}} \mybox{color515}{\strut{Martha}} \mybox{color514}{\strut{decided}} \mybox{color502}{\strut{to}} \mybox{color516}{\strut{give}} \mybox{color501}{\strut{the}} \mybox{color507}{\strut{coconut}} \mybox{color501}{\strut{to}} \mybox{color501}{\strut{\textbf{Paula}}}\\
\bottomrule
\end{tabular}}}
\label{table:ioi_gpt2_large_40}
\caption{GPT-2 Large \texttt{IOI} (why \textbf{Paula} instead of \textbf{Martha}).}
\end{table*}

\begin{table*}[t]
\small
\centering
\resizebox{!}{.4\paperheight}{%
\begin{tabular}{l} \toprule
L48\;| \mybox{color511}{\strut{When}} \mybox{color505}{\strut{Paula}} \mybox{color500}{\strut{and}} \mybox{color489}{\strut{Martha}} \mybox{color503}{\strut{got}} \mybox{color501}{\strut{a}} \mybox{color504}{\strut{coconut}} \mybox{color500}{\strut{at}} \mybox{color500}{\strut{the}} \mybox{color501}{\strut{zoo,}} \mybox{color501}{\strut{Martha}} \mybox{color501}{\strut{decided}} \mybox{color501}{\strut{to}} \mybox{color501}{\strut{give}} \mybox{color500}{\strut{the}} \mybox{color501}{\strut{coconut}} \mybox{color501}{\strut{to}} \mybox{color501}{\strut{\textbf{Paula}}}\\
\addlinespace
L47\;| \mybox{color500}{\strut{When}} \mybox{color494}{\strut{Paula}} \mybox{color499}{\strut{and}} \mybox{color484}{\strut{Martha}} \mybox{color500}{\strut{got}} \mybox{color500}{\strut{a}} \mybox{color500}{\strut{coconut}} \mybox{color501}{\strut{at}} \mybox{color501}{\strut{the}} \mybox{color502}{\strut{zoo,}} \mybox{color501}{\strut{Martha}} \mybox{color501}{\strut{decided}} \mybox{color500}{\strut{to}} \mybox{color501}{\strut{give}} \mybox{color501}{\strut{the}} \mybox{color501}{\strut{coconut}} \mybox{color501}{\strut{to}} \mybox{color501}{\strut{\textbf{Paula}}}\\
\addlinespace
L46\;| \mybox{color508}{\strut{When}} \mybox{color514}{\strut{Paula}} \mybox{color503}{\strut{and}} \mybox{color507}{\strut{Martha}} \mybox{color504}{\strut{got}} \mybox{color502}{\strut{a}} \mybox{color507}{\strut{coconut}} \mybox{color502}{\strut{at}} \mybox{color501}{\strut{the}} \mybox{color509}{\strut{zoo,}} \mybox{color504}{\strut{Martha}} \mybox{color505}{\strut{decided}} \mybox{color502}{\strut{to}} \mybox{color503}{\strut{give}} \mybox{color501}{\strut{the}} \mybox{color502}{\strut{coconut}} \mybox{color501}{\strut{to}} \mybox{color501}{\strut{\textbf{Paula}}}\\
\bottomrule
\end{tabular}}
\label{table:ioi_40_gpt2xl}
\caption{GPT-2 XL \texttt{IOI} (why \textbf{Paula} instead of \textbf{Martha}).}
\end{table*}

\end{document}